\documentclass{article}

\PassOptionsToPackage{sort&compress}{natbib}
\usepackage[preprint]{corl_2026} % Use this for the initial submission.
\usepackage{amsmath}
\usepackage{amsfonts}
\usepackage{booktabs}
\usepackage{graphicx}
\usepackage{multirow}
\usepackage{enumitem}
\usepackage{fontawesome5}
\usepackage[most]{tcolorbox}
\usepackage{array}
\usepackage{placeins}
\usepackage{caption}
\usepackage{stfloats}
\newcolumntype{L}[1]{>{\raggedright\arraybackslash}p{#1}}
\newcolumntype{C}[1]{>{\centering\arraybackslash}p{#1}}
\usepackage{wrapfig}
\newtcbox{\linkpill}{
  on line,
  box align=base,
  colback=gray!10,
  colframe=gray!10,
  arc=1.5mm,
  boxrule=0pt,
  left=2pt,
  right=2pt,
  top=1pt,
  bottom=1pt
}
\usepackage{wrapfig}
\definecolor{slsorange}{HTML}{528C03}

\title{Steering Robustness into World Action Models via Mechanistic Interpretability and Optimal Control}

\author{
  Jihoon Hong$^\star$, Julian Skifstad$^\star$, Qiyue Dai, Alice Chan, Glen Chou\\
  Georgia Institute of Technology\\
  \texttt{\{jhong392, jskifstad3, qdai41, ichan30, chou\}@gatech.edu} \\
    $^\star$Equal contribution\\
\linkpill{\href{https://trustworthyrobotics.github.io/steering_robust_wam_site/}{\textcolor{slsorange}{\faGlobe\ Website}}}
\quad
\linkpill{\href{https://github.com/trustworthyrobotics/steering_robustness_WAMs}{\textcolor{slsorange}{\faGithub\ Code}}}
\quad
\linkpill{\href{https://www.youtube.com/watch?v=067He19-TyQ}{\textcolor{slsorange}{\faYoutube\ Video}}}
}

\begin{document}
\maketitle

%===============================================================================

\begin{abstract}
\looseness-1World Action Models (WAMs) enable semantically- and physically-informed control but are brittle under distribution shift. In this work, we use mechanistic interpretability to study how robustness-relevant perturbations are represented in WAM activation space. Comparing activations across successful and unsuccessful rollouts, we find some WAM architectures exhibit low-dimensional linear separability for robustness-critical features, while others do not. This motivates the use of contrastive activation directions for training-free WAM steering. We also show that local linearity in WAM activation dynamics enables efficient feedback steering via model-based optimal control, yielding World-Action Linear Quadratic Regulator (WA-LQR), a minimally-invasive reduced-order LQR controller. 
Via mechanistic evaluations, we predict strong steerability in the Cosmos-Policy and DiT4DiT models but weak steerability in LingBot-VA, consistent with steering intervention results.
On Cosmos-Policy and DiT4DiT, WA-LQR generalizes contrastive directions to new tasks and improves robustness to camera, gripper, and visual-noise perturbations over unsteered and prompt steering baselines.

\end{abstract}

\vspace{-10pt}

%===============================================================================
\begin{figure}[h]
\vspace{-12pt}
    \centering
    \includegraphics[width=0.95\linewidth]{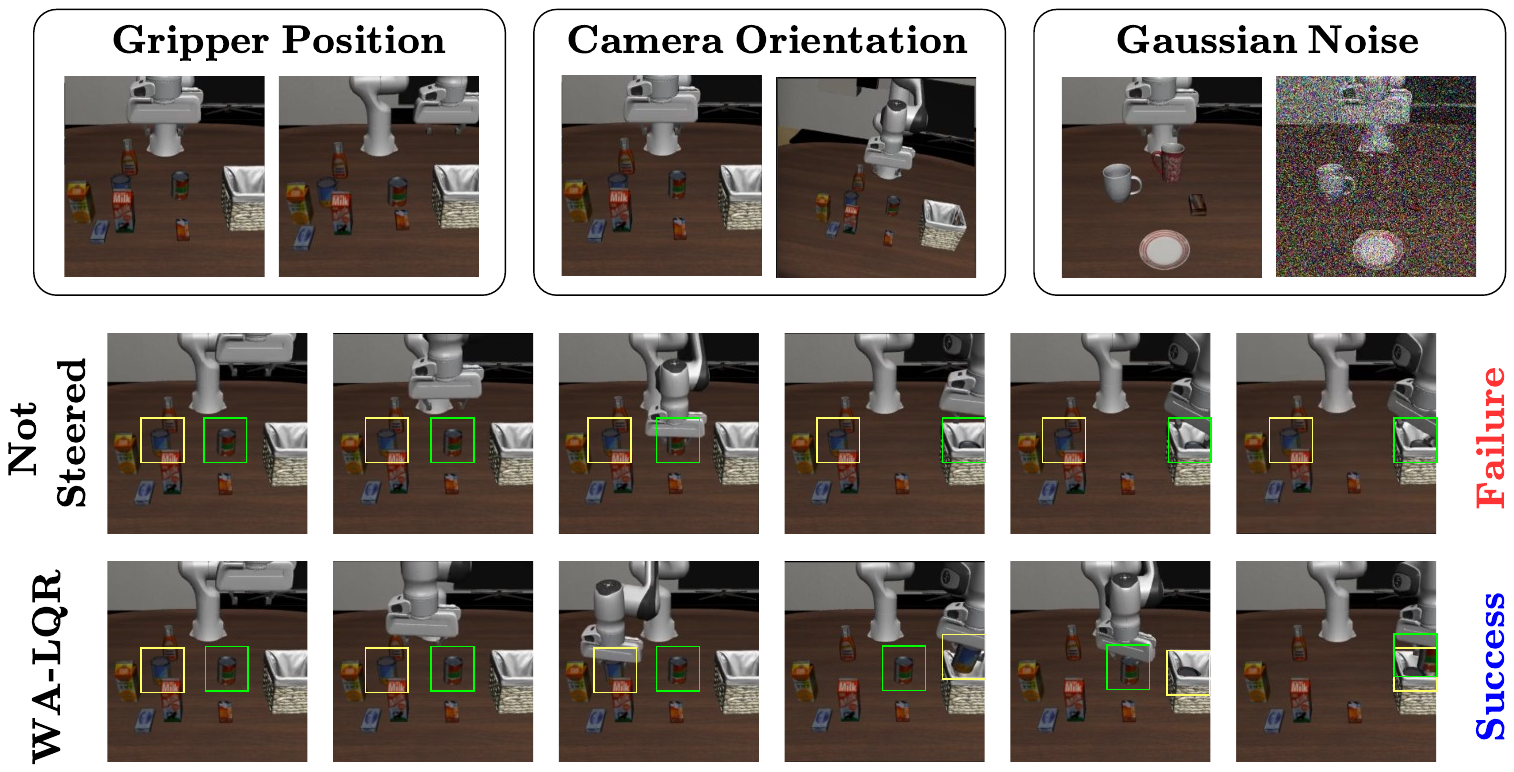}\vspace{-7pt}
    \caption{WA-LQR makes World Action Models more robust to perturbations including gripper-position changes, camera-orientation shifts, and Gaussian sensor noise. In these Cosmos-Policy examples from LIBERO-10, the yellow and green boxes mark the two objects that must be placed in the basket. Without steering, the WAM fails; with WA-LQR, it succeeds.}
    \label{fig:teaser}
    \vspace{-5pt}
\end{figure}

\vspace{-15pt}
\section{Introduction}
\vspace{-8pt}

Foundation models have advanced robot learning through policies that generalize across objects, tasks, and environments. While VLA models map observations and instructions directly to actions, World Action Models (WAMs) additionally couple action prediction with future-state modeling through video-generative backbones \cite{kim2024openvla, kim2026cosmos, ye2026world, ye2026gigaworld, zhang2026world}. This makes WAMs a promising route toward dynamically-coherent robot policies. However, they remain brittle under out-of-distribution shifts, including camera changes, robot initial-state perturbations, and visual corruption \cite{zhang2026world}. Because such nuisance factors are unavoidable in deployment and cannot be exhaustively covered with training data, it is essential to improve robustness without extensive new data collection or slow retraining.

In this paper, we ask if the internal activations of WAMs reveal why they fail under such perturbations, and if these representations can be used to improve robustness without finetuning. We study perturbations to camera position, initial gripper position, and Gaussian image noise. Using mechanistic interpretability (MI), we compare activations from nominal and perturbed rollouts to test for simple linear geometric activation space structure. By evaluating the degree of linear separability across models, we find this structure is perturbation- and architecture-dependent, suggesting that some WAMs contain steerable representations of robustness-critical features, while others may not.

By leveraging the degree of linear separability as a predictor of steerability, we can identify models that are well suited to training-free WAM activation steering, in which contrastive examples are used to construct directions that distinguish nominal from perturbed behavior. We use these directions for open-loop and closed-loop robustness interventions. In particular, we introduce \emph{World-Action Linear Quadratic Regulator} (WA-LQR), a reduced-order optimal-control method that projects activations into a low-dimensional contrastive subspace and uses local linear dynamics to synthesize a closed-loop LQR steering controller. Unlike open-loop activation addition, WA-LQR adapts online, steering only when activations deviate from the target feature strength while penalizing large perturbations.
We evaluate our mechanistic predictions and steering method on Cosmos-Policy \cite{kim2026cosmos}, DiT4DiT \cite{ma2026dit4dit}, and LingBot-VA \cite{lingbot-va2026}. Our low-dimensional linear separability analysis predicts strong steerability for Cosmos-Policy and DiT4DiT and weak steerability for LingBot-VA, which empirical intervention results validate. On Cosmos-Policy, contrastive directions transfer across LIBERO tasks, and WA-LQR improves robustness to camera, gripper, and visual-noise perturbations over non-steered, prompt-steering, and several open-loop steering baselines. These results show that MI can diagnose WAM robustness and guide inference-time interventions. Our contributions are: 
\begin{itemize}[leftmargin=1.1em]
    \item We conduct a mechanistic analysis of WAM activations under robustness-relevant perturbations, identifying when nuisance features exhibit low-dimensional linear separability. 
    \item We show that steerability is architecture-dependent: Cosmos-Policy \cite{kim2026cosmos} and DiT4DiT \cite{ma2026dit4dit} exhibit clear linear structure across multiple perturbations, whereas LingBot-VA \cite{lingbot-va2026} exhibits substantially weaker separability. Moreover, we show that separability is strongly correlated with steerability, making it a useful diagnostic for identifying models that are amenable to activation steering.
    \item We leverage these insights to construct contrastive activation directions for training-free WAM steering. First, we adapt open-loop activation addition techniques from the LLM literature to WAMs. We then show that the local linearity of the diffusion transformer dynamics in a reduced WAM activation space enables the efficient closed-form synthesis of \textit{closed-loop} steering controllers via WA-LQR. To the best of our knowledge, these are the first open- and closed-loop activation steering methods for WAMs.
    \item We evaluate WA-LQR on robustness benchmarks, showing that MI-based steerability predictions strongly correlate with intervention outcomes and that WA-LQR improves robustness on Cosmos-Policy across multiple perturbation types.
\end{itemize}

\vspace{-15pt}
\section{Related Work}
\vspace{-10pt}

\noindent\textbf{WAMs.}
\looseness-1 Foundation models have enabled general-purpose robotic policies that map observations to actions \cite{brohan2022rt,zitkovich2023rt,team2024octo,o2024open,kim2024openvla,black2024pi0,kim2025fine,pertsch2025fast}. VLA models are often reactive action predictors and do not explicitly model how the world evolves under robot interventions, limiting long-horizon reasoning and robustness under distribution shift \cite{hou2026world,zhang2026world}. WAMs address this by coupling action generation with future-state prediction, often adapting video generative backbones to robotic data \cite{ye2026world,kim2026cosmos,ye2026gigaworld,wang2026world}. Existing WAMs include inverse-dynamics-style models, where predicted futures are decoded into actions \cite{du2023learning,wen2024vidman,hu2024video,jia2025video2act,li2026causal}, and unified video-action models, where future states and actions are generated jointly \cite{kim2026cosmos,ye2026world,ye2026gigaworld}. We study both families and show that steerable, robot-relevant activation structure is architecture-dependent.

\noindent\textbf{Robustness of Action Models.}
Despite progress in general-purpose robot policies, robustness remains a major deployment obstacle. Benchmarks such as VLATest, COLOSSEUM, and LIBERO-Plus expose brittleness to camera viewpoint, robot initial state, object layout, lighting, background texture, sensor noise, and language phrasing \cite{liu2023libero,li2024evaluating,wang2025vlatest,pumacay2024colosseum,fei2025libero}. Prior studies suggest that robustness gains often require data diversity, wrist-camera observations, RL post-training, or robustness-oriented fine-tuning rather than arising from inherently stable representations \cite{zhou2025exploringlimitsvisionlanguageactionmanipulations,liu2026can,tan2025interactive,guo2025improving}. WAMs aim to improve robustness with future-state modeling and video-based priors \cite{hu2024video,kim2026cosmos,li2026causal,ye2026world,ye2026gigaworld,yuan2026fast}, but still fail under shifts such as camera viewpoint and robot initial-state changes \cite{zhang2026world}. This motivates inference-time methods that improve robustness without exhaustive perturbation data or costly retraining.

\noindent\textbf{MI and Activation Steering.}
MI identifies internal representations that modulate model behavior \cite{bereska2024mechanistic,sharkey2025open}. In LLMs, many semantic features appear approximately linear in activation space, motivating activation steering: inference-time hidden-state modifications that change behavior without retraining \cite{Elhage_Hume_Olsson_2022,park2023linear,marks2024the,zou2023representation,lee2025programming}. Most methods compute contrastive directions from examples with and without a target concept, then add or transform activations to steer behaviors \cite{Dathathri2020Plug,li2023inference,turner2024activation,rimsky2024steering,arditi2024refusal,rodriguez2025controlling,wu2024reft,Vu_Nguyen_2025}. Because these interventions are often layer-local and open-loop, recent work models activations as dynamical systems and uses feedback for more targeted steering \cite{bhargava2023s,Kong_Wang_Mu_Du_Zhuang_Zhou_Song_Zhang_Wang_Zhang_2024,Cheng_Alonso_2025,Nguyen_Vu_Pham_Zhang_Nguyen_2025,skifstad2026local}. Activation steering has also begun to extend to image and video generators \cite{rodriguez2025lineas,facchiano2025video,ekin2026unreasonable,hong2026activation}, where semantics are distributed across text, spatial, temporal, timestep, and layer representations. WAMs add a further challenge: their activations affect not only generated visual content, but also action-relevant predictions that determine robot behavior. 

Relatively little work applies MI to robotic foundation models, and existing studies focus on VLAs. Prior work identifies steerable VLA activation directions \cite{haon25mechanistic}, formalizes feature observability and controllability \cite{buurmeijer2026observing}, finetunes task-relevant attention heads \cite{mitra2025mechanistic}, discovers SAE-based motion primitives \cite{swann2026sparse}, measures causal reliance on visual regions \cite{zhang2026embodied}, and uses activation injection, probes, sparse latents, or conceptor subspaces to analyze and steer VLA behavior \cite{grant2026not,khan2025controlling,miao2026coast}. In contrast, we study WAMs, whose DiT-style video backbones jointly encode future visual states, action dynamics, and control outputs. We show that WAM steerability is architecture-dependent and develop feedback-based steering that treats WAM inference as a reduced-order dynamical system.

\section{Preliminaries and Problem Statement}
\vspace{-8pt}

\paragraph{Linear Quadratic Regulator (LQR)}

The LQR problem \eqref{eq:lqr_base} \cite{hespanha2018linear} seeks a controller that minimizes a quadratic state-control cost \eqref{eq:lqr_base} for linear time-varying dynamics \eqref{eq:dynamics}
\begin{subequations}\label{eq:lqr_base}
\begin{align}
\min_{\{u_k\}_{k=1}^{H-1}}
\quad &
\mathcal{L}
:=
\sum_{k=1}^{H-1}
\left(
z_k^\top Q_k z_k
+
u_k^\top R_k u_k
\right)
+
z_H^\top Q_H z_H
\\
\text{subject to}\quad &
z_{k+1}=A_k z_k+B_k u_k,
\qquad
\forall k=1,\ldots,H-1.
\label{eq:dynamics}
\end{align}
\end{subequations}
where $Q_k\succeq0$ penalizes state error and $R_k\succ0$ penalizes control effort. The optimal policy has closed form
$u_k^\ast=-K_kz_k$, where $K_k$ is obtained efficiently via Riccati recursions. 
Given a set of nominal setpoints $\{(\bar{z}_k, \bar{u}_k)\}_{k=1,\cdots,H}$, \eqref{eq:lqr_base} can be generalized to penalize deviations $\delta z_k:=z_k-\bar{z}_k$ and $\delta u_k:=u_k-\bar{u}_k$ from the setpoints.
By modifying \eqref{eq:lqr_base} to
\begin{subequations}\label{eq:lqr_tracking}
\begin{align}
\min_{\{\delta u_k\}_{k=1}^{H-1}}\quad
& \sum_{k=1}^{H-1} \left( \delta z_k^\top Q_k \delta z_k + \delta u_k^\top R_k \delta u_k \hspace{-1pt}\right) + \delta z_H^\top Q_H \delta z_H
\label{eq:err_lqr_objective} \\
\text{subject to} \quad\ \
& \delta z_{k+1} = A_k \delta z_k + B_k \delta u_k, \qquad k = 1,\dots,H-1,
\end{align}
\end{subequations}
the solution admits a closed-form tracking controller $u_k^{\ast}:=\bar{u}_k-K_k\delta z_k$.
When the dynamics $z_{k+1}=f(z_k,u_k)$ are non-linear, similar formulations to \eqref{eq:lqr_tracking} are possible by letting $A_k=\nabla_{z_k}f(z_k,u_k)$ and $B_k=\nabla_{u_k}f(z_k,u_k)$, and approximating $\delta z_{k+1}\approx A_k\delta z_k + B_k \delta u_k$ assuming local linearity.

% \vspace{-7pt}
\paragraph{WAM Architectures}

While the two types of WAMs vary in architecture, they both build on Diffusion Transformer (DiT) based video generation models \cite{peebles2023scalable}.
Starting from a random latent action representation $\hat{x}_T\sim\mathcal{N}(0,I)$, these models are used to gradually denoise it to a clean $x_{\text{out}}$ by 
\begin{equation}\label{eq:scheduler}
    \hat{x}_{t-1}=\texttt{STEP}(\hat{x}_t,t,M(\hat{x}_t,t,h)),
    \qquad
    x_{\text{out}}=\texttt{DEC}(\hat{x}_1),
\end{equation}
where $h$ is the embedding of the task prompt $p$, $M$ is the model, $\texttt{STEP}$ is a choice of ODE solver, and $\texttt{DEC}$ decodes latent actions.
Typically, $M$ is a sequence of $L$ DiT layers $\phi^{(l)}$ for $l=0,\cdots,L-1$:
\begin{equation}
    x_{0,t}:=\texttt{Tok}(\hat{x}_{t}),
    \qquad
    x_{l+1,t}=\phi^{(l)}(x_{l,t},t,h),
    \qquad
    M(\hat{x}_t,t,h):=\texttt{Detok}(x_{L,t}).
\end{equation}
After block $L$, the WAM output is detokenized and passed to the model scheduler \eqref{eq:scheduler}. We treat the scheduler transition in \eqref{eq:scheduler} as the mechanism that chains together $T$ independent $L$-horizon within-denoising-timestep controllers. 
In this work, we will perform activation steering by adding control inputs inside the DiT blocks. 
Let $u_{l,t}$ denote these perturbations. The steered WAM block is
\begin{equation}
\label{eq:steered_transformer_wam}
\begin{aligned}
    x_{l+1,t}
    &
    =f_{l,t}(x_{l,t},u_{l,t}):=
    \phi^{(l)}(x_{l,t}, t, h) + u_{l,t}.
\end{aligned}
\end{equation}

\paragraph{Problem Statement}

\looseness-1In this paper, we use MI to identify robustness-relevant features in WAM activation space and use these features for training-free robustness improvement via activation steering.

\looseness-1\textit{Problem 1: Linear feature discovery in WAM activation space (Sec. \ref{sec:mechanistic}).} Given a WAM, determine whether a target nuisance feature is represented by approximately linear directions in WAM activation space. Concretely, for each layer $l$ and denoising timestep $t$, we seek low-dimensional projections $P_{l,t}$ and linear directions $v_{l,t}$ such that the projected activations of $\xi^+$ and $\xi^-$ are separable.

\textit{Problem 2: Training-free robustness steering (Sec. \ref{sec:activation_steering})}. 
Given a WAM, synthesize an open- or closed-loop inference-time steering policy that modifies activations and modulates the features discovered in Problem 1.

% \vspace{-8pt}
\section{A Mechanistic Study for Interpreting WAMs}\label{sec:mechanistic}
\vspace{-8pt}

\looseness-1A central assumption in the activation steering literature is that model activation spaces exhibit simple, interpretable geometric structure, across domains including LLMs, VLAs, and video generation models \cite{turner2024activation,rimsky2024steering,skifstad2026local,hong2026activation,haon25mechanistic}. Specifically, most steering algorithms assume linear semantic feature directions, as determined by the emergent linear separability of activations corresponding to semantically contrastive inputs \cite{arditi2024refusal,marks2024geometry}. However, the existence of this structure in WAMs is not guaranteed, despite the steerable foundation model backbone. Indeed, prior work has shown that such representations are fragile to the finetuning process undergone by robotics foundation models \cite{haon25mechanistic,Huang_Zhang_Azarcon_Chou_Kira_2025}.

\begin{figure}
    \centering
    \includegraphics[width=\linewidth]{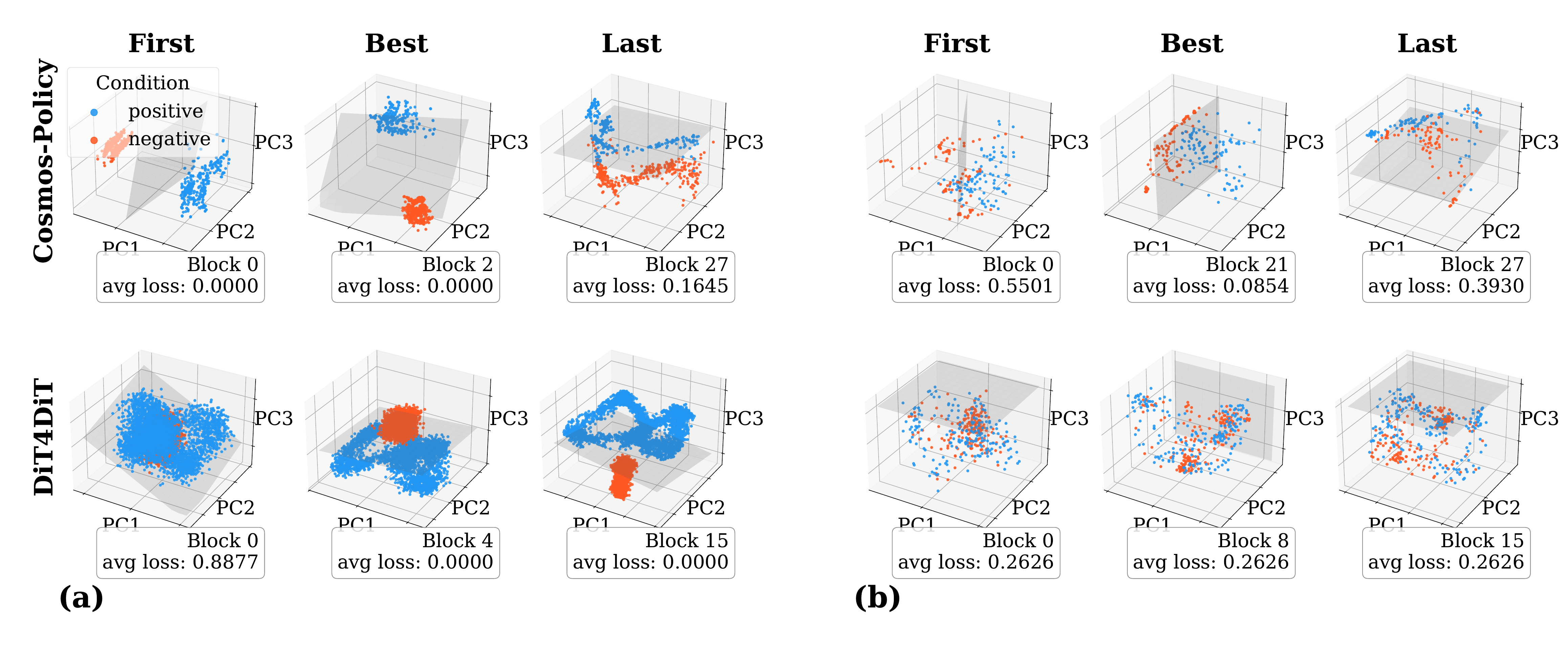}
    \vspace{-20pt}
    \caption{On Task 0 of LIBERO-10 \cite{liu2023libero}, we evaluate \textbf{(a)} activations corresponding to \textit{noise-perturbed} and clean inputs on the first, best intermediate, and final DiT block residual stream for Cosmos-Policy 2B and DiT4DiT \cite{kim2026cosmos,ma2026dit4dit}. Activations are projected onto the top three principal components, with the reported SVM (gray) and hinge loss. \textbf{(b)} Repeated for \textit{camera perturbations}.}
    \label{fig:noise_pca_3d}
    \vspace{-10pt}
\end{figure}

\paragraph{Setup} We study the emergent geometry in robustness features for WAMs, specifically Cosmos-Policy 2B, DiT4DiT, and LingBot-VA \cite{kim2026cosmos,ma2026dit4dit,lingbot-va2026}. We consider contrastive datasets related to three key sources of sensitivity in WAM manipulation tasks: perturbation to initial gripper position, initial camera position, and corruption of camera inputs with Gaussian noise. We collect activations corresponding to each dataset through a full model forward pass. That is, for prompts $p_+ \in \mathcal{D_+},$ $p_- \in \mathcal{D_-}$ where, e.g., $\mathcal{D_+} = \{\texttt{Clean inputs} \}$, $\mathcal{D_-} = \{\texttt{Noised inputs} \}$, we collect DiT activations $x_{t,\ell}^{p_+}$ and $x_{t,\ell}^{p_-}$ for all $t, \ell$. Typically, DiT activations have a shape of $(F,H,W,D)$, for some token frame, height, width and hidden dimension $F < H \leq W \ll D$, making direct storage intractable for a substantial sample size. Thus, for this section, we consider an average pooling over the token position, resulting in a summarized activation $\bar x \in \mathbb{R}^D$ for all sampled activations. We also perform an average pooling over robot action chunk timesteps; see Sec. \ref{subsec:activation_addition} for details.

Given contrastive sets of activations $\{\bar x^+_k \}_{k}$ and $\{\bar x^-_k \}_{k}$, we define a \textit{contrastive direction}
\begin{equation}\label{eq:contrastive_dir}
    d_k = \bar x^+_k - \bar x^-_k.
\end{equation}
Inspired by \cite{marks2024geometry}, we consider a simple geometric evaluation of the separability of contrastive datasets via principal component analysis (PCA) conducted on the set of contrastive directions, $\{d_k\}_k$. With the objective of enabling linear feature-based steering, we seek to identify linear separability of contrastive datasets. Note that in the high dimensionality of the original mean-pooled system, $N$ samples with $N \ll D$ are trivially linearly separable, rendering analysis in the full-dimensional space uninformative. Hence, we consider a low-dimensional approximation of the data. In fact, we find in a later section (Sec.~\ref{subsec:results_properties}) that the contrastive directions are well summarized by as few as three principal components.

Projecting the contrastive activations $\bar x^+$ and $\bar x^-$ onto the top three PCs, we construct a quantitative metric for the linear separability of the pairs of contrastive datasets. In particular, we fit a linear support vector machine (SVM) to the contrastive activations in three dimensions, and measure the classification loss as the average hinge loss per sample \cite{kecman2005svm}, 
\begin{equation}\label{eq:hinge_loss}
    \texttt{loss}(\mathcal{D}_+, \mathcal{D}_-) = \frac{1}{N} \sum_{i\in \left[N\right]} \max(0, 1-y_i(w^T\bar{x}_i+b)),
\end{equation}
\vspace{-12pt}

where $\{x\mid w^Tx+b=0\}$ denotes the SVM hyperplane and $y_i\in\{-1,1\}$. Intuitively, this measures how well the three-dimensional linear SVM distinguishes the two datasets, where perfect classification yields a loss of 0, and random classification yields a loss of 1. 

% \vspace{-8pt}
\begin{wrapfigure}{r}{.59\textwidth}
\vspace{-20pt}
    \centering
    \includegraphics[width=1\linewidth]{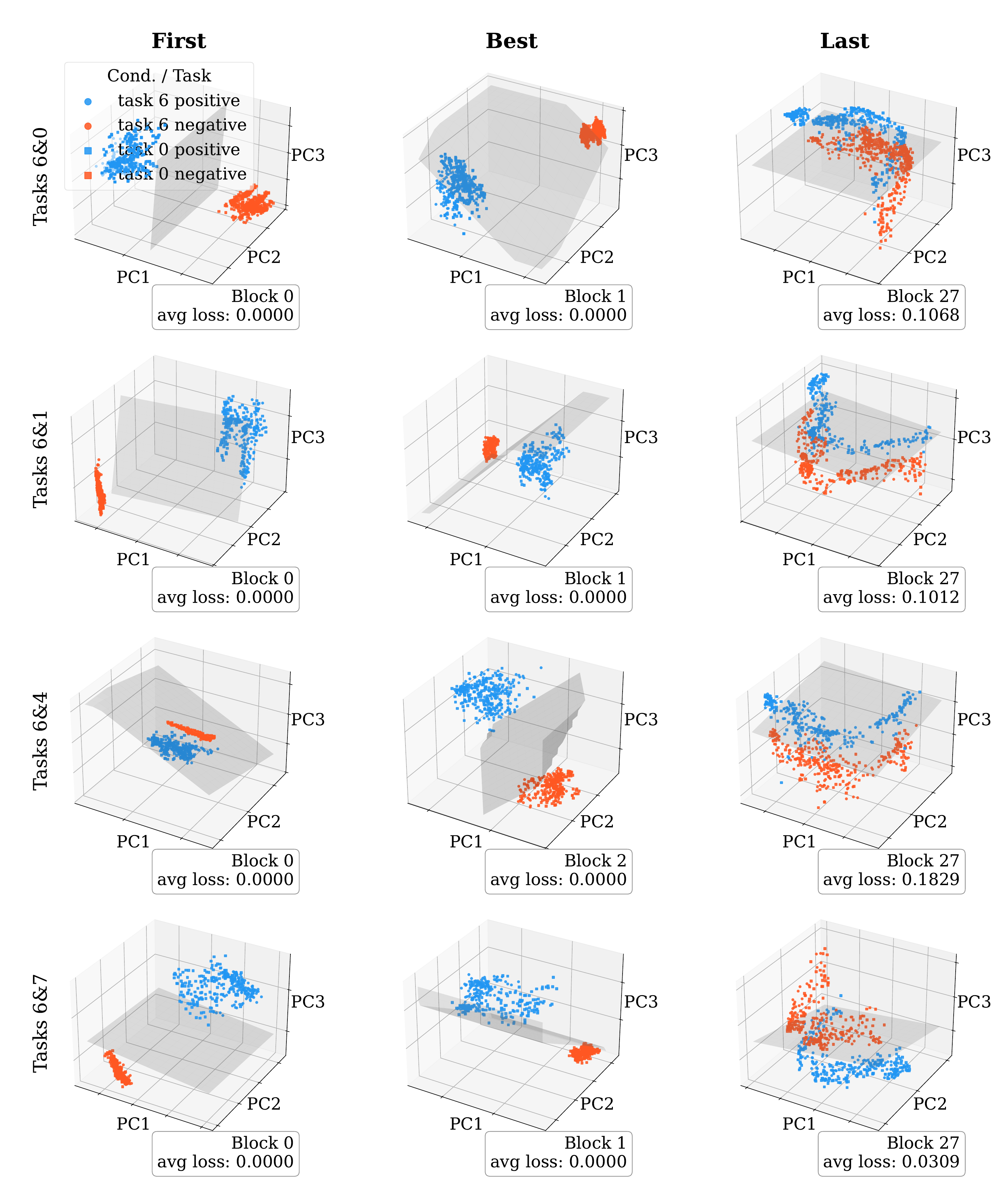}\vspace{-8pt}
    \caption{\looseness-1Pairwise separability for Cosmos-Policy under Gaussian noise corruption. Across task pairs, high linear separability and shared feature clusters suggest reusable representations that can be exploited for activation steering.}\vspace{-20pt}
    \label{fig:cosmos_noise_pairs}
% \vspace{-10pt}
\end{wrapfigure}

\vspace{-8pt}
\paragraph{Results} We perform this analysis across the perturbations for all 10 tasks of the LIBERO-10 dataset \cite{liu2023libero}, and visualize the results for Cosmos-Policy in Fig.~\ref{fig:noise_pca_3d} (further evaluations are provided in App.~\ref{ap:separability}, including for DiT4DiT \cite{ma2026dit4dit}). Notably, we find that the separability of different features is highly task-dependent, indicating that the latent representation corresponding to the same perturbation is not always shared across different scenes and tasks. A key observation of this study is that \emph{certain combinations of tasks share feature representations} corresponding to the same input perturbation -- see Fig.~\ref{fig:cosmos_noise_pairs} for an example with Cosmos-Policy. Using these clustered features, we are able to construct steering objectives which \textit{generalize} across tasks, as discussed in the following sections. This observation enables the activation steering formulation in Sec.~\ref{sec:activation_steering}.

\begin{figure}
    \centering
    \begin{minipage}[t]{0.64\linewidth}
        \vspace{0pt}
        \centering
        \includegraphics[width=\linewidth]{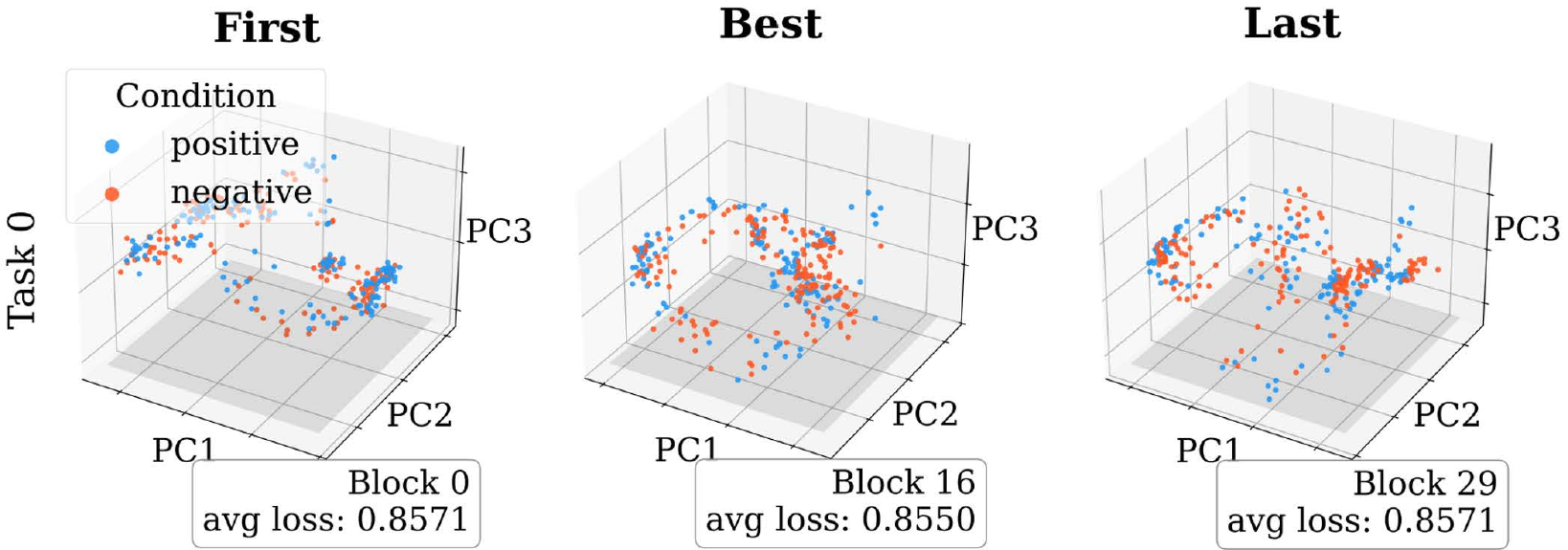}
    \end{minipage}
    \hfill
    \begin{minipage}[t]{0.34\linewidth}
        \vspace{0pt}
        \caption{
        Activations from camera-perturbed and clean LingBot-VA inputs at the first, best intermediate, and final transformer block residual streams. We observe much weaker separability than in the Cosmos-Policy setting.
        }
        \label{fig:noise_pca_3d_ling}
    \end{minipage}
    \vspace{-10pt}
\end{figure}

Applying the same analysis to the action inference activations for LingBot-VA yields substantially milder results. Across tasks and perturbations, we observe little to no separability as reported by the SVM loss, as well as qualitatively by inspection (see Fig.~\ref{fig:noise_pca_3d_ling}).

% \vspace{-9pt}
\section{Activation Steering for Robustifying WAMs}\label{sec:activation_steering}
% \vspace{-10pt}

\looseness-1 Motivated by the MI results of Sec. \ref{sec:mechanistic}, we give an overview of our steering method. We first construct contrastive vectors isolating the desired robustness feature and discuss a simple open-loop method for steering WAMs with the contrastive vectors (Sec. \ref{subsec:activation_addition}). We also propose a reduced-order LQR-based steering approach (Sec. \ref{subsec:latentactivationdynamics}) to enable scalable control-theoretic steering for WAMs. 

\subsection{Open-Loop Activation Addition from Contrastive Vectors}
\label{subsec:activation_addition}
\vspace{-8pt}

Given paired inputs
$(\xi_n^+,\xi_n^-)$ that differ primarily in a desired feature, e.g., nominal
versus perturbed camera pose or clean versus corrupted observations, we compute \textit{contrastive activation
directions} by subtracting hidden states from the two forward passes (App. \ref{app:contrastive}). For layer $l$, denoising timestep $t$, and action-chunk timestep
$\tau\in[H_a]$, this gives
\begin{equation}
\label{eq:wamlqr_actadd_contrastive}
    d_{l,t,\tau}^{(n)}
    :=
    x_{l,t,\tau}^{(\xi_n^+)}
    -
    x_{l,t,\tau}^{(\xi_n^-)}.
\end{equation}

\vspace{-8pt}
We thus describe the simplest contrastive steering method:
\emph{activation addition} (ActAdd)~\cite{turner2024activation}. ActAdd assumes
that the target feature is roughly linear in activation space. A steering
vector is formed by averaging contrastive directions
\eqref{eq:wamlqr_actadd_contrastive} over pairs/action-chunk positions 
\begin{equation}
    a_{l,t}
    :=
    \frac{1}{NH_a}
    \sum_{n=1}^{N}
    \sum_{\tau=1}^{H_a}
    d_{l,t,\tau}^{(n)}.
\end{equation} 
At inference time, ActAdd steers by adding $a_{l,t}$ with strength $\gamma$:
\begin{equation}
\label{eq:wamlqr_actadd}
    x_{l,t,\tau}
    \leftarrow
    x_{l,t,\tau}
    +
    \gamma a_{l,t},
    \qquad
    \forall \tau\in[H_a].
\end{equation}
\looseness-1The same direction is applied across all action-chunk timesteps, with $\gamma$
setting steering strength: positive values push toward $\xi^+$, while negative
values reverse the effect. ActAdd is training-free and weight-preserving, but
open-loop: it ignores the current activation, WAM dynamics, and whether
the feature is already at the desired strength, which can cause oversteering and
action degradation.

\subsection{Steering Latent World Activation Dynamics via WA-LQR}
\label{subsec:latentactivationdynamics}

\looseness-1WA-LQR generalizes ActAdd with feedback control, adapting the T2V steering method of \cite{hong2026activation} to WAMs. Instead of adding a fixed
$\gamma a_{l,t}$, it projects contrastive vectors into a latent
feature space, measures deviation from a latent setpoint, and computes
a minimum-cost LQR intervention. Thus, it preserves ActAdd's training-free
nature while adapting interventions to the realized WAM activation.

\paragraph{Dimensionality Reduction} 
As in T2V models \cite{hong2026activation}, full activation-space LQR is infeasible due to the high dimensionality of WAM activations. We instead assume (and validate in Sec. \ref{sec:results}) that robustness-relevant factors lie largely in a low-dimensional contrastive subspace. For each layer-denoising pair $(l,t)$, we construct this subspace by pooling contrastive directions \eqref{eq:wamlqr_actadd_contrastive} over prompt pairs and action-chunk timesteps, then applying streaming randomized singular value decomposition (SVD) \cite{halko2011finding} to the matrix whose rows are
$\{d_{l,t,\tau}^{(n)}:n\in[N],\ \tau\in[H_a]\}$.
This yields a compact orthonormal basis
$V_{l,t}\in\mathbb R^{d_x\times d_z}$ with $d_z\ll d_x$. 
Define the projection matrix
$P_{l,t}
    :=
    V_{l,t}^\top
    \in\mathbb R^{d_z\times d_x},$
which is shared across robot action-chunk timesteps $\tau \in [H_a]$. 
For each $(l,t,\tau)$, define the latent activation 
    $z_{l,t,\tau}
    :=
    P_{l,t}x_{l,t,\tau}
    \in\mathbb R^{d_z}$.
Let $A_{l,t}$ and $B_{l,t}$ denote the raw Jacobians of the controlled block dynamics in \eqref{eq:steered_transformer_wam} along a nominal trajectory:
    $A_{l,t}
    :=
    \frac{\partial f_{l,t}}{\partial x}
    \big|_{\bar x_{l,t},\bar u_{l,t}}$, 
    $B_{l,t}
    :=
    \frac{\partial f_{l,t}}{\partial u}
    \big|_{\bar x_{l,t},\bar u_{l,t}}$.
The reduced latent dynamics are
\begin{equation}
\label{eq:wamlqr_latent_dynamics}
    \delta z_{l+1,t,\tau}
    \approx
    \widetilde A_{l,t}\delta z_{l,t,\tau}
    +
    \widetilde B_{l,t}\delta u_{l,t,\tau},
    \qquad
    l=0,\ldots,L-1,
\end{equation}
where $\widetilde A_{l,t}
    :=
    P_{l+1,t}A_{l,t}P_{l,t}^\top
    \in\mathbb R^{d_z\times d_z}$ and $
    \widetilde B_{l,t}
    :=
    P_{l+1,t}B_{l,t}
    \in\mathbb R^{d_z\times d_u}$. 
Neither $A_{l,t}$ nor $B_{l,t}$ is materialized explicitly; products with $\widetilde A_{l,t}$ and $\widetilde B_{l,t}$ are computed efficiently using Jacobian-vector products (JVPs) or vector-Jacobian products (VJPs).

\paragraph{Defining Feature Setpoints}

We define feature setpoints in the latent space for robustness-relevant WAM features. We first compute a latent contrastive direction averaged over action chunk indices
\begin{equation} 
e^z_{l,t}
    :=
    \frac{1}{NH_a}
    \sum_{n=1}^{N}
    \sum_{\tau=1}^{H_a}
    P_{l,t}
    (
        x_{l,t,\tau}^{(\xi_n^+)}
        -
        x_{l,t,\tau}^{(\xi_n^-)}
    ), \qquad
    v^z_{l,t}
    :=
    \frac{e^z_{l,t}}{\|e^z_{l,t}\|_2}. 
\end{equation}
For a realized latent activation $z_{l,t,\tau}$, the feature strength is
$\beta^z_{l,t,\tau} := (v^z_{l,t})^\top z_{l,t,\tau}$.
We set the desired feature strength as
$\beta_{l,t}^{z,\ast} := \lambda \|e^z_{l,t}\|_2$, 
where $\lambda$ controls steering strength. The feature tracking error for action-chunk timestep $\tau$ is
$\alpha_{l,t,\tau}
    :=
    \beta_{l,t}^{z,\ast}
    -
    (v^z_{l,t})^\top z_{l,t,\tau}$, 
    with $\delta z_{l,t,\tau}
    :=
    -\alpha_{l,t,\tau}v^z_{l,t}.$
Thus, $\delta z_{l,t,\tau}$ is the latent tracking error that, if corrected, would bring the chunk-$\tau$ activation to the desired feature setpoint along the contrastive direction.

\vspace{-4pt}
\paragraph{Reaching Feature Setpoints via WA-LQR}

Using the latent dynamics in \eqref{eq:wamlqr_latent_dynamics}, we compute LQR controllers that steer WAM activations toward the feature setpoints. Unlike the T2V setting \cite{hong2026activation}, we do not solve one $T\!\times\!L$-step LQR over both transformer blocks and scheduler transitions. Instead, for each chunk index $\tau$ and denoising timestep $t$, we solve an independent $L$-step LQR over the transformer blocks:
\begin{equation}
\label{eq:wamlqr_per_timestep_problem}
\begin{aligned}
    \min_{\{\delta u_{l,t}\}_{l=0}^{L-1}}
    \quad
    &
    \sum_{l=0}^{L-1}
    \left(
        \delta z_{l,t}^\top Q_{l,t}\delta z_{l,t}
        +
        \delta u_{l,t}^\top
        R^{\mathrm{chunk}}_{l,t}(\tau)
        \delta u_{l,t}
    \right)
    +
    \delta z_{L,t}^\top Q_{L,t}\delta z_{L,t}
    \\
    \mathrm{s.t.}
    \quad
    &
    \delta z_{l+1,t}
    =
    \widetilde A_{l,t}\delta z_{l,t}
    +
    \widetilde B_{l,t}\delta u_{l,t},
    \qquad
    l=0,\ldots,L-1.
\end{aligned}
\end{equation}
    
The control penalty incorporates an action-decay schedule over robot action chunk indices. Let $\tau\in\{0,\ldots,H_a-1\}$ index the action chunks executed over
time. For chunk $\tau$, we set
$r(\tau)
    :=
    \min\!\left(
        R_{\mathrm{final}},
        R_{\mathrm{init}}\exp(\tau/\tau_R)
    \right)$, 
where $R_{\mathrm{init}}$ is the initial steering penalty,
$R_{\mathrm{final}}$ is a large saturation value, and $\tau_R$ controls the
decay rate. The LQR control-cost matrix at chunk $\tau$ is then
$R_{l,t}^{\mathrm{chunk}}(\tau)
    :=
    r(\tau) I_{d_u}$, 
which is directly used in LQR \eqref{eq:wamlqr_per_timestep_problem}. Since $r(\tau)$ increases with
$\tau$ and saturates at $R_{\mathrm{final}}$, steering is strongest for early
chunks and gradually decays as the robot proceeds. A larger $\tau_R$ slows this
growth, causing more chunks to be steered before saturation. In our experiments,
$R_{\mathrm{final}}$ is chosen large enough that saturated chunks receive
negligible steering. Notably, \eqref{eq:wamlqr_per_timestep_problem} can be efficiently solved in $\mathcal{O}(Ld_z^3)$ time \cite{rawlings2020model} on the CPU or $\mathcal{O}(\log L \cdot \log^2 d_z)$ on the GPU \cite{fang2026safe}.

For each action chunk index $\tau$ and denoising timestep $t$, solving \eqref{eq:wamlqr_per_timestep_problem} yields gains
$K_{l,t,\tau}\in\mathbb R^{d_u\times d_z}$, which are used only within the corresponding denoising pass. At inference time, the controller is
\begin{equation}
\label{eq:wamlqr_controller}
    u_{l,t,\tau}^\ast
    :=
    \bar u_{l,t,\tau}
    -
    K_{l,t,\tau}\delta z_{l,t,\tau}
    =
    \bar u_{l,t,\tau}
    +
    K_{l,t,\tau}\alpha_{l,t,\tau}
    v^z_{l,t}.
\end{equation}

% \vspace{-4pt}
When $\bar u_{l,t,\tau}=0$, the intervention magnitude is proportional to the online feature-tracking error for robot action chunk $\tau$. The chunk-specific policy $K_{l,t,\tau}$ is then applied to the activation at the corresponding action-chunk index. 
The full WA-LQR procedure chains the $T$ per-timestep controllers through WAM inference. For action chunk index $\tau$ and denoising timestep $t$, we apply the $L$ feedback gains
$\{K_{0,t,\tau},\ldots,K_{L-1,t,\tau}\}$ inside the DiT blocks, as in \eqref{eq:steered_transformer_wam}. At each chunk index $\tau$, the steered block output $x_{L,t}$ is then passed to the scheduler, which produces $\hat x_{t-1}$ and initializes the next denoising pass, yielding $T$ chained $L$-step LQR controllers. Compared with open-loop ActAdd, WA-LQR adapts to the realized latent activation at each layer and denoising timestep, steering only when the WAM deviates from the desired robustness feature setpoint.

\vspace{-8pt}
\section{Results}\label{sec:results}
\vspace{-8pt}

In this section, we evaluate whether the mechanistic structure identified above can be used to improve WAM robustness through activation steering. We first assess the validity of the assumptions made by WA-LQR to justify its applicability to WAM steering (Sec. \ref{subsec:results_properties}). Next, we compare open-loop activation addition (ActAdd) and closed-loop WA-LQR across Cosmos-Policy, DiT4DiT, and LingBot-VA under OOD perturbations to camera orientation, initial gripper position, and camera noise (Sec. \ref{subsec:steering_results}). Overall, the results show that steering is effective when feature-relevant activations are linearly separable, improving success rates by up to 41\%. WA-LQR further improves robustness while helping avoid oversteering in settings where open-loop control is less reliable. See App. \ref{app:experimental_details} for further experimental details.

\begin{figure*}[!b]
\centering
\vspace{-20pt}
\hspace{-25pt}
    \includegraphics[width=\linewidth]{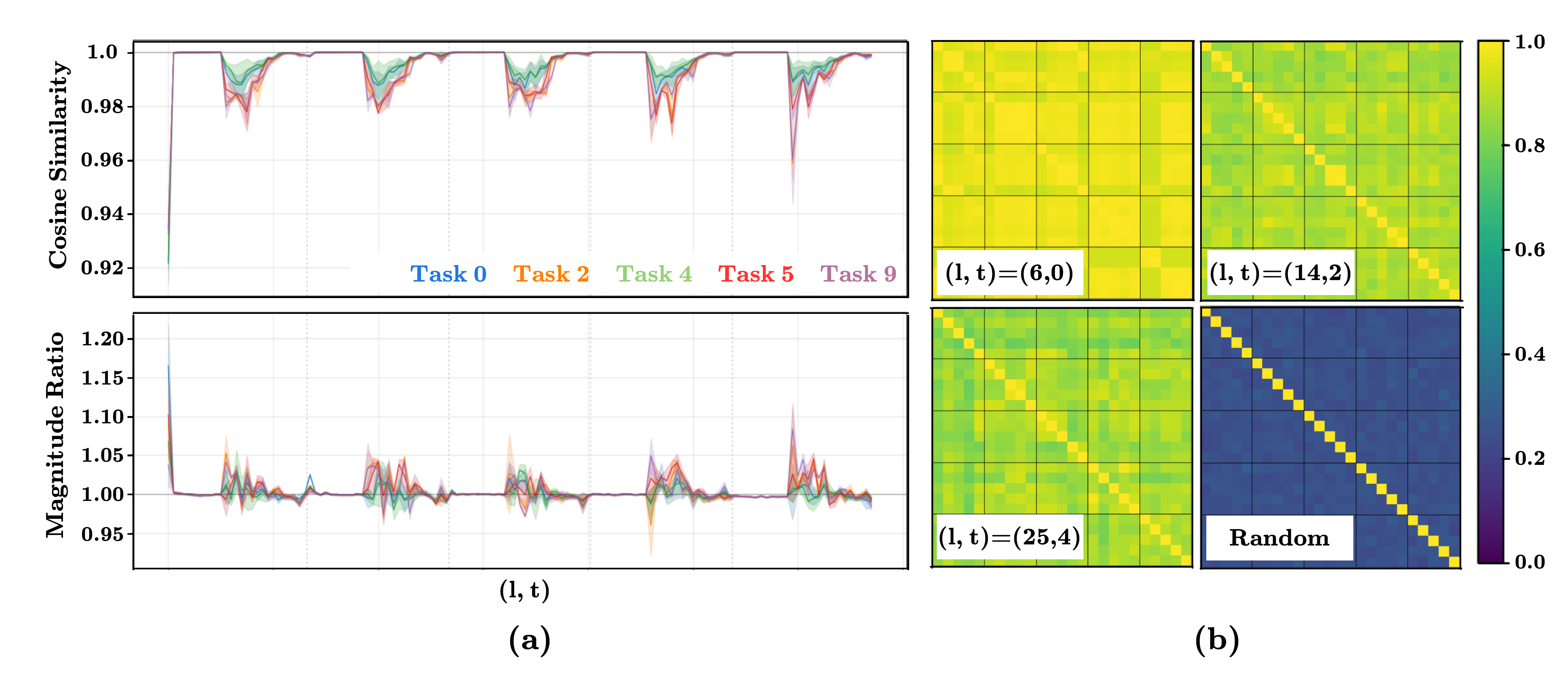}\vspace{-16pt}
    \caption{\looseness-1\textbf{(a)} The cosine similarity/magnitude ratio between the linear approximation using $\tilde{A}_{l,t}$ and actual latent activation, under change in camera orientation. \textbf{(b)} Overlap between subspaces spanned by 16 top right singular vectors of $\tilde{A}_{l,t}$ matrices, obtained from 25 random inputs across 5 tasks. \vspace{-10pt}}
    % \vspace{-10pt}
    \label{fig:linearity}
\end{figure*}

\vspace{-5pt}
\subsection{Activation Properties}\label{subsec:results_properties}
\vspace{-5pt}

\begin{wrapfigure}{r}{.58\textwidth}
\centering
\vspace{-15pt}
    \includegraphics[width=0.9\linewidth]{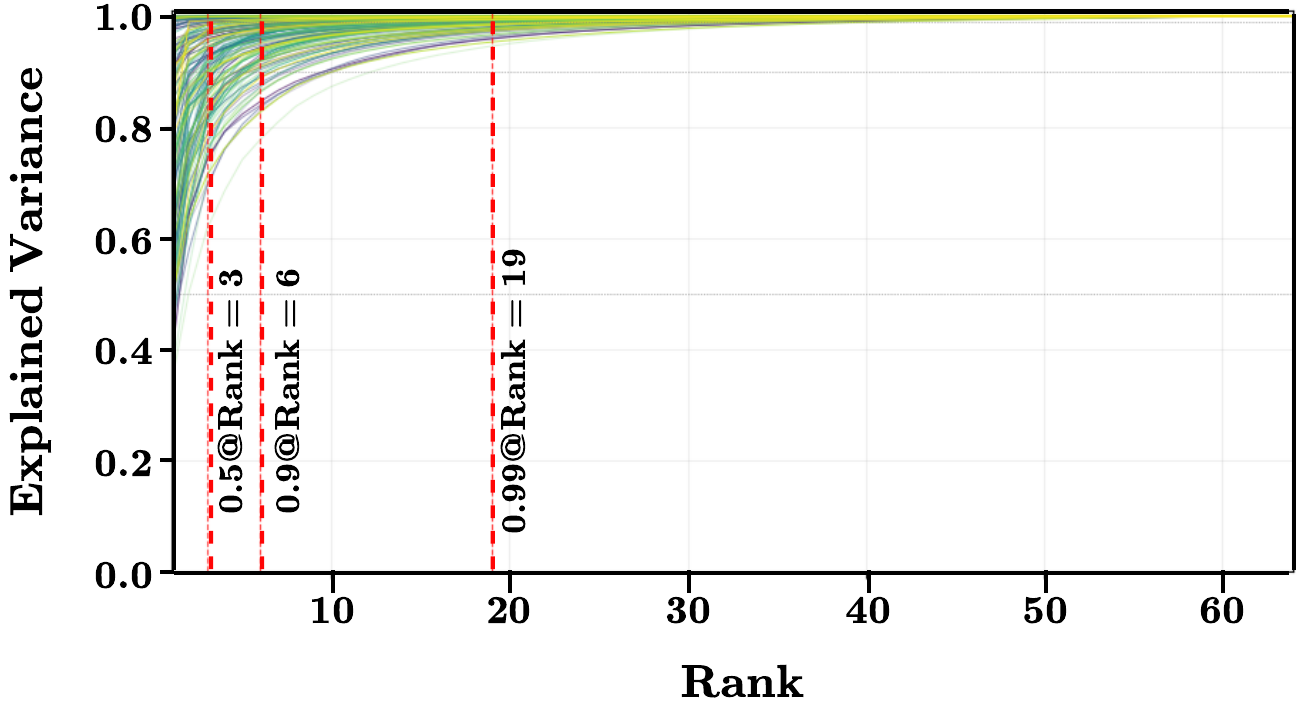}
    % \vspace{-10pt}
    \caption{Cumulative variance of latent contrastive vectors for camera orientation perturbation on Cosmos-Policy 2B \cite{kim2026cosmos} explained by the top-$k$ singular vectors.} \vspace{-10pt}
    % \vspace{-8pt}
    \label{fig:result_lowdim}
\end{wrapfigure}

\looseness-1We first empirically verify the assumptions that enable LQR-based steering of latent activations: 1) preservation of the information in the contrastive vectors in the latent subspace and 2) linearity of the WAM dynamics within the subspace.
We first study contrastive information preservation. In Fig.~\ref{fig:result_lowdim} we show that the variance of contrastive vectors projected to a 64-dimensional latent subspace, $P_{l,t}(x_{l,t,\tau}^{(\xi_n^+)} -x_{l,t,\tau}^{(\xi_n^-)})$, is mostly captured by the top few dimensions.
This is caused by rapid decay of singular values, which suggests that the influence of perturbations such as the change in camera orientation on activations manifests in a significantly lower dimensional subspace compared to raw activations.

To assess dynamics linearity, we observe that WAMs are locally linear (aligned with recent discoveries in LLM \cite{skifstad2026local} and T2V \cite{hong2026activation} models), which allows them to be effectively steered using LQR.
Fig.~\ref{fig:linearity}\textbf{(a)} shows the cosine similarity and the magnitude ratio between $P_{l+1,t}\phi^{(l)}\left(x_{l,t}+P_{l,t}^T\epsilon,t,h\right)$ and $P_{l+1,t}\phi^{(l)}\left(x_{l,t},t,h\right)+\tilde{A}_{l,t}\epsilon$ for random perturbation $\epsilon$ whose norm is proportional to that of $\|x_{l,t}\|$.
The cosine similarities and magnitude ratios remain close to 1 throughout inference, demonstrating that the latent dynamics are well captured by a first-order approximation.
Furthermore, we find that the matrices $\tilde{A}_{l,t}$ are highly similar across different inputs.

Motivated by the concentration of variance on the top few singular vectors, we visualize in Fig.~\ref{fig:linearity}\textbf{(b)} the overlap between the subspaces spanned by the top 16 right singular vectors of $\tilde{A}_{l,t}$ over 25 random inputs from 5 different tasks, measured by the mean squared cosine of principal angles (as inspired by a similar metric in \cite{skifstad2026local}).
While not shown here, the left singular vectors also demonstrate significant overlap, enabling the reuse of $\tilde{A}_{l,t}$ computed from a single input for steering model behavior on other tasks.

\begin{table}[t]
\centering
\caption{LIBERO-10 \cite{liu2023libero} success rates on Cosmos-Policy 2B \cite{kim2026cosmos}. Task $i\rightarrow$ Task $j$ denotes all $P_{l,t}, \tilde{A}_{l,t}, \tilde{B}_{l,t}$ matrices, $e^z_{l,t}$ vectors are computed with task $i$, and used to steer task $j$. 30 trials/task.}
\renewcommand{\arraystretch}{0.9}
\setlength{\tabcolsep}{9pt}
\label{tab:cosmos_steering_results}
% \resizebox{\textwidth}{!}{%
% \begin{tabular}{lc cccc}
\begin{tabular}{L{1.6cm} C{2.35cm} *{4}{C{1.5cm}}}
% \begin{tabular}{L{1.5cm} C{0.8cm} C{1.4cm} @{\hspace{12pt}} C{1.4cm} @{\hspace{12pt}} C{1.4cm} @{\hspace{12pt}} C{1.4cm}}
\toprule
\textbf{Perturbation} & \textbf{Tasks} & \textbf{No Steering} & \textbf{Prompt Steering} \cite{Wang_Chen_Liu_Ye_Chen_Lu_Liu_Yu_Jia_2026} & \textbf{ActAdd} \cite{turner2024activation} & \textbf{WA-LQR} \\%& \textbf{Hinge Loss $(\downarrow)$} \\
\midrule

\multirow{6}{2cm}{\textbf{Camera Orientation}} 
 & Task 0 $\rightarrow$ Task 0 & 33.3\%$\pm$8.6\%  & 40.0\%$\pm$8.9\%  & 43.3\%$\pm$9.0\%  & \textbf{60.0\%$\pm$8.9\%} \\%& 0.59 \\
 & Task 0 $\rightarrow$ Task 2  & 63.3\%$\pm$8.8\%  & 60.0\%$\pm$8.9\%  & \textbf{73.3\%$\pm$8.1\%}  & 70.0\%$\pm$8.4\% \\%& 0.55  \\
 & Task 0 $\rightarrow$ Task 4  & 10.0\%$\pm$5.5\%  & 6.7\%$\pm$4.6\%   & \textbf{16.7\%$\pm$6.8\%}  & 13.3\%$\pm$6.2\% \\%& 0.48 \\
 & Task 0 $\rightarrow$ Task 5  & 56.7\%$\pm$9.0\%  & 70.0\%$\pm$8.4\%  & 56.7\%$\pm$9.0\%  & \textbf{83.3\%$\pm$6.8\%} \\%& 0.57 \\
 & Task 0 $\rightarrow$ Task 9  & 66.7\%$\pm$8.6\%  & 56.7\%$\pm$9.0\%  & 56.7\%$\pm$9.0\%  & \textbf{70.0\%$\pm$8.4\%} \\%& 0.55 \\
 \cmidrule{2-6}
 & \textbf{Average}            & 46.0\%$\pm$4.1\%  & 46.7\%$\pm$4.1\%  & 49.3\%$\pm$4.1\%  & \textbf{59.3\%$\pm$4.0\%} \\%& 0.55 \\

\midrule
\multirow{6}{2cm}{\textbf{Initial Gripper Position}} 
 & Task 1 $\rightarrow$ Task 1  & 46.7\%$\pm$9.1\%  & 46.7\%$\pm$9.1\%  & \textbf{66.7\%$\pm$8.6\%}  & 60.0\%$\pm$8.9\% \\%& 0.16 \\
 & Task 1 $\rightarrow$ Task 2  & 83.3\%$\pm$6.8\%  & 83.3\%$\pm$6.8\%  & 80.0\%$\pm$7.3\%  & \textbf{100.0\%$\pm$0.0\%} \\%& 0.19 \\
 & Task 1 $\rightarrow$ Task 3  & 46.7\%$\pm$9.1\%  & 53.3\%$\pm$9.1\%  & 56.7\%$\pm$9.0\%  & \textbf{60.0\%$\pm$8.9\%} \\%& 0.15 \\
 & Task 1 $\rightarrow$ Task 7  & 80.0\%$\pm$7.3\%  & 63.3\%$\pm$8.8\%  & 60.0\%$\pm$8.9\%  & \textbf{83.3\%$\pm$6.8\%} \\%& 0.22 \\
 & Task 1 $\rightarrow$ Task 9  & 50.0\%$\pm$9.1\%  & 56.7\%$\pm$9.0\%  & 53.3\%$\pm$9.1\%  & \textbf{60.0\%$\pm$8.9\%} \\%& 0.18 \\
 \cmidrule{2-6}
 & \textbf{Average}            & 61.3\%$\pm$4.0\%  & 60.7\%$\pm$4.0\%  & 63.3\%$\pm$3.9\%  & \textbf{72.7\%$\pm$3.6\%} \\
 
\midrule
\multirow{6}{2cm}{\textbf{Camera Gaussian Noise}} 
 & Task 6 $\rightarrow$ Task 0  & 3.3\%$\pm$3.3\%   & 0.0\%$\pm$0.0\%   & \textbf{43.3\%$\pm$9.0\%}  & 33.3\%$\pm$8.6\% \\%& 0.04 \\
 & Task 6 $\rightarrow$ Task 1  & 53.3\%$\pm$9.1\%  & 0.0\%$\pm$0.0\%   & 70.0\%$\pm$8.4\%  & \textbf{73.3\%$\pm$8.1\%} \\%& 0.04 \\
 & Task 6 $\rightarrow$ Task 4  & 10.0\%$\pm$5.5\%  & 0.0\%$\pm$0.0\%   & \textbf{80.0\%$\pm$7.3\%}  & 53.3\%$\pm$9.1\% \\%& 0.05 \\
 & Task 6 $\rightarrow$ Task 6  & 36.7\%$\pm$8.8\%  & 3.3\%$\pm$3.3\%   & \textbf{76.7\%$\pm$7.7\%}  & 73.3\%$\pm$8.1\% \\%& 0.04 \\
 & Task 6 $\rightarrow$ Task 7  & 30.0\%$\pm$8.4\%  & 0.0\%$\pm$0.0\%   & \textbf{66.7\%$\pm$8.6\%}  & 60.0\%$\pm$8.9\% \\%& 0.02 \\
 \cmidrule{2-6}
 & \textbf{Average}             & 26.7\%$\pm$3.6\%  & 0.7\%$\pm$0.7\%   & \textbf{67.3\%$\pm$3.8\%}  & 58.7\%$\pm$4.0\% \\%& 0.04 \\
 
\bottomrule
\end{tabular}%
% }
\vspace{-15pt}
\end{table}

\subsection{Steering to Improve Robustness}\label{subsec:steering_results}

\begin{wrapfigure}{r}{.50\textwidth}
\vspace{-15pt}
    \includegraphics[width=0.9\linewidth]{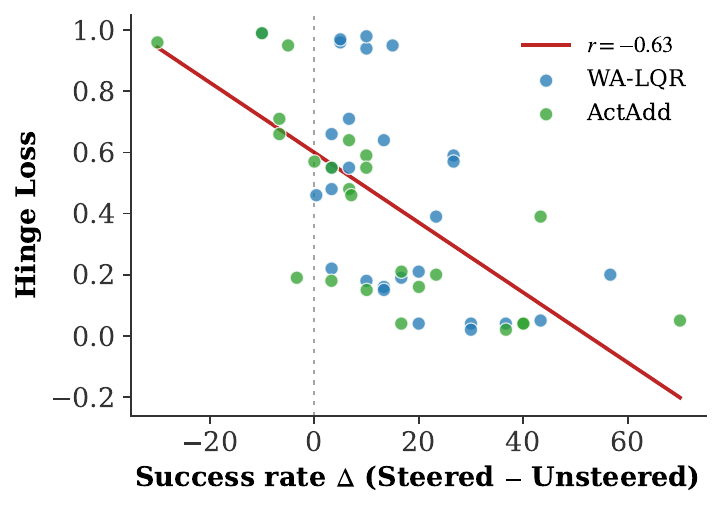}
    \vspace{-5pt}
    \caption{Hinge loss vs. steering performance across tasks and models, with reported line of best fit (red) and correlation coefficient.}% \vspace{-10pt}}
    \vspace{-10pt}
    \label{fig:result_loss_hinge_fit}
\end{wrapfigure}

We leverage the mechanistic insights from Sec.~\ref{sec:mechanistic} to inform steering objectives designed to improve robustness to OOD perturbations in Cosmos-Policy 2B \cite{kim2026cosmos}, DiT4DiT \cite{ma2026dit4dit}, and LingBot-VA \cite{lingbot-va2026}, using WA-LQR (Sec.~\ref{subsec:latentactivationdynamics}) and a variant of simple activation addition adapted to WAMs (Sec. \ref{subsec:activation_addition}) \cite{turner2024activation}. Specifically, we seek to improve success rate under perturbations to initial gripper position, camera orientation, and Gaussian noise corruption in the image data, applied to LIBERO-10 evaluation tasks \cite{liu2023libero}. According to the mechanistic analysis, it is infeasible to construct linear feature directions that apply for all scenes and tasks (see App.~\ref{ap:separability}). Thus, we consider groups of activations which we empirically find have shared representation as suggested by our analysis in Sec. \ref{sec:mechanistic}. For baselines that do not involve finetuning, we compare to the original model under perturbation and a baseline that decomposes the prompt into subtasks, inspired by \cite{Wang_Chen_Liu_Ye_Chen_Lu_Liu_Yu_Jia_2026}.

\begin{wraptable}{r}{0.5\textwidth}
% \vspace{-15pt}
% \begin{minipage}[t]{0.4\linewidth}
\centering
\caption{Success rates on DiT4DiT \cite{ma2026dit4dit}.}\label{tab:dit_results}
\vspace{-5pt}
\scriptsize
\resizebox{\linewidth}{!}{%
\begin{tabular}{lc *{3}{C{1.3cm}}}
\toprule
& Tasks & No Steer. & ActAdd & WA-LQR \\%& Hinge \\
\midrule
\multirow{5}{1.3cm}{\textbf{Initial Gripper Position}} &
T0$\rightarrow$T0 & 53.3\%$\pm$9.1\% & 60.0\%$\pm$8.9\% & \textbf{66.7\%$\pm$8.6\%} \\
& T0$\rightarrow$T1 & 69.6\% & \textbf{76.7\%$\pm$7.7\%} & 70.0\%$\pm$8.4\% \\
& T0$\rightarrow$T2 & 70.0\%$\pm$8.4\% & 63.3\%$\pm$8.8\% & \textbf{73.3\%$\pm$8.1\%} \\
& T0$\rightarrow$T3 & 70.0\%$\pm$8.4\% & 63.3\%$\pm$8.8\% & \textbf{76.7\%$\pm$7.7\%} \\
\cmidrule{2-5}
& \textbf{Avg.} & 65.7\%$\pm$4.3\% & 65.8\%$\pm$4.3\% & \textbf{71.7\%$\pm$4.1\%} \\
\midrule
\multirow{4}{1.3cm}{\textbf{Camera Gaussian Noise}} 
& T1$\rightarrow$T0 & 16.7\%$\pm$6.8\% & 40.0\%$\pm$8.9\% & \textbf{73.3\%$\pm$8.1\%} \\
& T1$\rightarrow$T1 & 20.0\%$\pm$7.3\% & \textbf{63.3\%$\pm$8.8\%} & 43.3\%$\pm$9.0\% \\
& T1$\rightarrow$T2 & 10.0\%$\pm$5.5\% & 26.7\%$\pm$8.1\% & \textbf{30.0\%$\pm$8.4\%} \\
\cmidrule{2-5}
& \textbf{Avg.} & 15.6\%$\pm$3.8\% & 43.3\%$\pm$5.2\% & \textbf{48.9\%$\pm$5.3\%} \\
\bottomrule
\end{tabular}}
\vspace{2 pt}

% --- Table 2: LingBot-VA ---
\centering
\caption{LingBot-VA \cite{lingbot-va2026} success rates.}\label{tab:ling_results}
\vspace{-5pt}
\scriptsize
\resizebox{\linewidth}{!}{%
\begin{tabular}{l c *{3}{C{1.3cm}}}
\toprule
& Tasks & No Steer. & ActAdd & WA-LQR \\%& Hinge\\
\midrule
\multirow{6}{1.4cm}{\textbf{Camera Orientation}} &
T0$\rightarrow$T0 & 65.0\%$\pm$10.7\% & 35.0\%$\pm$10.7\% & \textbf{70.0\%$\pm$10.2\%} \\
& T0$\rightarrow$T2 & \textbf{55.0\%$\pm$11.1\%} & 45.0\%$\pm$11.1\% & 45.0\%$\pm$11.1\% \\
& T0$\rightarrow$T4 & 15.0\%$\pm$8.0\% & 10.0\%$\pm$6.7\% & \textbf{30.0\%$\pm$10.2\%} \\
& T0$\rightarrow$T5 & 70.0\%$\pm$10.2\% & \textbf{75.0\%$\pm$9.7\%} & 60.0\%$\pm$11.0\% \\
& T0$\rightarrow$T9 & 35.0\%$\pm$10.7\% & 40.0\%$\pm$11.0\% & \textbf{50.0\%$\pm$11.2\%} \\
\cmidrule{2-5}
& \textbf{Avg.} & 48.0\%$\pm$5.0\% & 41.0\%$\pm$4.9\% & \textbf{51.0\%$\pm$5.0\%} \\

\midrule

\multirow{6}{1.4cm}{\textbf{Initial Gripper Position}} 
 & T1$\rightarrow$T1  & 65.0\%$\pm$10.7\% & \textbf{75.0\%$\pm$9.7\%} & \textbf{75.0\%$\pm$9.7\%}  \\
 & T1$\rightarrow$T2  & 70.0\%$\pm$10.3\% & 70.0\%$\pm$10.3\% & \textbf{80.0\%$\pm$8.9\%}  \\
 & T1$\rightarrow$T3  & 70.0\%$\pm$10.3\% & \textbf{85.0\%$\pm$8.0\%} & 75.0\%$\pm$9.7\%  \\
 & T1$\rightarrow$T7  & 80.0\%$\pm$8.9\% & 60.0\%$\pm$11.0\% & \textbf{80.0\%$\pm$8.9\%} \\
 & T1$\rightarrow$T9  & 75.0\%$\pm$9.7\% & \textbf{70.0\%$\pm$10.3\%} & 65.0\%$\pm$10.7\%  \\
 \cmidrule{2-5}
 & \textbf{Avg.}             & 72.0\%$\pm$4.5\% & 72.0\%$\pm$4.5\% & \textbf{75.0\%$\pm$4.3\%}  \\
 
\midrule

\multirow{6}{1.4cm}{\textbf{Camera Gaussian Noise}} 
 & T6$\rightarrow$T0  & \textbf{35.0\%$\pm$10.7\%} & 15.0\%$\pm$8.0\% & 20.0\%$\pm$8.9\%  \\
 & T6$\rightarrow$T1  & 90.0\%$\pm$6.7\% & 60.0\%$\pm$11.0\% & \textbf{95.0\%$\pm$4.9\%}  \\
 & T6$\rightarrow$T4  & \textbf{75.0\%$\pm$9.7\%} & 45.0\%$\pm$11.1\% & \textbf{75.0\%$\pm$9.7\%}  \\
 & T6$\rightarrow$T6  & \textbf{85.0\%$\pm$8.0\%} & 55.0\%$\pm$11.1\% & 70.0\%$\pm$10.3\%  \\
 & T6$\rightarrow$T7  & 10.0\%$\pm$6.7\% & 15.0\%$\pm$8.0\% & \textbf{20.0\%$\pm$8.9\%}  \\
 \cmidrule{2-5}
 & \textbf{Avg.}             & \textbf{59.0\%$\pm$4.9\%} & 38.0\%$\pm$4.9\% & 56.0\%$\pm$5.0\% \\

\bottomrule
\end{tabular}}
\vspace{-10pt}

\end{wraptable}

\begin{figure}
\centering
\vspace{-24pt}
    \includegraphics[width=\linewidth]{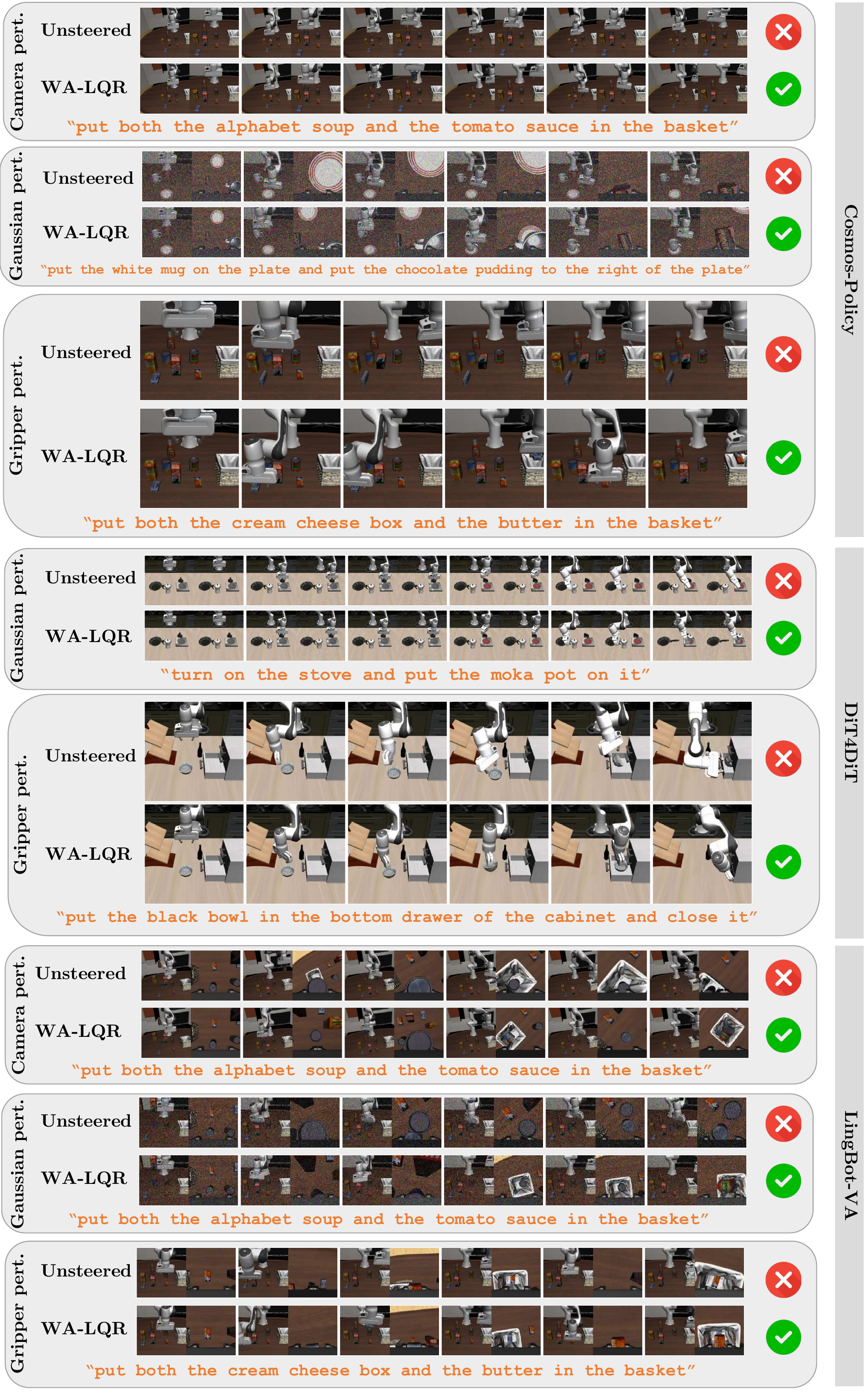}\vspace{-4pt}
    \caption{Snapshots from rollouts where steering enables task success despite unsteered failure, across perturbations, LIBERO-10 tasks, and WAM architectures. Each rollout shows six equally spaced snapshots, with time increasing from left to right. \vspace{-10pt}}
    % \vspace{-10pt}
    \label{fig:qualitative}
\end{figure}

\paragraph{Results} \looseness-1The grouping and steering results for Cosmos-Policy are summarized in Tab.~\ref{tab:cosmos_steering_results}, and we provide qualitative examples of unsteered and steered trajectory rollouts in Fig. \ref{fig:teaser}. We provide a full set of snapshots from qualitative unsteered and steered rollouts in Fig. \ref{fig:qualitative}. WA-LQR is the most effective method in improving robustness to camera orientation and gripper position perturbations, with both steering methods outperforming the unsteered or prompt-steering baselines. As one exception, ActAdd is more effective than WA-LQR on camera Gaussian noise corruption, which we hypothesize results from the strong separability of this feature, enabling simple open-loop control directly in activation space to be viable. We also show in Sec.~\ref{app:actadd_sensitivity} that ActAdd is highly sensitive to the strength parameter $\gamma$, highlighting the benefit of the closed-loop steering provided by WA-LQR, which adaptively modulates the steering magnitude based on alignment with the desired robustness feature direction. On DiT4DiT (Tab.~\ref{tab:dit_results}), ActAdd yields substantial improvements on Gaussian noise but does not change performance on initial gripper position, while WA-LQR outperforms all baselines and ActAdd on Gaussian noise and initial gripper position. Across evaluations, we find that prompt steering is ineffective, either matching unsteered performance or strongly degrading performance.

Task groupings did not emerge on LingBot-VA due to the poor alignment results in Sec.~\ref{sec:mechanistic}, so we instead map the groups discovered for Cosmos-Policy onto LingBot-VA. The results are summarized in Tab.~\ref{tab:ling_results} (more results are in App.~\ref{app:lingbot_results}). Consistent with Sec.~\ref{sec:mechanistic}, we do not observe major robustness improvements from steering for LingBot-VA. In fact, ActAdd often degrades performance due to over-steering, while WA-LQR avoids this because of its closed-loop modulation of steering magnitude. We summarize the relationship between separability and steering performance in Fig.~\ref{fig:result_loss_hinge_fit}, where we observe a negative correlation between hinge loss and robustness improvements through steering, supporting the use of hinge loss as a predictor for steering success. Overall, these results suggest that hinge loss is an effective predictor of steering success, that both open-loop and closed-loop activation steering are effective in improving success rate across models with low separability loss, and that closed-loop steering can be more effective than open-loop perturbations.

\vspace{-12pt}
\section{Discussion, Limitations, and Conclusion}
\vspace{-10pt}

\looseness-1We investigate mechanistic interpretability and activation steering in WAMs. Our analysis reveals clear linear structure in feature-relevant activations under OOD perturbations for Cosmos-Policy \cite{kim2026cosmos} and DiT4DiT \cite{ma2026dit4dit}, but not for LingBot-VA \cite{lingbot-va2026}. We further find that linear separability loss is a strong predictor of steering performance. Motivated by these observations, we adapt activation addition to the WAM setting and introduce WA-LQR, a novel activation steering framework that exploits locally linear dynamics within a feature-relevant activation subspace to improve OOD robustness without finetuning, achieving success-rate improvements of up to 41\%. Overall, our results suggest that mechanistic control of hidden model representations is promising for improving WAM robustness.

\noindent\textbf{Limitations.\ } A key limitation of WA-LQR is its limited applicability across tasks and models. To our knowledge, there is currently no interpretable method for predicting when tasks or environments share transferable representations, requiring mechanistic analysis on a per-setting basis. We also observe strong dependence on model architecture. These results highlight the need to better understand how steerability emerges in robotics foundation models and motivate the design of WAMs that are both steerable and able to preserve representations from their base foundation models.

%===============================================================================

\clearpage

\bibliography{example}  % .bib

@article{wang2026world,
  title={World Action Models: The Next Frontier in Embodied AI},
  author={Wang, Siyin and Shi, Junhao and Fu, Zhaoyang and He, Xinzhe and Liu, Feihong and Yang, Chenchen and Zhou, Yikang and Fei, Zhaoye and Gong, Jingjing and Fu, Jinlan and others},
  journal={arXiv preprint arXiv:2605.12090},
  year={2026}
}

@article{park2023linear,
  title={The linear representation hypothesis and the geometry of large language models},
  author={Park, Kiho and Choe, Yo Joong and Veitch, Victor},
  journal={arXiv preprint arXiv:2311.03658},
  year={2023}
}

@article{turner2024activation,
  title={Activation addition: Steering language models without optimization},
  author={Turner, Alexander Matt and Thiergart, Lisa and Leech, Gavin and Udell, David and Mini, Ulisse and MacDiarmid, Monte},
  year={2024}
}

@inproceedings{rimsky2024steering,
  title={Steering llama 2 via contrastive activation addition},
  author={Rimsky, Nina and Gabrieli, Nick and Schulz, Julian and Tong, Meg and Hubinger, Evan and Turner, Alexander},
  booktitle={Proceedings of the 62nd Annual Meeting of the Association for Computational Linguistics (Volume 1: Long Papers)},
  pages={15504--15522},
  year={2024}
}

@article{arditi2024refusal,
  title={Refusal in language models is mediated by a single direction},
  author={Arditi, Andy and Obeso, Oscar and Syed, Aaquib and Paleka, Daniel and Panickssery, Nina and Gurnee, Wes and Nanda, Neel},
  journal={Advances in Neural Information Processing Systems},
  volume={37},
  pages={136037--136083},
  year={2024}
}

@inproceedings{hong2026activation,
  title={Activation Steering of Video Generation Models via Reduced-Order Linear Optimal Control},
  author={Hong, Jihoon and Chan, Alice and Dai, Qiyue and Skifstad, Julian and Chou, Glen},
  year={2026}
}

@inproceedings{rodriguez2025controlling,
  title={Controlling language and diffusion models by transporting activations},
  author={Rodriguez, Pau and Blaas, Arno and Klein, Michal and Zappella, Luca and Apostoloff, Nicholas and Suau, Xavier and others},
  booktitle={International Conference on Learning Representations},
  volume={2025},
  pages={89812--89855},
  year={2025}
}

@article{skifstad2026local,
  title={Local Linearity of LLMs Enables Activation Steering via Model-Based Linear Optimal Control},
  author={Skifstad, Julian and Yang, Xinyue Annie and Chou, Glen},
  journal={arXiv preprint arXiv:2604.19018},
  year={2026}
}

@article{hou2026world,
  title={World Model for Robot Learning: A Comprehensive Survey},
  author={Hou, Bohan and Li, Gen and Jia, Jindou and An, Tuo and Guo, Xinying and Leng, Sicong and Geng, Haoran and Ze, Yanjie and Harada, Tatsuya and Torr, Philip and others},
  journal={arXiv preprint arXiv:2605.00080},
  year={2026}
}

@article{black2024pi0,
  title={$\pi_0$: A Vision-Language-Action Flow Model for General Robot Control},
  author={Black, Kevin and Brown, Noah and Driess, Danny and Esmail, Adnan and Equi, Michael and Finn, Chelsea and Fusai, Niccolo and Groom, Lachy and Hausman, Karol and Ichter, Brian and others},
  journal={arXiv preprint arXiv:2410.24164},
  year={2024}
}

@article{kim2024openvla,
  title={Openvla: An open-source vision-language-action model},
  author={Kim, Moo Jin and Pertsch, Karl and Karamcheti, Siddharth and Xiao, Ted and Balakrishna, Ashwin and Nair, Suraj and Rafailov, Rafael and Foster, Ethan and Lam, Grace and Sanketi, Pannag and others},
  journal={arXiv preprint arXiv:2406.09246},
  year={2024}
}

@inproceedings{zitkovich2023rt,
  title={Rt-2: Vision-language-action models transfer web knowledge to robotic control},
  author={Zitkovich, Brianna and Yu, Tianhe and Xu, Sichun and Xu, Peng and Xiao, Ted and Xia, Fei and Wu, Jialin and Wohlhart, Paul and Welker, Stefan and Wahid, Ayzaan and others},
  booktitle={Conference on Robot Learning},
  pages={2165--2183},
  year={2023},
  organization={PMLR}
}

@article{kim2026cosmos,
  title={Cosmos policy: Fine-tuning video models for visuomotor control and planning},
  author={Kim, Moo Jin and Gao, Yihuai and Lin, Tsung-Yi and Lin, Yen-Chen and Ge, Yunhao and Lam, Grace and Liang, Percy and Song, Shuran and Liu, Ming-Yu and Finn, Chelsea and others},
  journal={arXiv preprint arXiv:2601.16163},
  year={2026}
}

@article{ye2026world,
  title={World action models are zero-shot policies},
  author={Ye, Seonghyeon and Ge, Yunhao and Zheng, Kaiyuan and Gao, Shenyuan and Yu, Sihyun and Kurian, George and Indupuru, Suneel and Tan, You Liang and Zhu, Chuning and Xiang, Jiannan and others},
  journal={arXiv preprint arXiv:2602.15922},
  year={2026}
}

@article{wen2024vidman,
  title={Vidman: Exploiting implicit dynamics from video diffusion model for effective robot manipulation},
  author={Wen, Youpeng and Lin, Junfan and Zhu, Yi and Han, Jianhua and Xu, Hang and Zhao, Shen and Liang, Xiaodan},
  journal={Advances in Neural Information Processing Systems},
  volume={37},
  pages={41051--41075},
  year={2024}
}

@article{jia2025video2act,
  title={Video2Act: A Dual-System Video Diffusion Policy with Robotic Spatio-Motional Modeling},
  author={Jia, Yueru and Liu, Jiaming and Liu, Shengbang and Zhou, Rui and Yu, Wanhe and Yan, Yuyang and Chi, Xiaowei and Guo, Yandong and Shi, Boxin and Zhang, Shanghang},
  journal={arXiv preprint arXiv:2512.03044},
  year={2025}
}

@article{hu2024video,
  title={Video prediction policy: A generalist robot policy with predictive visual representations},
  author={Hu, Yucheng and Guo, Yanjiang and Wang, Pengchao and Chen, Xiaoyu and Wang, Yen-Jen and Zhang, Jianke and Sreenath, Koushil and Lu, Chaochao and Chen, Jianyu},
  journal={arXiv preprint arXiv:2412.14803},
  year={2024}
}

@article{brohan2022rt,
  title={Rt-1: Robotics transformer for real-world control at scale},
  author={Brohan, Anthony and Brown, Noah and Carbajal, Justice and Chebotar, Yevgen and Dabis, Joseph and Finn, Chelsea and Gopalakrishnan, Keerthana and Hausman, Karol and Herzog, Alex and Hsu, Jasmine and others},
  journal={arXiv preprint arXiv:2212.06817},
  year={2022}
}

@article{buurmeijer2026observing,
  title={Observing and Controlling Features in Vision-Language-Action Models},
  author={Buurmeijer, Hugo and Alonso, Carmen Amo and Swann, Aiden and Pavone, Marco},
  journal={arXiv preprint arXiv:2603.05487},
  year={2026}
}

@article{mitra2025mechanistic,
  title={Mechanistic Finetuning of Vision-Language-Action Models via Few-Shot Demonstrations},
  author={Mitra, Chancharik and Luo, Yusen and Saravanan, Raj and Niu, Dantong and Pai, Anirudh and Thomason, Jesse and Darrell, Trevor and Anwar, Abrar and Ramanan, Deva and Herzig, Roei},
  journal={arXiv preprint arXiv:2511.22697},
  year={2025}
}

@article{zhang2026embodied,
  title={Embodied Interpretability: Linking Causal Understanding to Generalization in Vision-Language-Action Models},
  author={Zhang, Hanxin and Xu, Mingshuo and Dhafer, Abdulqader and Yue, Shigang and Dong, Hongbiao and Hao, Zhou Daniel},
  journal={arXiv preprint arXiv:2605.00321},
  year={2026}
}

@article{pertsch2025fast,
  title={Fast: Efficient action tokenization for vision-language-action models},
  author={Pertsch, Karl and Stachowicz, Kyle and Ichter, Brian and Driess, Danny and Nair, Suraj and Vuong, Quan and Mees, Oier and Finn, Chelsea and Levine, Sergey},
  journal={arXiv preprint arXiv:2501.09747},
  year={2025}
}

@article{li2026causal,
  title={Causal World Modeling for Robot Control},
  author={Li, Lin and Zhang, Qihang and Luo, Yiming and Yang, Shuai and Wang, Ruilin and Han, Fei and Yu, Mingrui and Gao, Zelin and Xue, Nan and Zhu, Xing and others},
  journal={arXiv preprint arXiv:2601.21998},
  year={2026}
}

@article{fei2025libero,
  title={Libero-plus: In-depth robustness analysis of vision-language-action models},
  author={Fei, Senyu and Wang, Siyin and Shi, Junhao and Dai, Zihao and Cai, Jikun and Qian, Pengfang and Ji, Li and He, Xinzhe and Zhang, Shiduo and Fei, Zhaoye and others},
  journal={arXiv preprint arXiv:2510.13626},
  year={2025}
}

@article{yuan2026fast,
  title={Fast-WAM: Do World Action Models Need Test-time Future Imagination?},
  author={Yuan, Tianyuan and Dong, Zibin and Liu, Yicheng and Zhao, Hang},
  journal={arXiv preprint arXiv:2603.16666},
  year={2026}
}

@article{pumacay2024colosseum,
  title={The colosseum: A benchmark for evaluating generalization for robotic manipulation},
  author={Pumacay, Wilbert and Singh, Ishika and Duan, Jiafei and Krishna, Ranjay and Thomason, Jesse and Fox, Dieter},
  journal={arXiv preprint arXiv:2402.08191},
  year={2024}
}

@inproceedings{guo2025improving,
  title={Improving vision-language-action model with online reinforcement learning},
  author={Guo, Yanjiang and Zhang, Jianke and Chen, Xiaoyu and Ji, Xiang and Wang, Yen-Jen and Hu, Yucheng and Chen, Jianyu},
  booktitle={2025 IEEE International Conference on Robotics and Automation (ICRA)},
  pages={15665--15672},
  year={2025},
  organization={IEEE}
}

@article{tan2025interactive,
  title={Interactive post-training for vision-language-action models},
  author={Tan, Shuhan and Dou, Kairan and Zhao, Yue and Kr{\"a}henb{\"u}hl, Philipp},
  journal={arXiv preprint arXiv:2505.17016},
  year={2025}
}

@article{liu2026can,
  title={What can rl bring to vla generalization? an empirical study},
  author={Liu, Jijia and Gao, Feng and Wei, Bingwen and Chen, Xinlei and Liao, Qingmin and Wu, Yi and Yu, Chao and Wang, Yu},
  journal={Advances in Neural Information Processing Systems},
  volume={38},
  pages={97121--97151},
  year={2026}
}

@misc{zhou2025exploringlimitsvisionlanguageactionmanipulations,
      title={Exploring the Limits of Vision-Language-Action Manipulations in Cross-task Generalization}, 
      author={Jiaming Zhou and Ke Ye and Jiayi Liu and Teli Ma and Zifan Wang and Ronghe Qiu and Kun-Yu Lin and Zhilin Zhao and Junwei Liang},
      year={2025},
      eprint={2505.15660},
      archivePrefix={arXiv},
      primaryClass={cs.RO},
      url={https://arxiv.org/abs/2505.15660}, 
}

@article{kim2025fine,
  title={Fine-tuning vision-language-action models: Optimizing speed and success},
  author={Kim, Moo Jin and Finn, Chelsea and Liang, Percy},
  journal={arXiv preprint arXiv:2502.19645},
  year={2025}
}

@article{du2023learning,
  title={Learning universal policies via text-guided video generation},
  author={Du, Yilun and Yang, Sherry and Dai, Bo and Dai, Hanjun and Nachum, Ofir and Tenenbaum, Josh and Schuurmans, Dale and Abbeel, Pieter},
  journal={Advances in neural information processing systems},
  volume={36},
  pages={9156--9172},
  year={2023}
}

@article{wang2025vlatest,
  title={Vlatest: Testing and evaluating vision-language-action models for robotic manipulation},
  author={Wang, Zhijie and Zhou, Zhehua and Song, Jiayang and Huang, Yuheng and Shu, Zhan and Ma, Lei},
  journal={Proceedings of the ACM on Software Engineering},
  volume={2},
  number={FSE},
  pages={1615--1638},
  year={2025},
  publisher={ACM New York, NY, USA}
}

@article{li2024evaluating,
  title={Evaluating real-world robot manipulation policies in simulation},
  author={Li, Xuanlin and Hsu, Kyle and Gu, Jiayuan and Pertsch, Karl and Mees, Oier and Walke, Homer Rich and Fu, Chuyuan and Lunawat, Ishikaa and Sieh, Isabel and Kirmani, Sean and others},
  journal={arXiv preprint arXiv:2405.05941},
  year={2024}
}

@article{team2024octo,
  title={Octo: An open-source generalist robot policy},
  author={Team, Octo Model and Ghosh, Dibya and Walke, Homer and Pertsch, Karl and Black, Kevin and Mees, Oier and Dasari, Sudeep and Hejna, Joey and Kreiman, Tobias and Xu, Charles and others},
  journal={arXiv preprint arXiv:2405.12213},
  year={2024}
}

@inproceedings{o2024open,
  title={Open x-embodiment: Robotic learning datasets and rt-x models: Open x-embodiment collaboration 0},
  author={O’Neill, Abby and Rehman, Abdul and Maddukuri, Abhiram and Gupta, Abhishek and Padalkar, Abhishek and Lee, Abraham and Pooley, Acorn and Gupta, Agrim and Mandlekar, Ajay and Jain, Ajinkya and others},
  booktitle={2024 IEEE International Conference on Robotics and Automation (ICRA)},
  pages={6892--6903},
  year={2024},
  organization={IEEE}
}

@article{ye2026gigaworld,
  title={GigaWorld-Policy: An Efficient Action-Centered World--Action Model},
  author={Ye, Angen and Wang, Boyuan and Ni, Chaojun and Huang, Guan and Zhao, Guosheng and Li, Hao and Li, Hengtao and Li, Jie and Lv, Jindi and Liu, Jingyu and others},
  journal={arXiv preprint arXiv:2603.17240},
  year={2026}
}

@article{zhang2026world,
  title={Do World Action Models Generalize Better than VLAs? A Robustness Study},
  author={Zhang, Zhanguang and Li, Zhiyuan and Rahmati, Behnam and Yang, Rui Heng and Ma, Yintao and Rasouli, Amir and Pakdamansavoji, Sajjad and Wu, Yangzheng and Zhang, Lingfeng and Cao, Tongtong and others},
  journal={arXiv preprint arXiv:2603.22078},
  year={2026}
}

@article{Huang_Zhang_Azarcon_Chou_Kira_2025, title={MAPS: Preserving Vision-Language Representations via Module-Wise Proximity Scheduling for Better Vision-Language-Action Generalization}, url={http://arxiv.org/abs/2511.19878}, DOI={10.48550/arXiv.2511.19878}, journal={arXiv preprint arXiv:2511.19878}, author={Huang, Chengyue and Zhang, Mellon M. and Azarcon, Robert and Chou, Glen and Kira, Zsolt}, year={2025}, month=nov }

@article{marks2024geometry, title={The Geometry of Truth: Emergent Linear Structure in Large Language Model Representations of True/False Datasets}, url={http://arxiv.org/abs/2310.06824}, DOI={10.48550/arXiv.2310.06824}, journal={arXiv preprint arXiv:2310.06824}, publisher={arXiv}, author={Marks, Samuel and Tegmark, Max}, year={2024}, month=aug }

@InProceedings{haon25mechanistic,
  title = 	 {Mechanistic Interpretability for Steering Vision-Language-Action Models},
  author =       {H\"{a}on, Bear and Stocking, Kaylene Caswell and Chuang, Ian and Tomlin, Claire},
  booktitle = 	 {Proceedings of The 9th Conference on Robot Learning},
  pages = 	 {2743--2762},
  year = 	 {2025},
  editor = 	 {Lim, Joseph and Song, Shuran and Park, Hae-Won},
  volume = 	 {305},
  series = 	 {Proceedings of Machine Learning Research},
  month = 	 {27--30 Sep},
  publisher =    {PMLR},
  url = 	 {https://proceedings.mlr.press/v305/haon25a.html},
}

@inbook{kecman2005svm, address={Berlin, Heidelberg}, title={Support Vector Machines – An Introduction}, ISBN={9783540323846}, url={https://doi.org/10.1007/10984697_1}, DOI={10.1007/10984697_1}, booktitle={Support Vector Machines: Theory and Applications}, publisher={Springer}, author={Kecman, V.}, editor={Wang, Lipo}, year={2005}, pages={1–47}, language={en} }

@article{liu2023libero,
  title={Libero: Benchmarking knowledge transfer for lifelong robot learning},
  author={Liu, Bo and Zhu, Yifeng and Gao, Chongkai and Feng, Yihao and Liu, Qiang and Zhu, Yuke and Stone, Peter},
  journal={Advances in Neural Information Processing Systems},
  volume={36},
  pages={44776--44791},
  year={2023}
}

@article{halko2011finding,
  title={Finding structure with randomness: Probabilistic algorithms for constructing approximate matrix decompositions},
  author={Halko, Nathan and Martinsson, Per-Gunnar and Tropp, Joel A},
  journal={SIAM review},
  volume={53},
  number={2},
  pages={217--288},
  year={2011},
  publisher={SIAM}
}

@article{facchiano2025video,
  title={Video unlearning via low-rank refusal vector},
  author={Facchiano, Simone and Saravalle, Stefano and Migliarini, Matteo and De Matteis, Edoardo and Sampieri, Alessio and Pilzer, Andrea and Rodol{\`a}, Emanuele and Spinelli, Indro and Franco, Luca and Galasso, Fabio},
  journal={arXiv preprint arXiv:2506.07891},
  year={2025}
}

@article{ekin2026unreasonable,
  title={The Unreasonable Effectiveness of Text Embedding Interpolation for Continuous Image Steering},
  author={Ekin, Yigit and Gandelsman, Yossi},
  journal={arXiv preprint arXiv:2603.17998},
  year={2026}
}

@article{Cheng_Alonso_2025, title={Linearly Controlled Language Generation with Performative Guarantees}, url={http://arxiv.org/abs/2405.15454}, DOI={10.48550/arXiv.2405.15454}, abstractNote={The increasing prevalence of Large Language Models (LMs) in critical applications highlights the need for controlled language generation strategies that are not only computationally efficient but that also enjoy performance guarantees. To achieve this, we use a common model of concept semantics as linearly represented in an LM’s latent space. In particular, we take the view that natural language generation traces a trajectory in this continuous semantic space, realized by the language model’s hidden activations. This view permits a control-theoretic treatment of text generation in latent space, in which we propose a lightweight, gradient-free intervention that dynamically steers trajectories away from regions corresponding to undesired meanings. In particular, we propose to directly intervene the activations of the token that is being generated in embedding space in an online fashion. Crucially, we do not simply steer activations towards a desirable region. Instead, our method relies on classical techniques from control theory to precisely control activations in a context-dependent way, and guarantees that they are brought into a specific pre-defined region of embedding space that corresponds to allowed semantics. Our intervention is computed in closed-form according to an optimal controller formulation, minimally impacting generation time. This control of the activations in embedding space allows for fine-grained steering of attributes of the generated sequence. We demonstrate the effectiveness of our approach on different objectives -- toxicity avoidance and sentiment control -- while maintaining text quality.}, note={arXiv:2405.15454}, number={arXiv:2405.15454}, publisher={arXiv}, author={Cheng, Emily and Alonso, Carmen Amo}, year={2025}, month=sep }

@article{Nguyen_Vu_Pham_Zhang_Nguyen_2025, title={Activation Steering with a Feedback Controller}, url={http://arxiv.org/abs/2510.04309}, DOI={10.48550/arXiv.2510.04309}, abstractNote={Controlling the behaviors of large language models (LLM) is fundamental to their safety alignment and reliable deployment. However, existing steering methods are primarily driven by empirical insights and lack theoretical performance guarantees. In this work, we develop a control-theoretic foundation for activation steering by showing that popular steering methods correspond to the proportional (P) controllers, with the steering vector serving as the feedback signal. Building on this finding, we propose Proportional-Integral-Derivative (PID) Steering, a principled framework that leverages the full PID controller for activation steering in LLMs. The proportional (P) term aligns activations with target semantic directions, the integral (I) term accumulates errors to enforce persistent corrections across layers, and the derivative (D) term mitigates overshoot by counteracting rapid activation changes. This closed-loop design yields interpretable error dynamics and connects activation steering to classical stability guarantees in control theory. Moreover, PID Steering is lightweight, modular, and readily integrates with state-of-the-art steering methods. Extensive experiments across multiple LLM families and benchmarks demonstrate that PID Steering consistently outperforms existing approaches, achieving more robust and reliable behavioral control.}, note={arXiv:2510.04309}, number={arXiv:2510.04309}, publisher={arXiv}, author={Nguyen, Dung V. and Vu, Hieu M. and Pham, Nhi Y. and Zhang, Lei and Nguyen, Tan M.}, year={2025}, month=oct }

@article{Vu_Nguyen_2025, title={Angular Steering: Behavior Control via Rotation in Activation Space}, url={http://arxiv.org/abs/2510.26243}, DOI={10.48550/arXiv.2510.26243}, abstractNote={Controlling specific behaviors in large language models while preserving their general capabilities is a central challenge for safe and reliable artificial intelligence deployment. Current steering methods, such as vector addition and directional ablation, are constrained within a two-dimensional subspace defined by the activation and feature direction, making them sensitive to chosen parameters and potentially affecting unrelated features due to unintended interactions in activation space. We introduce Angular Steering, a novel and flexible method for behavior modulation that operates by rotating activations within a fixed two-dimensional subspace. By formulating steering as a geometric rotation toward or away from a target behavior direction, Angular Steering provides continuous, fine-grained control over behaviors such as refusal and compliance. We demonstrate this method using refusal steering emotion steering as use cases. Additionally, we propose Adaptive Angular Steering, a selective variant that rotates only activations aligned with the target feature, further enhancing stability and coherence. Angular Steering generalizes existing addition and orthogonalization techniques under a unified geometric rotation framework, simplifying parameter selection and maintaining model stability across a broader range of adjustments. Experiments across multiple model families and sizes show that Angular Steering achieves robust behavioral control while maintaining general language modeling performance, underscoring its flexibility, generalization, and robustness compared to prior approaches. Code and artifacts are available at https://github.com/lone17/angular-steering/.}, note={arXiv:2510.26243}, number={arXiv:2510.26243}, publisher={arXiv}, author={Vu, Hieu M. and Nguyen, Tan M.}, year={2025}, month=oct }

@article{fang2026safe,
  title={Safe Large-Scale Robust Nonlinear MPC in Milliseconds via Reachability-Constrained System Level Synthesis on the GPU},
  author={Fang, Jeffrey and Chou, Glen},
  journal={arXiv preprint arXiv:2604.07644},
  year={2026}
}

@article{Elhage_Hume_Olsson_2022, title={Toy Models of Superposition}, url={http://arxiv.org/abs/2209.10652}, DOI={10.48550/arXiv.2209.10652}, abstractNote={Neural networks often pack many unrelated concepts into a single neuron - a puzzling phenomenon known as “polysemanticity” which makes interpretability much more challenging. This paper provides a toy model where polysemanticity can be fully understood, arising as a result of models storing additional sparse features in ‘superposition.’ We demonstrate the existence of a phase change, a surprising connection to the geometry of uniform polytopes, and evidence of a link to adversarial examples. We also discuss potential implications for mechanistic interpretability.}, note={arXiv:2209.10652}, number={arXiv:2209.10652}, publisher={arXiv}, author={Elhage, Nelson and Hume, Tristan and Olsson, Catherine and Schiefer, Nicholas and Henighan, Tom and Kravec, Shauna and Hatfield-Dodds, Zac and Lasenby, Robert and Drain, Dawn and Chen, Carol and Grosse, Roger and McCandlish, Sam and Kaplan, Jared and Amodei, Dario and Wattenberg, Martin and Olah, Christopher}, year={2022}, month=sept }

@article{Kong_Wang_Mu_Du_Zhuang_Zhou_Song_Zhang_Wang_Zhang_2024, title={Aligning Large Language Models with Representation Editing: A Control Perspective}, url={http://arxiv.org/abs/2406.05954}, DOI={10.48550/arXiv.2406.05954}, abstractNote={Aligning large language models (LLMs) with human objectives is crucial for real-world applications. However, fine-tuning LLMs for alignment often suffers from unstable training and requires substantial computing resources. Test-time alignment techniques, such as prompting and guided decoding, do not modify the underlying model, and their performance remains dependent on the original model’s capabilities. To address these challenges, we propose aligning LLMs through representation editing. The core of our method is to view a pre-trained autoregressive LLM as a discrete-time stochastic dynamical system. To achieve alignment for specific objectives, we introduce external control signals into the state space of this language dynamical system. We train a value function directly on the hidden states according to the Bellman equation, enabling gradient-based optimization to obtain the optimal control signals at test time. Our experiments demonstrate that our method outperforms existing test-time alignment techniques while requiring significantly fewer resources compared to fine-tuning methods. Our code is available at https://github.com/Lingkai-Kong/RE-Control.}, note={arXiv:2406.05954}, number={arXiv:2406.05954}, publisher={arXiv}, author={Kong, Lingkai and Wang, Haorui and Mu, Wenhao and Du, Yuanqi and Zhuang, Yuchen and Zhou, Yifei and Song, Yue and Zhang, Rongzhi and Wang, Kai and Zhang, Chao}, year={2024}, month=nov }

@article{
bereska2024mechanistic,
title={Mechanistic Interpretability for {AI} Safety - A Review},
author={Leonard Bereska and Stratis Gavves},
journal={Transactions on Machine Learning Research},
issn={2835-8856},
year={2024},
url={https://openreview.net/forum?id=ePUVetPKu6},
note={Survey Certification, Expert Certification}
}

@article{zou2023representation,
  title={Representation engineering: A top-down approach to ai transparency},
  author={Zou, Andy and Phan, Long and Chen, Sarah and Campbell, James and Guo, Phillip and Ren, Richard and Pan, Alexander and Yin, Xuwang and Mazeika, Mantas and Dombrowski, Ann-Kathrin and others},
  journal={arXiv preprint arXiv:2310.01405},
  year={2023}
}

@article{bhargava2023s,
  title={What's the magic word? a control theory of llm prompting},
  author={Bhargava, Aman and Witkowski, Cameron and Looi, Shi-Zhuo and Thomson, Matt},
  journal={arXiv preprint arXiv:2310.04444},
  year={2023}
}

@inproceedings{
Dathathri2020Plug,
title={Plug and Play Language Models: A Simple Approach to Controlled Text Generation},
author={Sumanth Dathathri and Andrea Madotto and Janice Lan and Jane Hung and Eric Frank and Piero Molino and Jason Yosinski and Rosanne Liu},
booktitle={International Conference on Learning Representations},
year={2020},
url={https://openreview.net/forum?id=H1edEyBKDS}
}

@article{li2023inference,
  title={Inference-time intervention: Eliciting truthful answers from a language model},
  author={Li, Kenneth and Patel, Oam and Vi{\'e}gas, Fernanda and Pfister, Hanspeter and Wattenberg, Martin},
  journal={Advances in Neural Information Processing Systems},
  volume={36},
  pages={41451--41530},
  year={2023}
}

@article{wu2024reft,
  title={Reft: Representation finetuning for language models},
  author={Wu, Zhengxuan and Arora, Aryaman and Wang, Zheng and Geiger, Atticus and Jurafsky, Dan and Manning, Christopher D and Potts, Christopher},
  journal={Advances in Neural Information Processing Systems},
  volume={37},
  pages={63908--63962},
  year={2024}
}

@inproceedings{
lee2025programming,
title={Programming Refusal with Conditional Activation Steering},
author={Bruce W. Lee and Inkit Padhi and Karthikeyan Natesan Ramamurthy and Erik Miehling and Pierre Dognin and Manish Nagireddy and Amit Dhurandhar},
booktitle={The Thirteenth International Conference on Learning Representations},
year={2025},
url={https://openreview.net/forum?id=Oi47wc10sm}
}

@inproceedings{
marks2024the,
title={The Geometry of Truth: Emergent Linear Structure in Large Language Model Representations of True/False Datasets},
author={Samuel Marks and Max Tegmark},
booktitle={First Conference on Language Modeling},
year={2024},
url={https://openreview.net/forum?id=aajyHYjjsk}
}

@book{rawlings2020model,
  title={Model predictive control: theory, computation, and design},
  author={Rawlings, James Blake and Mayne, David Q and Diehl, Moritz and others},
  volume={2},
  year={2020},
  publisher={Nob Hill Publishing Madison, WI}
}

@article{rodriguez2025lineas,
  title={LinEAS: End-to-end Learning of Activation Steering with a Distributional Loss},
  author={Rodriguez, Pau and Klein, Michal and Gualdoni, Eleonora and Maiorca, Valentino and Blaas, Arno and Zappella, Luca and Cuturi, Marco and Suau, Xavier},
  journal={arXiv preprint arXiv:2503.10679},
  year={2025}
}

@article{grant2026not,
  title={Not All Features Are Created Equal: A Mechanistic Study of Vision-Language-Action Models},
  author={Grant, Bryce and Zhao, Xijia and Wang, Peng},
  journal={arXiv preprint arXiv:2603.19233},
  year={2026}
}

@inproceedings{peebles2023scalable,
  title={Scalable diffusion models with transformers},
  author={Peebles, William and Xie, Saining},
  booktitle={Proceedings of the IEEE/CVF international conference on computer vision},
  pages={4195--4205},
  year={2023}
}

@inproceedings{khan2025controlling,
  title={Controlling Vision--Language--Action Policies through Sparse Latent Directions},
  author={Khan, Momin Ahmad and Boskov, Novak and Anwar, Fatima M and Khan, Manzoor A},
  booktitle={Mechanistic Interpretability Workshop at NeurIPS 2025}
}

@book{hespanha2018linear,
  title={Linear systems theory},
  author={Hespanha, Joao P},
  year={2018},
  publisher={Princeton university press}
}

@article{miao2026coast,
  title={Contrastive Conceptor Activation Steering (COAST): Unlocking Vision-Language-Action Models through Hidden States},
  author={Miao, Miranda Muqing and Kim, Subin and Yang, Brandon and Ungar, Lyle},
  journal={arXiv preprint arXiv:2605.17144},
  year={2026}
}

@article{swann2026sparse,
  title={Sparse autoencoders reveal interpretable and steerable features in vla models},
  author={Swann, Aiden and McGranahan, Lachlain and Buurmeijer, Hugo and Kennedy III, Monroe and Schwager, Mac},
  journal={arXiv preprint arXiv:2603.19183},
  year={2026}
}

@article{sharkey2025open,
  title={Open problems in mechanistic interpretability},
  author={Sharkey, Lee and Chughtai, Bilal and Batson, Joshua and Lindsey, Jack and Wu, Jeff and Bushnaq, Lucius and Goldowsky-Dill, Nicholas and Heimersheim, Stefan and Ortega, Alejandro and Bloom, Joseph and others},
  journal={arXiv preprint arXiv:2501.16496},
  year={2025}
}

@article{lingbot-va2026,
  title={Causal World Modeling for Robot Control},
  author={Li, Lin and Zhang, Qihang and Luo, Yiming and Yang, Shuai and Wang, Ruilin and Han, Fei and Yu, Mingrui and Gao, Zelin and Xue, Nan and Zhu, Xing and Shen, Yujun and Xu, Yinghao},
  journal={arXiv preprint arXiv:2601.21998},
  year={2026}
}

@article{Wang_Chen_Liu_Ye_Chen_Lu_Liu_Yu_Jia_2026, title={VP-VLA: Visual Prompting as an Interface for Vision-Language-Action Models}, url={http://arxiv.org/abs/2603.22003}, DOI={10.48550/arXiv.2603.22003}, note={arXiv:2603.22003}, journal={arXiv preprint arXiv:2603.22003}, publisher={arXiv}, author={Wang, Zixuan and Chen, Yuxin and Liu, Yuqi and Ye, Jinhui and Chen, Pengguang and Lu, Changsheng and Liu, Shu and Yu, Bei and Jia, Jiaya}, year={2026}, month=may }

@article{ma2026dit4dit,
  title={DiT4DiT: Jointly Modeling Video Dynamics and Actions for Generalizable Robot Control},
  author={Ma, Teli and Zheng, Jia and Wang, Zifan and Jiang, Chunli and Cui, Andy and Liang, Junwei and Yang, Shuo},
  journal={arXiv preprint arXiv:2603.10448},
  year={2026}
}

\clearpage
\appendix

\begin{center}
    \Large \textbf{Appendices}
\end{center}

In the following, we provide an overview of our appendices. In App. \ref{app:contrastive}, we describe how contrastive vectors are constructed for Gaussian noise, camera perturbations, and gripper-position perturbations, including the definitions of the desirable and undesirable activation sets used for steering. In App. \ref{app:lingbot_results}, we provide supplemental experimental results for LingBot-VA, including additional WA-LQR evaluations on camera orientation, initial gripper-position, and Gaussian-noise perturbations in both the action and video modules. In App. \ref{app:actadd_sensitivity}, we provide a parameter sensitivity analysis on ActAdd. 
% In App. \ref{app:dit4dit_results}, we provide supplemental experimental results on another WAM, DiT4DiT, including a mechanistic analysis of feature separability and an evaluation of WA-LQR for improving robustness under Gaussian-noise and gripper-position perturbations. 
In App. \ref{app:experimental_details}, we provide additional experimental details for the mechanistic study and for the model evaluations, including activation collection, dimensionality reduction, SVM fitting, and LQR steering implementation details. In App. \ref{ap:separability}, we provide the complete low-dimensional separability results across all LIBERO-10 tasks for all models, including both task-level and pairwise separability plots. In App. \ref{app:full_model_separation}, we provide full model separation plots across layers for Cosmos-Policy and DiT4DiT.

\section{Contrastive Vectors}\label{app:contrastive}

To construct contrastive vectors we define a \textit{desirable} or target set of inputs $\mathcal D_+$, with rollouts emblematic of behavior we seek to induce via steering, and \textit{undesirable} set of inputs $\mathcal D_-$, which are perturbed by some nuisance and result in failed rollouts. The definition of this set is consistent between models, for each perturbation. 

For Gaussian noise, we consider noised and clean camera images, i.e., $\mathcal D_+=\{\texttt{Clean inputs}\}$ and $\mathcal{D}_- = \{\texttt{Noised inputs}\}$. For pairs of inputs in $\xi^+ \in \mathcal D_+$ and $\xi^- \in \mathcal{D}_-$, we perform a model forward pass on each input and collect their corresponding activations $x^+,x^-$. 

Similarly, the camera perturbation contrastive input sets are defined as $\mathcal{D}_+ = \{\texttt{No perturbation}\}$ and $\mathcal{D}_- = \{\texttt{Perturbed camera view}\}$. The corresponding activations $x^+,x^-$ are collected from a forward pass of the model. 

% \[
% e_k = x^{+}_k - x^-_k
% \]
% \[\{e_k\}_k\]

For gripper position perturbation, the definition of $\mathcal D_+$ and $\mathcal D_-$ differs slightly. Rather than defining the sets as unperturbed and perturbed, respectively, we let $\mathcal D_+$ be inputs corresponding to successful rollouts and $\mathcal D_-$ be inputs corresponding to unsuccessful inputs, both under gripper perturbation. This is to account for an observed variable sensitivity to different perturbations, i.e., directly including perturbed inputs in the negative dataset would result in many successful rollouts in the negative activations, resulting intuitively in ``steering away'' from desired behavior. Note, however, that we do not observe a meaningful difference in the mechanistic analysis when making the distinction between successful and unsuccessful rollouts in our dataset configuration (see App.~\ref{app:subsec:mechanistic_details}).
% This is to account for the fact that there is variability in the nominal initial position in the original LIBERO dataset. 

Given sets of contrastive vectors $\{x^+_k\}$ and $\{x^-_k\}$, we compute the contrastive direction simply as the difference between pairs of positive and negative activations, with different pooling and processing as described in Sec.~\ref{sec:activation_steering}.

\section{Supplemental Experimental Results on LingBot-VA}\label{app:lingbot_results}

Although we show some results of WA-LQR on LingBot in Table~\ref{tab:ling_results}, we conducted a more comprehensive evaluation of WA-LQR applied in both the action module and the video module separately. Specifically, relative to Table \ref{tab:ling_results}, we further evaluate on Gaussian noise perturbations, on more task transfers, and on steering in the video module (denoted ``(Video)" in Table \ref{tab:ling_full_results}). To allow a fair comparison with Cosmos, we evaluate our method on LingBot-VA on the same set of tasks and perturbations as shown in Table ~\ref{tab:ling_full_results}. We did not observe significant effectiveness of steering, and this is supported by the result of our mechanistic analysis on the extracted activations from LingBot-VA's action and video modules, where it is shown that there is poor linear separability (with high numerical classification losses, especially relative to Cosmos), as presented in Fig \ref{fig:ling_video_cam}, \ref{fig:ling_video_noise}, and \ref{fig:ling_video_gripper}.

% \begin{table}[h]
% \centering
% \caption{LIBERO-10 \cite{liu2023libero} success rates on LingBot-VA \cite{lingbot-va2026}. Task $i\rightarrow$ Task $j$ denotes all $P_{l,t}, \tilde{A}_{l,t}, \tilde{B}_{l,t}$ matrices, $e^z_{l,t}$ vectors are computed with task $i$, and used to steer task $j$. 20 trials/task.}
% \renewcommand{\arraystretch}{0.9}
% \label{tab:cosmos_steering_results}
% \resizebox{0.65\textwidth}{!}{%
% % \begin{tabular}{lc cccc}
% \begin{tabular}{l c *{2}{C{1.5cm}}}
% \toprule
% \textbf{Perturbation} & \textbf{Tasks} & \textbf{No Steering} & \textbf{Ours} \\

% \midrule
% \multirow{6}{*}{\textbf{Camera Gaussian Noise}} 
%  & Task 6 $\rightarrow$ Task 0  & 35.0\%   & 20.0\%  \\
%  & Task 6 $\rightarrow$ Task 1  & 90.0\%  & 95.0\%  \\
%  & Task 6 $\rightarrow$ Task 4  & 75.0\%  & 75.0\%   \\
%  & Task 6 $\rightarrow$ Task 6  & 85.0\%  & 70.0\%   \\
%  & Task 6 $\rightarrow$ Task 7  & 10.0\%  & 20.0\% \\
%  \cmidrule{2-4}
%  & \textbf{Average}             & 59.0\%  & 56.0\%  \\
 
% \bottomrule
% \end{tabular}%
% }\vspace{-15pt}
% \end{table}

\begin{table}[h]
\centering
\caption{LIBERO-10 \cite{liu2023libero} success rates on LingBot-VA \cite{lingbot-va2026}. Task $i\rightarrow$ Task $j$ denotes all $P_{l,t}, \tilde{A}_{l,t}, \tilde{B}_{l,t}$ matrices, $e^z_{l,t}$ vectors are computed with task $i$, and used to steer task $j$. 20 trials/task.}
\renewcommand{\arraystretch}{0.9}
\setlength{\tabcolsep}{10pt}
\label{tab:ling_full_results}
\resizebox{0.85\textwidth}{!}{%
% \begin{tabular}{lc cccc}
\begin{tabular}{l c *{4}{C{1.5cm}}}
\toprule
\textbf{Perturbation} & \textbf{Tasks} & \textbf{No Steering} &   \textbf{ActAdd}&\textbf{Ours (Action)} & \textbf{Ours (Video)} \\
\midrule
\multirow{6}{*}{\textbf{Camera Orientation}} 
 & Task 0 $\rightarrow$ Task 0  & 65.0\%$\pm$10.7\% & 35.0\%$\pm$10.7\% & 70.0\%$\pm$10.3\% & 65.0\%$\pm$10.7\% \\
 & Task 0 $\rightarrow$ Task 2  & 55.0\%$\pm$11.1\% & 45.0\%$\pm$11.1\% & 45.0\%$\pm$11.1\% & 55.0\%$\pm$11.1\% \\
 & Task 0 $\rightarrow$ Task 4  & 15.0\%$\pm$8.0\% & 10.0\%$\pm$6.7\% & 30.0\%$\pm$10.3\% & 10.0\%$\pm$6.7\% \\
 & Task 0 $\rightarrow$ Task 5  & 70.0\%$\pm$10.3\% & 75.0\%$\pm$9.7\% & 60.0\%$\pm$11.0\% & 65.0\%$\pm$10.7\% \\
 & Task 0 $\rightarrow$ Task 9  & 35.0\%$\pm$10.7\% & 40.0\%$\pm$11.0\% & 50.0\%$\pm$11.2\% & 60.0\%$\pm$11.0\% \\
 \cmidrule{2-6}
 & \textbf{Average}             & 48.0\%$\pm$5.0\% & 41.0\%$\pm$4.9\% & 51.0\%$\pm$5.0\% & 51.0\%$\pm$5.0\% \\

\midrule

\multirow{6}{*}{\textbf{Initial Gripper Position}} 
 & Task 1 $\rightarrow$ Task 1  & 65.0\%$\pm$10.7\% & 75.0\%$\pm$9.7\% & 75.0\%$\pm$9.7\% & 70.0\%$\pm$10.3\% \\
 & Task 1 $\rightarrow$ Task 2  & 70.0\%$\pm$10.3\% & 70.0\%$\pm$10.3\% & 80.0\%$\pm$8.9\% & 70.0\%$\pm$10.3\% \\
 & Task 1 $\rightarrow$ Task 3  & 70.0\%$\pm$10.3\% & 85.0\%$\pm$8.0\% & 75.0\%$\pm$9.7\% & 85.0\%$\pm$8.0\% \\
 & Task 1 $\rightarrow$ Task 7  & 80.0\%$\pm$8.9\% & 60.0\%$\pm$11.0\% & 80.0\%$\pm$8.9\% & 80.0\%$\pm$8.9\% \\
 & Task 1 $\rightarrow$ Task 9  & 75.0\%$\pm$9.7\% & 70.0\%$\pm$10.3\% & 65.0\%$\pm$10.7\% & 70.0\%$\pm$10.3\% \\
 \cmidrule{2-6}
 & \textbf{Average}             & 72.0\%$\pm$4.5\% & 72.0\%$\pm$4.5\% & 75.0\%$\pm$4.3\% & 75.0\%$\pm$4.3\% \\
 
\midrule

\multirow{6}{*}{\textbf{Camera Gaussian Noise}} 
 & Task 6 $\rightarrow$ Task 0  & 35.0\%$\pm$10.7\% & 15.0\%$\pm$8.0\% & 20.0\%$\pm$8.9\% & 65.0\%$\pm$10.7\% \\
 & Task 6 $\rightarrow$ Task 1  & 90.0\%$\pm$6.7\% & 60.0\%$\pm$11.0\% & 95.0\%$\pm$4.9\% & 20.0\%$\pm$8.9\% \\
 & Task 6 $\rightarrow$ Task 4  & 75.0\%$\pm$9.7\% & 45.0\%$\pm$11.1\% & 75.0\%$\pm$9.7\% & 100.0\%$\pm$0.0\% \\
 & Task 6 $\rightarrow$ Task 6  & 85.0\%$\pm$8.0\% & 55.0\%$\pm$11.1\% & 70.0\%$\pm$10.3\% & 75.0\%$\pm$9.7\% \\
 & Task 6 $\rightarrow$ Task 7  & 10.0\%$\pm$6.7\% & 15.0\%$\pm$8.0\% & 20.0\%$\pm$8.9\% & 20.0\%$\pm$8.9\% \\
 \cmidrule{2-6}
 & \textbf{Average}             & 59.0\%$\pm$4.9\% & 38.0\%$\pm$4.9\% & 56.0\%$\pm$5.0\% & 56.0\%$\pm$5.0\% \\
 
\bottomrule
\end{tabular}%
}\vspace{-15pt}
\end{table}

\clearpage
\section{ActAdd Sensitivity Analysis}\label{app:actadd_sensitivity}

We perform a sensitive analysis of the effectiveness of ActAdd on improving the performance of Cosmos-Policy 2B \cite{kim2026cosmos} under Gaussian noise perturbation in the sensor input. The steering vectors are acquired from Task06 and used to steer the model over 5 tasks. As shown in Tab.~\ref{tab:gamma-performance}, ActAdd demonstrates a sharp peak in success rates at a specific hyperparameter ($\gamma=0.1$), and this performance boost quickly vanishes with smaller or larger $\gamma$.

\begin{table}[h]
\centering
\caption{Performance across different values of $\gamma$ under sensor input perturbed with Gaussian noise.}
\renewcommand{\arraystretch}{0.9}
\setlength{\tabcolsep}{10pt}
\label{tab:gamma-performance}
\resizebox{0.85\textwidth}{!}{%
\begin{tabular}{c c *{5}{C{1.5cm}}}
\toprule
$\boldsymbol{\gamma}$
& \textbf{Mean}
& \textbf{Task06}
& \textbf{Task00}
& \textbf{Task01}
& \textbf{Task04}
& \textbf{Task07} \\
\midrule
0.0   & 26.6\% & 37\% & 3\%  & 53\% & 10\% & 30\% \\
0.02  & 36.2\% & 37\% & 17\% & 67\% & 13\% & 47\% \\
0.05  & 50.8\% & 57\% & 37\% & 73\% & 27\% & 60\% \\
0.075 & 52.6\% & 70\% & 37\% & 70\% & 43\% & 43\% \\
0.1   & 67.4\% & 77\% & 43\% & 70\% & 80\% & 67\% \\
0.15  & 32.6\% & 13\% & 10\% & 17\% & 80\% & 43\% \\
0.2   & 0.0\%  & 0\%  & 0\%  & 0\%  & 0\%  & 0\% \\
0.25  & 0.0\%  & 0\%  & 0\%  & 0\%  & 0\%  & 0\% \\
0.5   & 0.0\%  & 0\%  & 0\%  & 0\%  & 0\%  & 0\% \\
\bottomrule
\end{tabular}%
}
\vspace{-15pt}
\end{table}

\clearpage
\section{Experimental Details}\label{app:experimental_details}

\subsection{Mechanistic Study}\label{app:subsec:mechanistic_details}
To analyze the geometry of each model's latent activations, we operate directly in the full activation space, using mean-pooled activations across token positions as described in Sec.~\ref{sec:mechanistic}. For Cosmos-Policy, which utilizes a unified single-backbone architecture, we consider the activations across all latent frames. For LingBot-VA and DiT4DiT, which adopt inverse-dynamics style architectures with distinct DiTs for different modalities, we consider the activations from the action-generation DiT. This is consistent with our steering formulation (App.~\ref{app:subsec:cosmos_exp_details}-\ref{app:subsec:ling_exp_details}). 

To collect contrastive examples for each perturbation type, we conduct a similar procedure to the contrastive vector collection process in App.~\ref{app:contrastive}. That is, we consider successful vs. unsuccessful rollout as $\mathcal D_+$ and $\mathcal D_-$, respectively, under camera or gripper position perturbation. For Gaussian noise, we let $\mathcal{D_+} = \{\texttt{Clean inputs} \}$, and $\mathcal{D_-} = \{\texttt{Noised inputs} \}$. We also evaluated the gripper and camera perturbation in terms of $\mathcal{D_+} = \{\texttt{Without perturbation} \}$, and $\mathcal{D_-} = \{\texttt{With perturbation} \}$, but did not observe a meaningful difference.

Given $\mathcal{D_+}$ and $\mathcal{D_-}$, we compute contrastive directions, and conduct Principal Component Analysis on the set of contrastive directions, as described in Sec.~\ref{sec:mechanistic}. To fit the support vector machine (SVM), we initialize the separating $\{x|\hat w^Tx+\hat b=0\}$ hyperplane as the zero vector and perform gradient descent with the loss function:
\begin{equation}\label{app:eq:svm_loss}
    \texttt{loss}_\texttt{SVM}(\hat w, \hat b) = (1/2) \hat w^\top \hat w + C \cdot \sum_{i\in \left[N\right]} \max(0, 1-y_i(\hat w^T\bar{x}_i+\hat b)),
\end{equation}
where $C$ is some constant (in our experiments, we set $C = 10$). Note the second term in Eq.~\ref{app:eq:svm_loss} resembles the hinge loss in Eq.~\ref{eq:hinge_loss}, evaluated on the intermediate hyperplane parameterized by $\hat w, \hat b$. The SVM is always computed in the three-dimensional subspace defined by the top three principal components of the contrastive directions. 
% $\textstyle\texttt{loss}(\mathcal{D}_+, \mathcal{D}_-) = (1/N) \sum_{i\in \left[N\right]} \max(0, 1-y_i(w^T\bar{x}_i+b)),$

\subsection{Cosmos-Policy Evaluations on LIBERO-10}\label{app:subsec:cosmos_exp_details}
Cosmos-Policy is a diffusion-based world-action model with a diffusion transformer backbone, where it jointly denoises a robot action chunk and future state predictions (future proprioception, wrist image, and third-person image). In our method, steering is applied at all latent outputs except for the action chunks. 
To construct contrastive vectors, activations are collected from all 28 transformer blocks across 5 denoising timesteps for matched pairs of clean and perturbed observations of the same task and scene.

For each layer and timestep, a randomized SVD is run on the paired matrix to produce a rank-64 basis to reduce dimensionality. The compact subspace is used to find the linearized dynamics that approximates how activations propagate from one layer to the next. The LQR gains are pre-computed via backward Riccati recursion, where the control cost grows exponentially with the action chunk index. An LQR hyperparameter search is conducted before settling on the best-performing parameters for all other tasks of the same scene. During inference time, the projected activation error relative to the nominal trajectory is used to inject a correction through the precomputed gains.

\subsection{LingBot-VA Evaluations on LIBERO-10}\label{app:subsec:ling_exp_details}

LingBot-VA is a mixture-of-transformers (MoT) based world-action model. It first uses a video module to predict future visual frames, which are then passed to a lightweight action module with a smaller hidden dimension to generate robot actions. In our experiments, we apply our method to either the video module or the action module while keeping the other component unchanged. 

For both modules, we collect activations from the outputs of all 30 transformer blocks. Since the video and action modules use different denoising schedules, we select module-specific denoising timesteps. For the action module, we use timesteps $(t \in {0,10,20,30,40})$. For the video module, we use timesteps $(t \in {0,4,9,14,19})$. These timesteps are chosen to cover the corresponding denoising trajectory of each module. 

To reduce dimensionality, we partition the 30 transformer layers into three groups: layers 0--9, 10--19, and 20--29. For each layer partition and denoising timestep, we pool activation differences of contrastive pairs from all layers within the corresponding partition and compute a shared rank-64 SVD basis. The produced compact subspace for each layer group and timestep pair is used by our LQR injector during evaluation. At evaluation time, the LQR injector follows the same layer partitioning and timestep mapping. For each activation, we project it onto the corresponding low-dimensional SVD subspace, compute the LQR correction in this reduced space, and map the correction back to the full activation space before applying it to the model.

\subsection{DiT4DiT Evaluations on LIBERO-10}\label{app:subsec:dit4dit_exp_details}

Similar to LingBot-VA, DiT4DiT is a mixture-of-transformers (MoT) based world-action model with distinct video and action modules. In our experiments, we apply our method to the action module, and do not intervene on the video generation module. Otherwise, our evaluation procedure structurally follows Cosmos-Policy on all token positions, intervening across all 4 diffusion timesteps and 16 model layers.

The dimensionality reduction procedure also follows the other models, with action generation layers partitioned into three groups: layers 0-5, 6-10, and 11-15. For each layer pool and diffusion timestep, we pool contrastive vectors and analogously construct a 64-dimensional SVD basis, which defines the projection onto the low-dimensional latent space. Jacobian construction, LQR, and control synthesis follow the same procedure as the other two models.

\clearpage

\section{Complete Low-Dimensional Separability Results}\label{ap:separability}
\FloatBarrier

\begin{figure}
    \centering
    \includegraphics[width=1\linewidth]{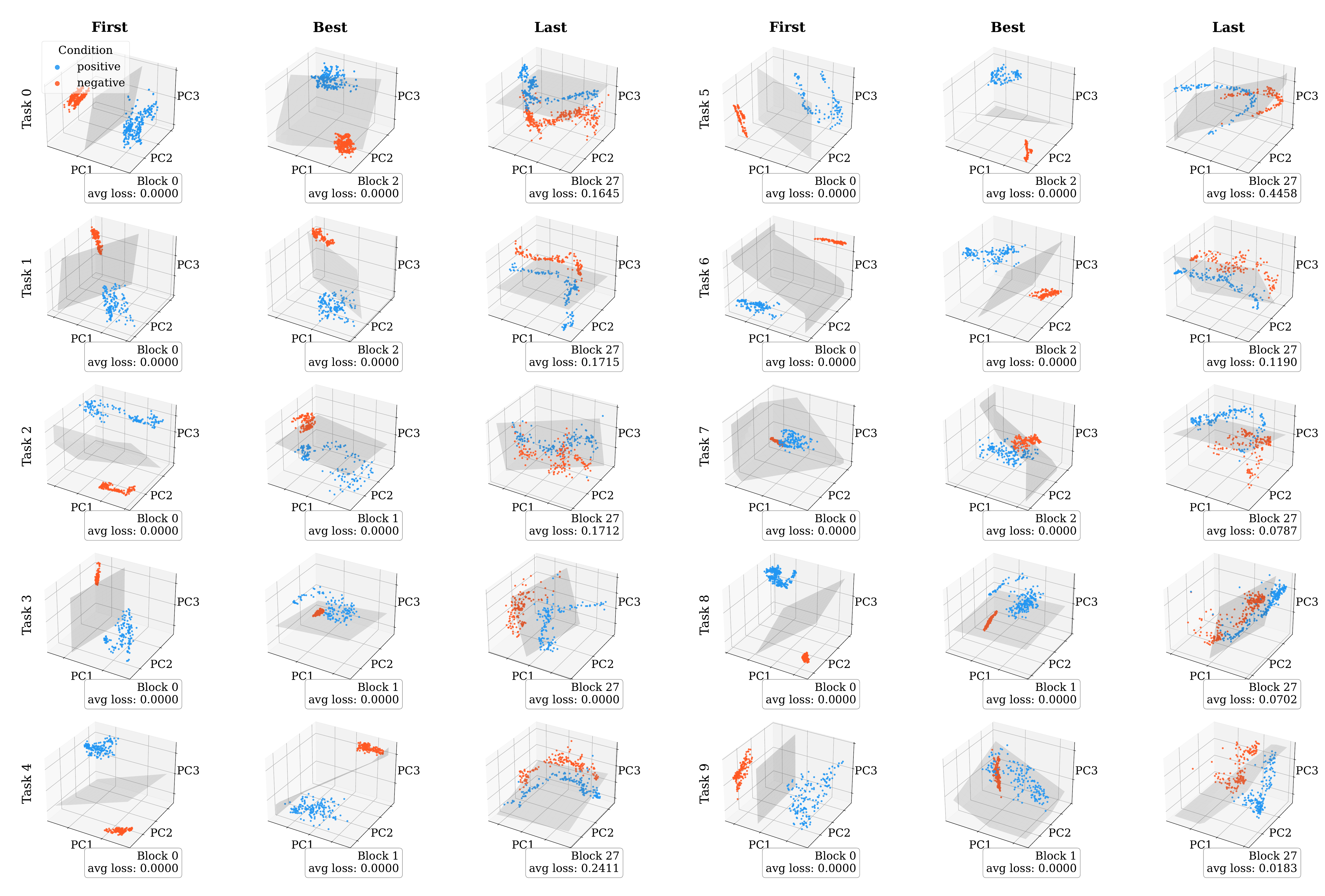}
    \caption{Cosmos-Policy noise corruption separation for all LIBERO-10 tasks}
    \label{fig:cosmos_full_noise}
\end{figure}

\begin{figure}
    \centering
    \includegraphics[width=1\linewidth]{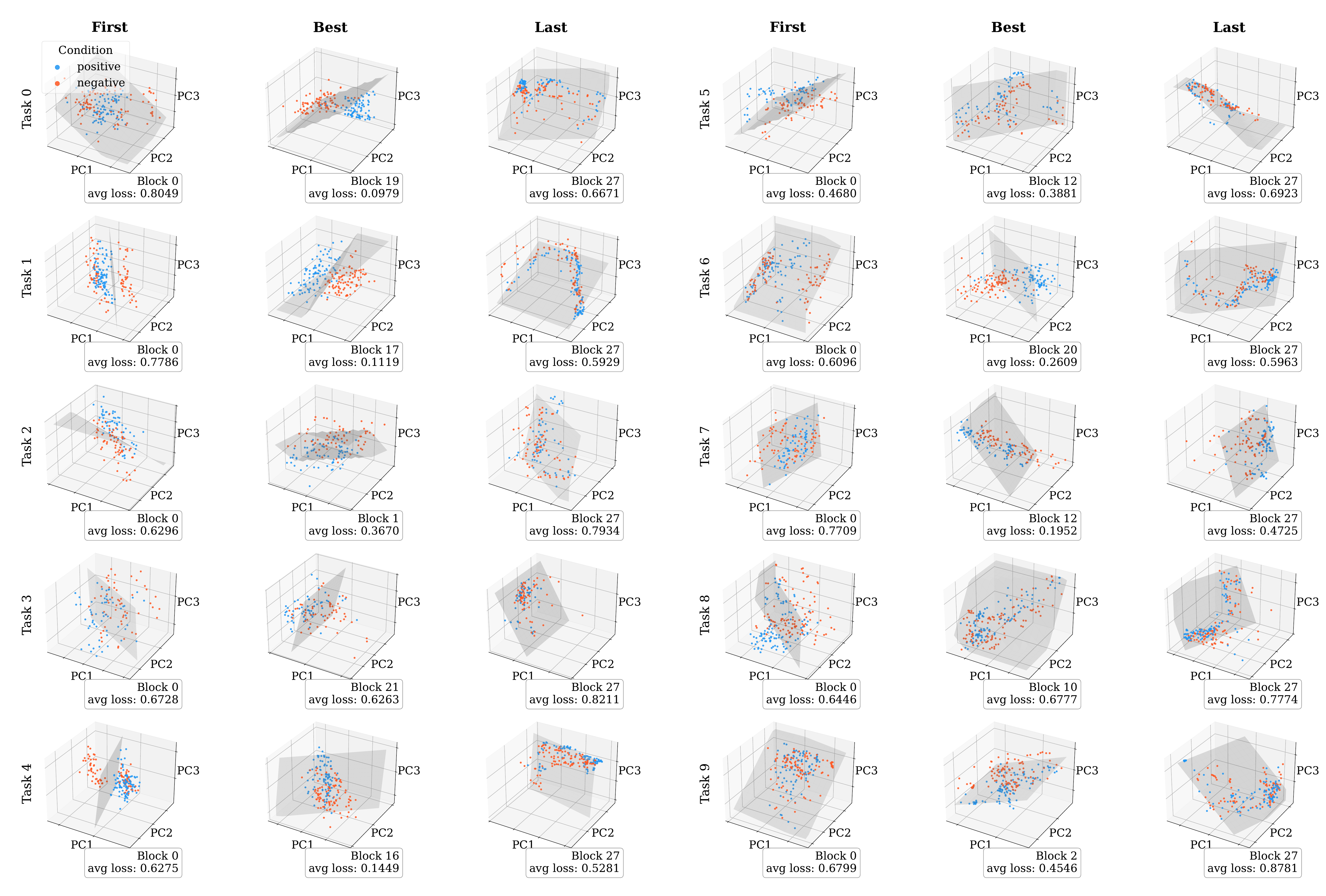}
    \caption{Cosmos-Policy camera perturbation separation for all LIBERO-10 tasks.}
    \label{fig:cosmos_full_cam}
\end{figure}

\begin{figure}
    \centering
    \includegraphics[width=1\linewidth]{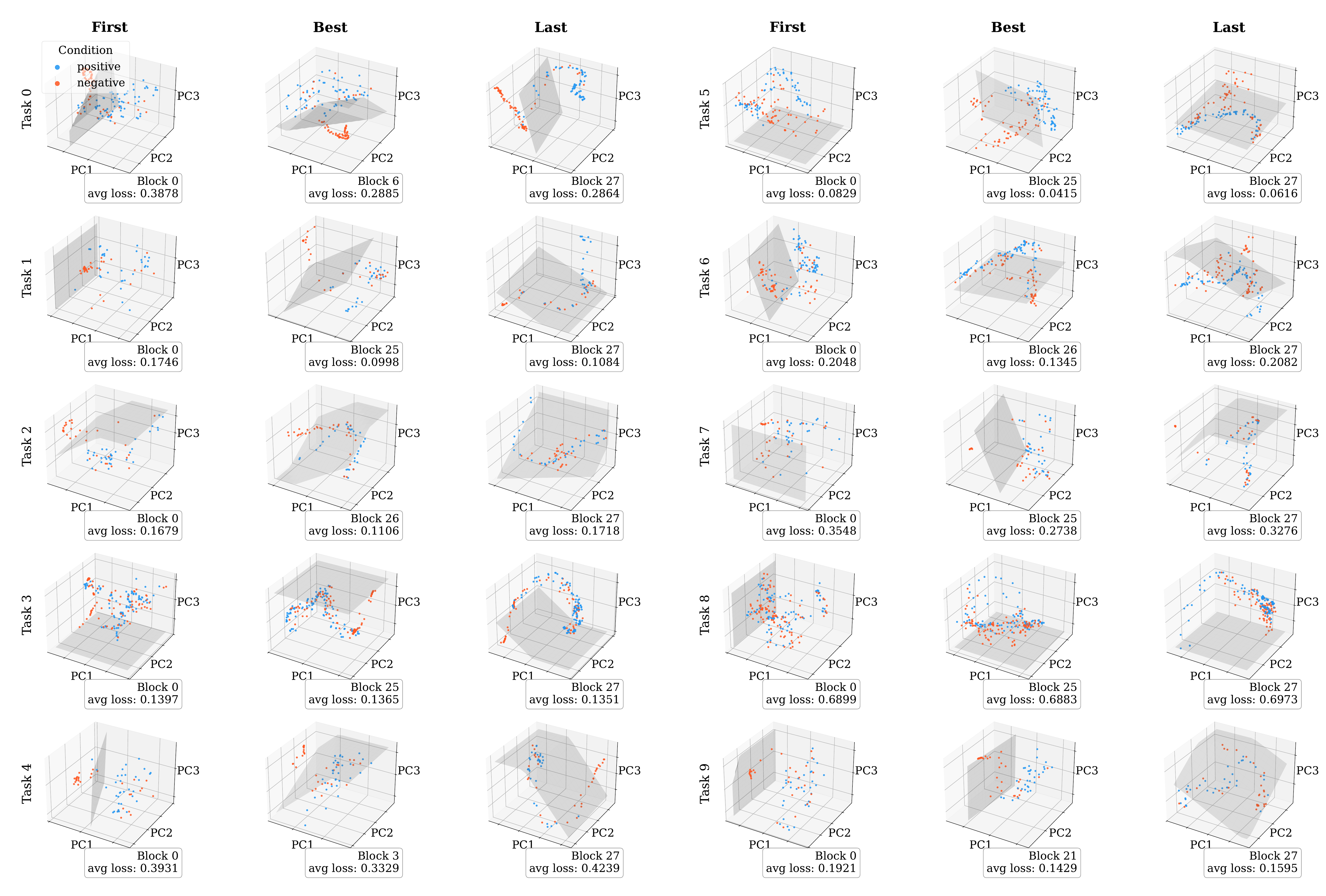}
    \caption{Cosmos-Policy gripper perturbation separation for all LIBERO-10 tasks.}
    \label{fig:cosmos_full_gripper}
\end{figure}

\begin{figure}
    \centering
    \includegraphics[width=1\linewidth]{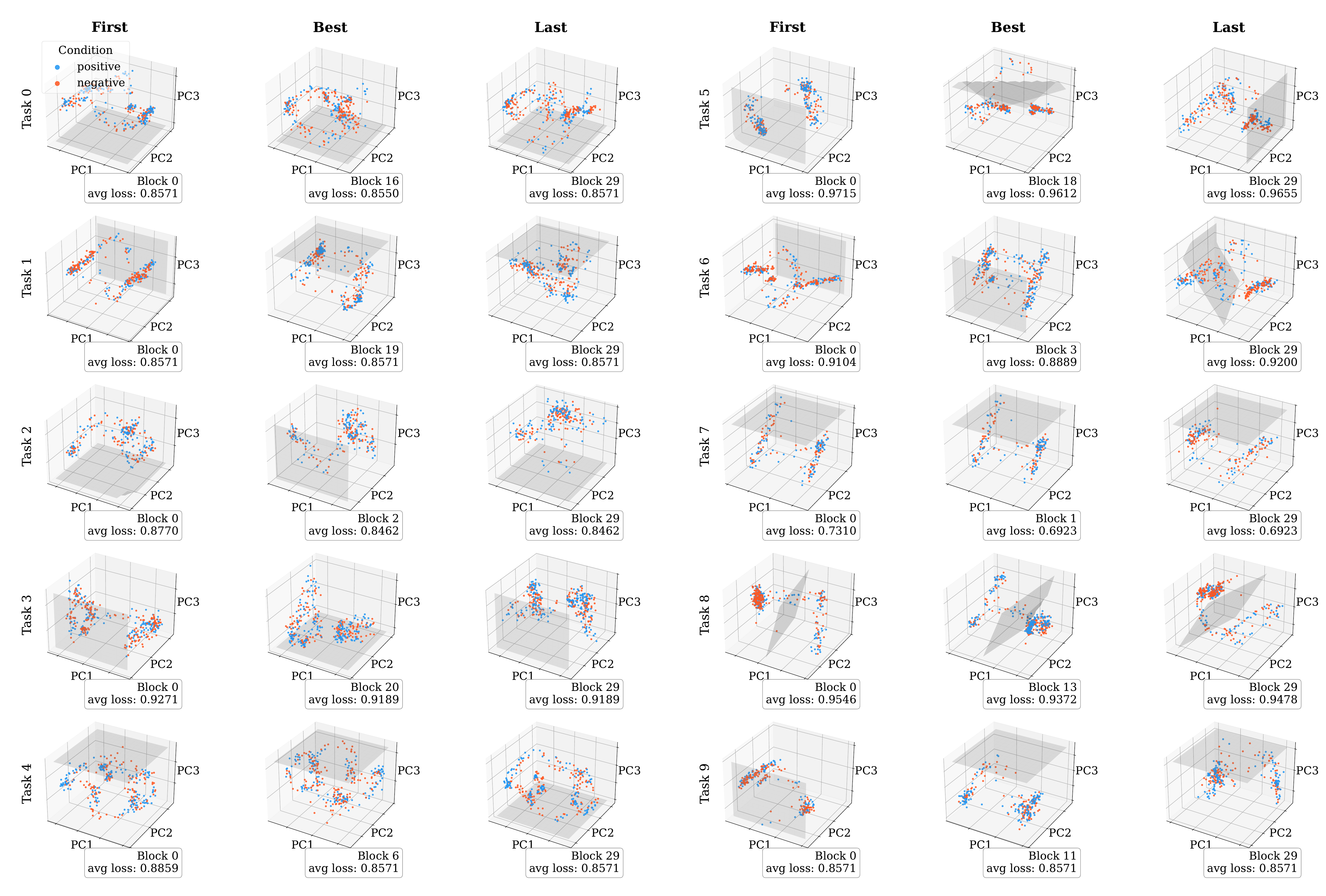}
    \caption{LingBot-VA camera perturbation separation for all LIBERO-10 tasks (Action Module).}
    \label{fig:ling_full_cam}
\end{figure}

\begin{figure}
    \centering
    \includegraphics[width=1\linewidth]{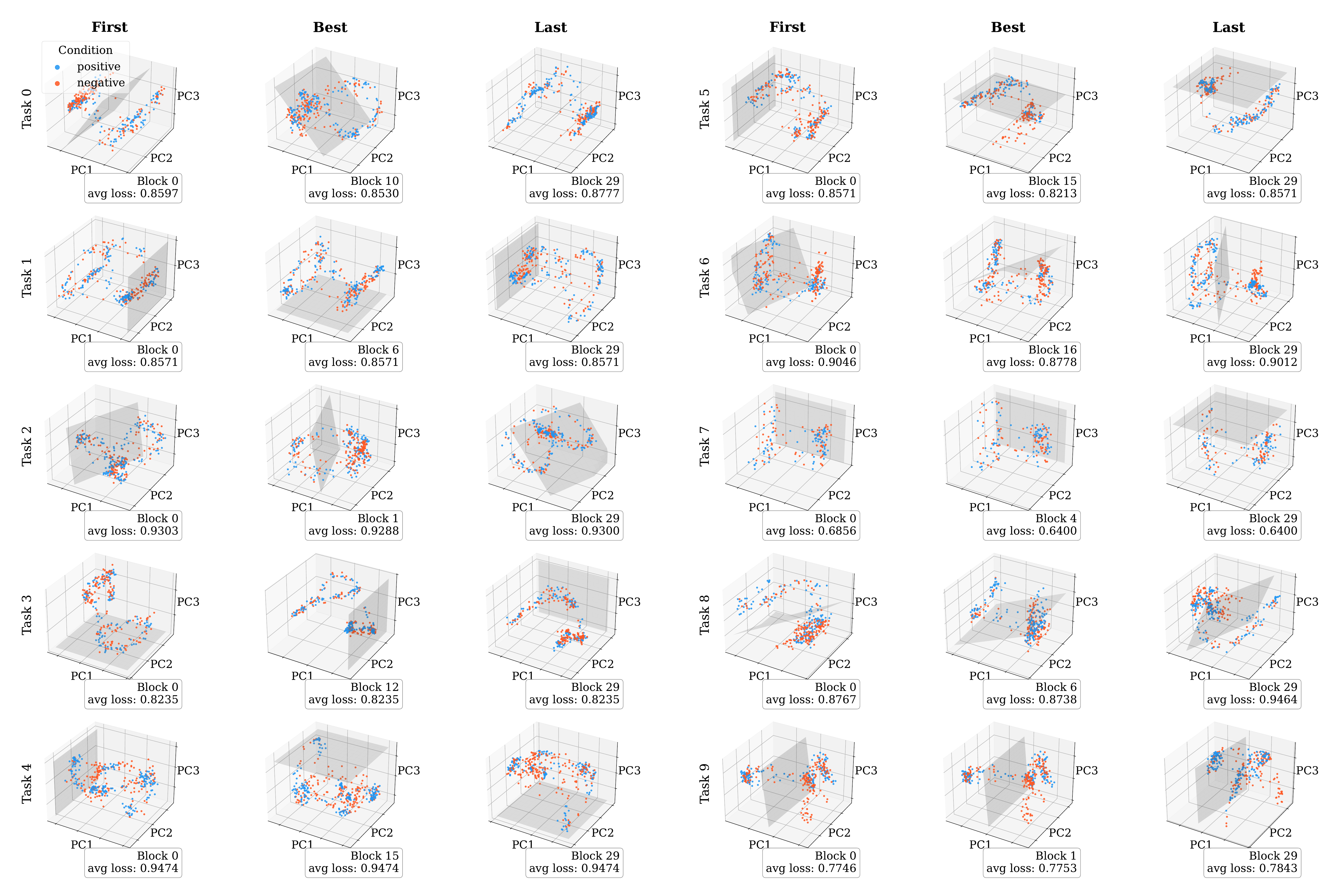}
    \caption{LingBot-VA gripper perturbation separation for all LIBERO-10 tasks (Action Module).}
    \label{fig:ling_full_gripper}
\end{figure}

\begin{figure}
    \centering
    \includegraphics[width=1\linewidth]{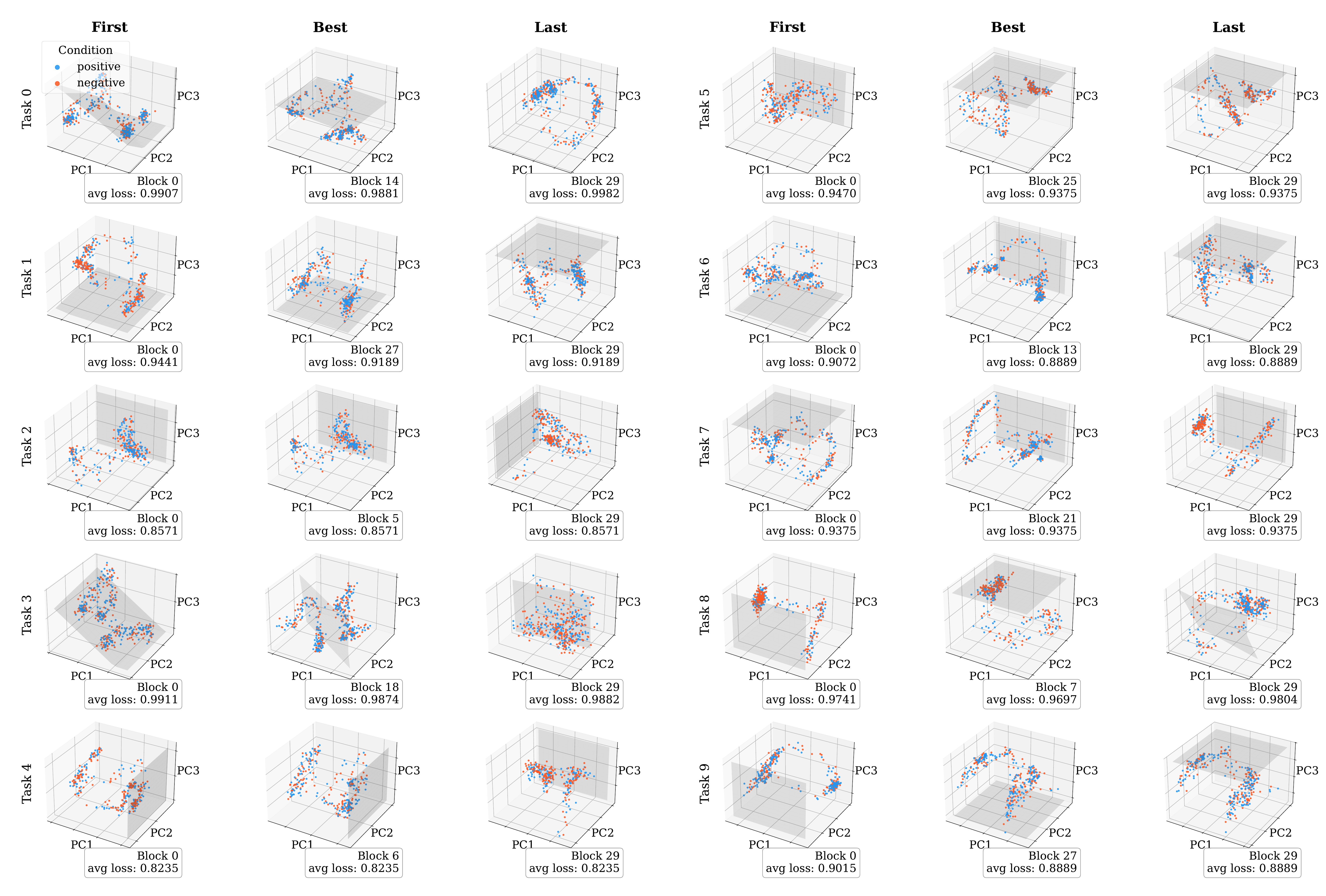}
    \caption{LingBot-VA noise corruption separation for all LIBERO-10 tasks (Action Module).}
    \label{fig:ling_full_noise}
\end{figure}

\begin{figure}
    \centering
    \includegraphics[width=1\linewidth]{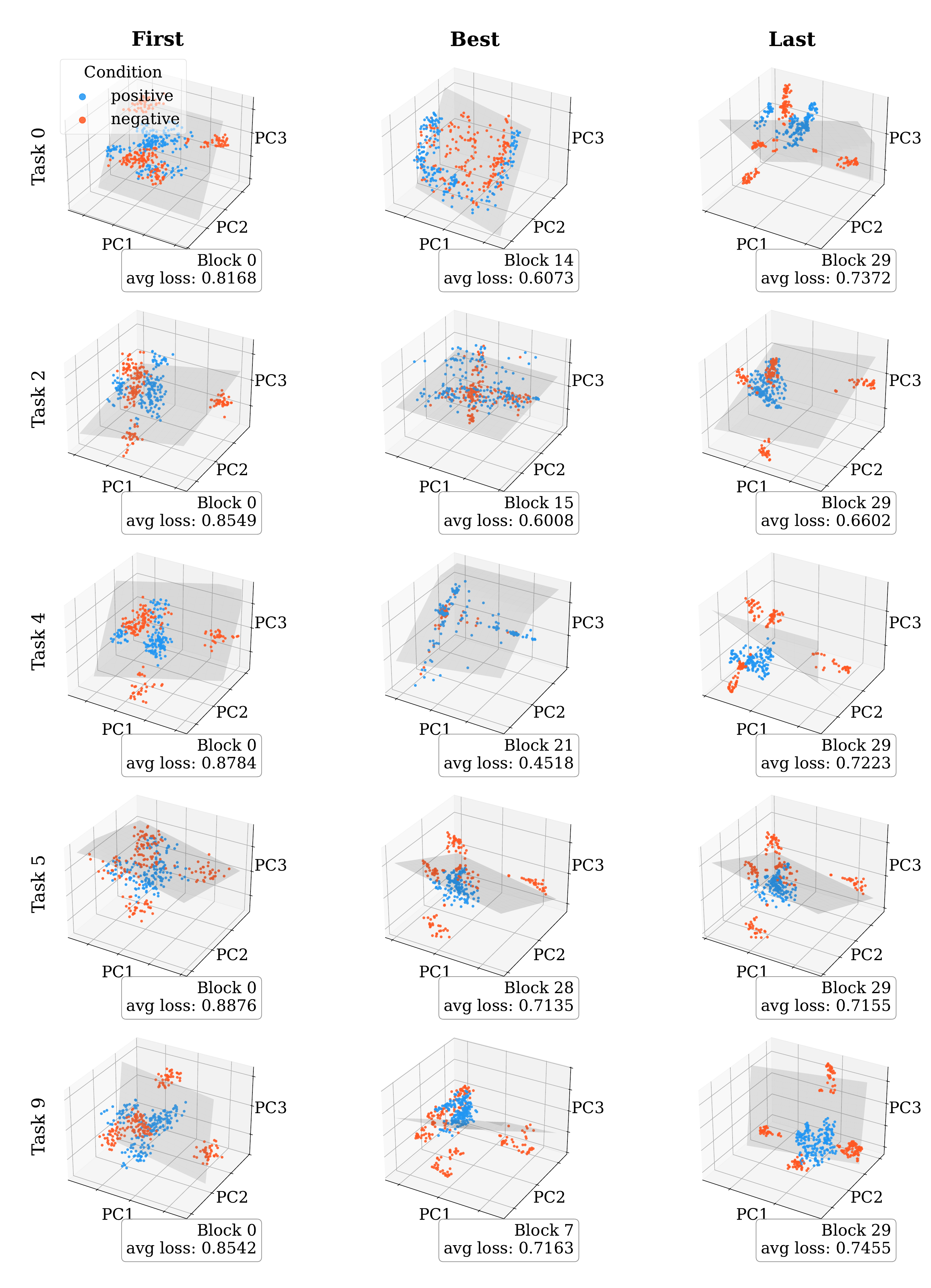}
    \caption{LingBot-VA camera orientation perturbation separation for LIBERO-10 task 0, 2, 4, 5, 9 (Video Module).}
    \label{fig:ling_video_cam}
\end{figure}

\begin{figure}
    \centering
    \includegraphics[width=1\linewidth]{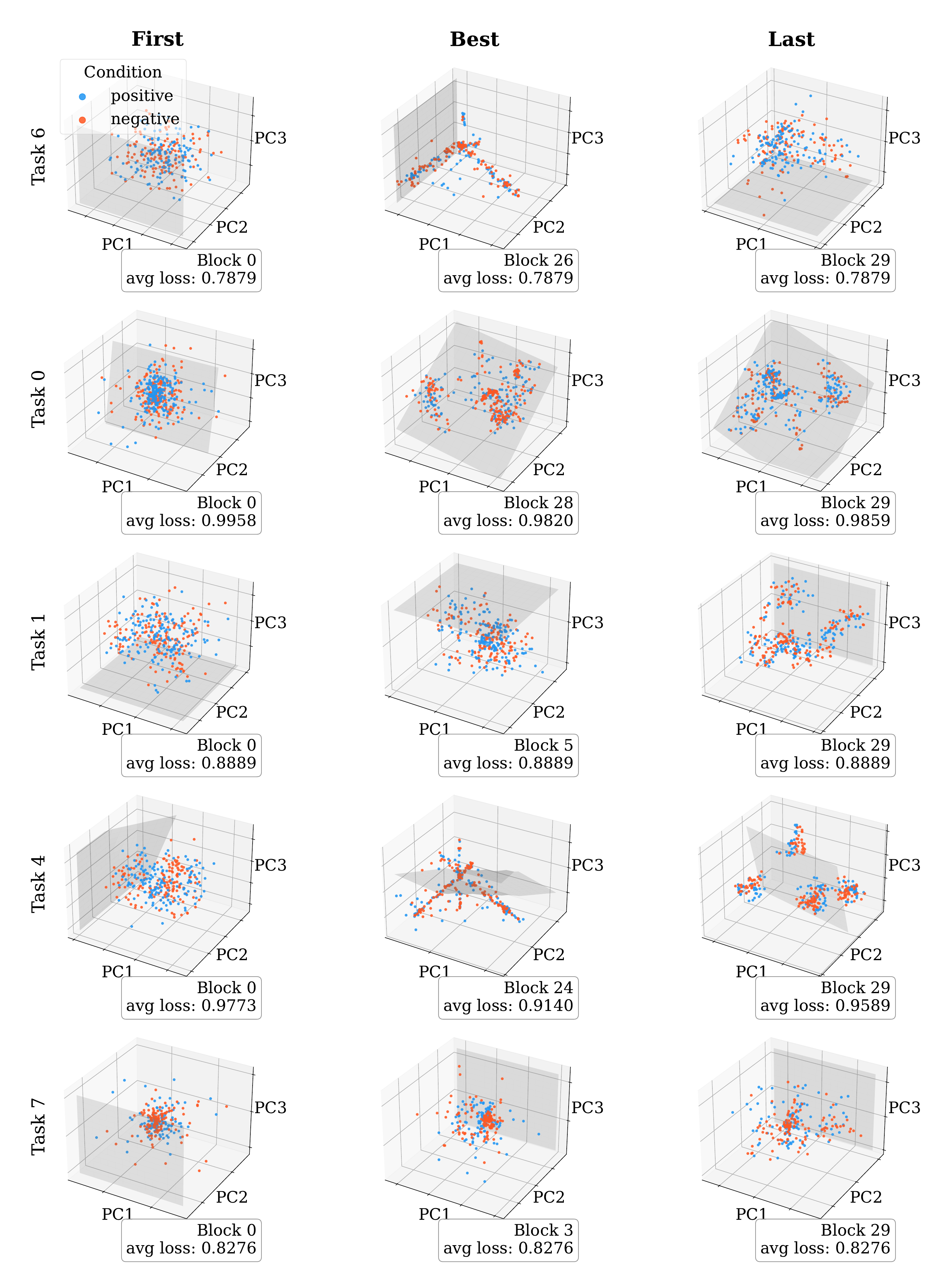}
    \caption{LingBot-VA noise corruption separation for LIBERO-10 task 0, 1, 4, 6, 7 (Video Module).}
    \label{fig:ling_video_noise}
\end{figure}

\begin{figure}
    \centering
    \includegraphics[width=1\linewidth]{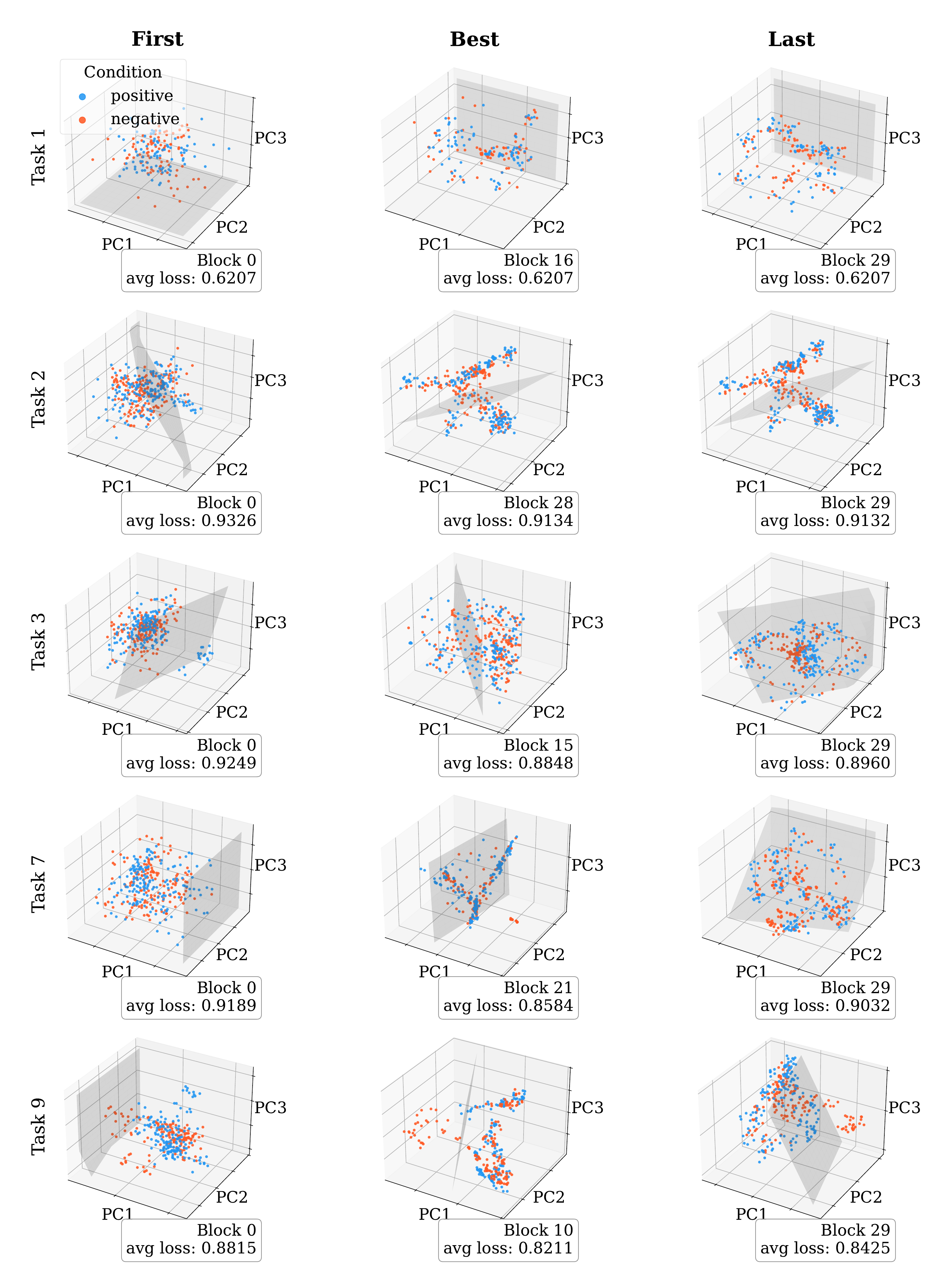}
    \caption{LingBot-VA gripper perturbation separation for LIBERO-10 task 1, 2, 3, 7, 9 (Video Module).}
    \label{fig:ling_video_gripper}
\end{figure}

\begin{figure}
    \centering
    \includegraphics[width=1\linewidth]{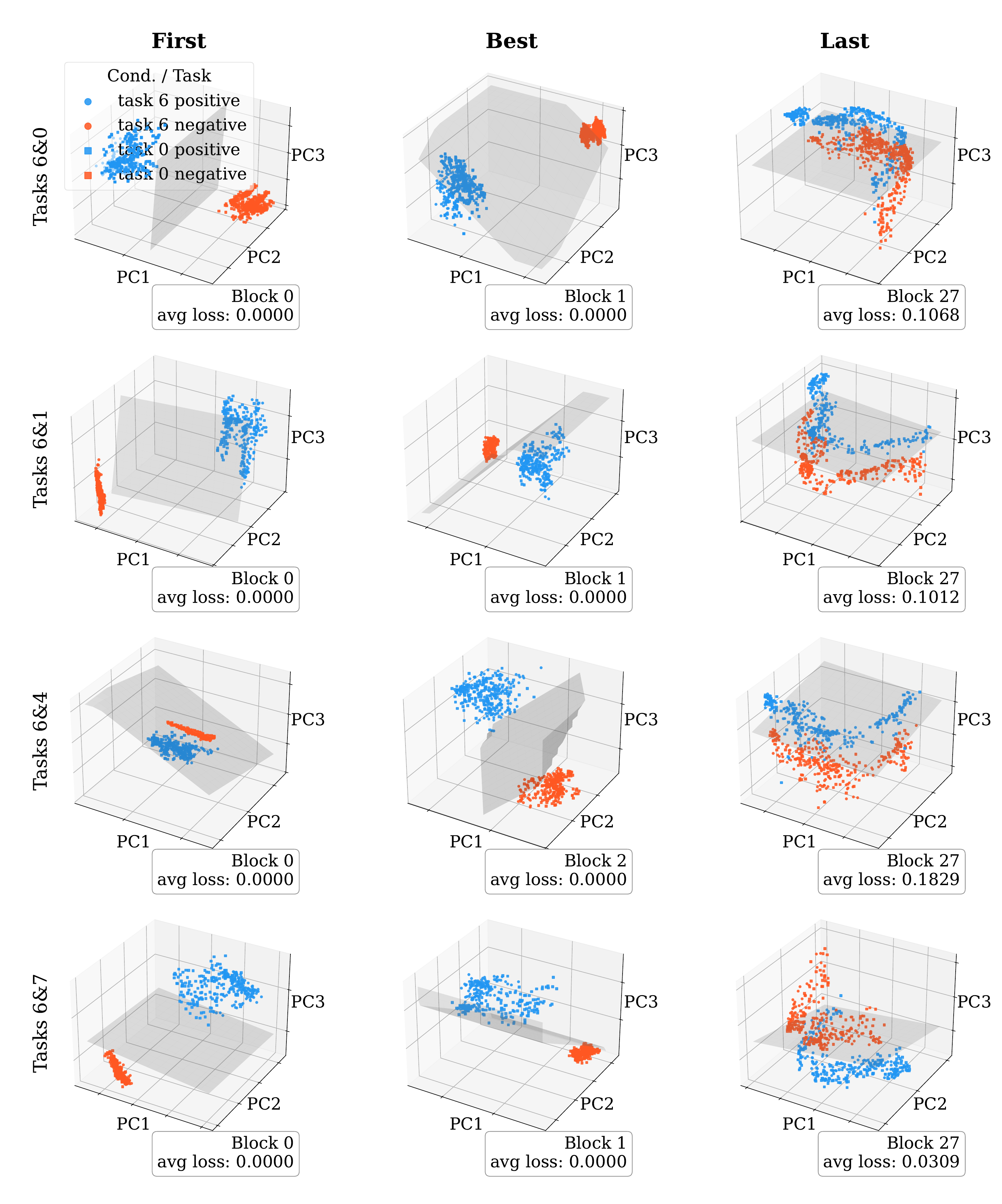}
    \caption{Pairwise separability for Cosmos-Policy with Gaussian noise corruption. }
    \label{ap:fig:cosmos_noise_pairs}
\end{figure}

\begin{figure}
    \centering
    \includegraphics[width=1\linewidth]{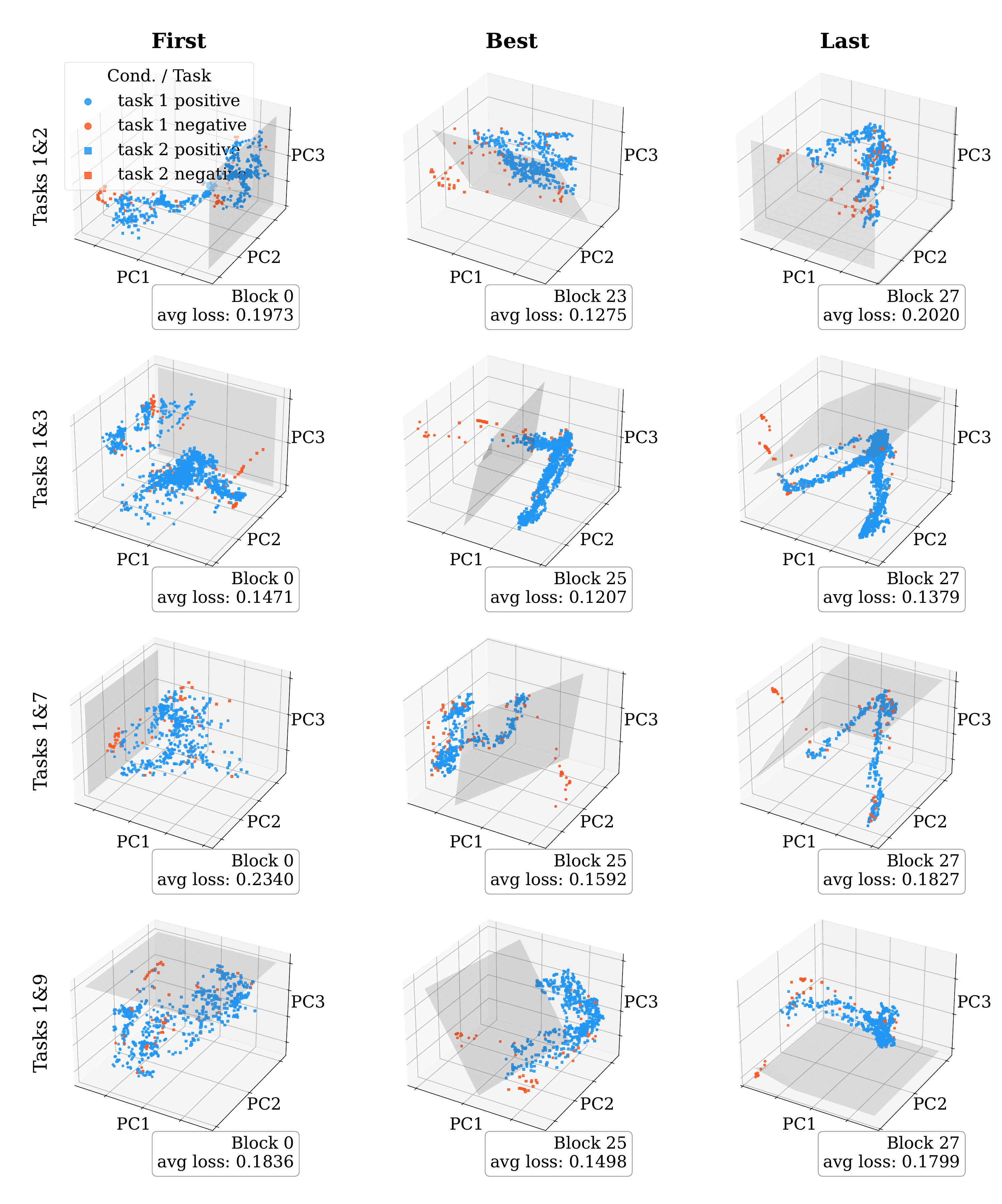}
    \caption{Pairwise separability for Cosmos-Policy with gripper position perturbation. }
    \label{fig:cosmos_gripper_pairs}
\end{figure}

\begin{figure}
    \centering
    \includegraphics[width=1\linewidth]{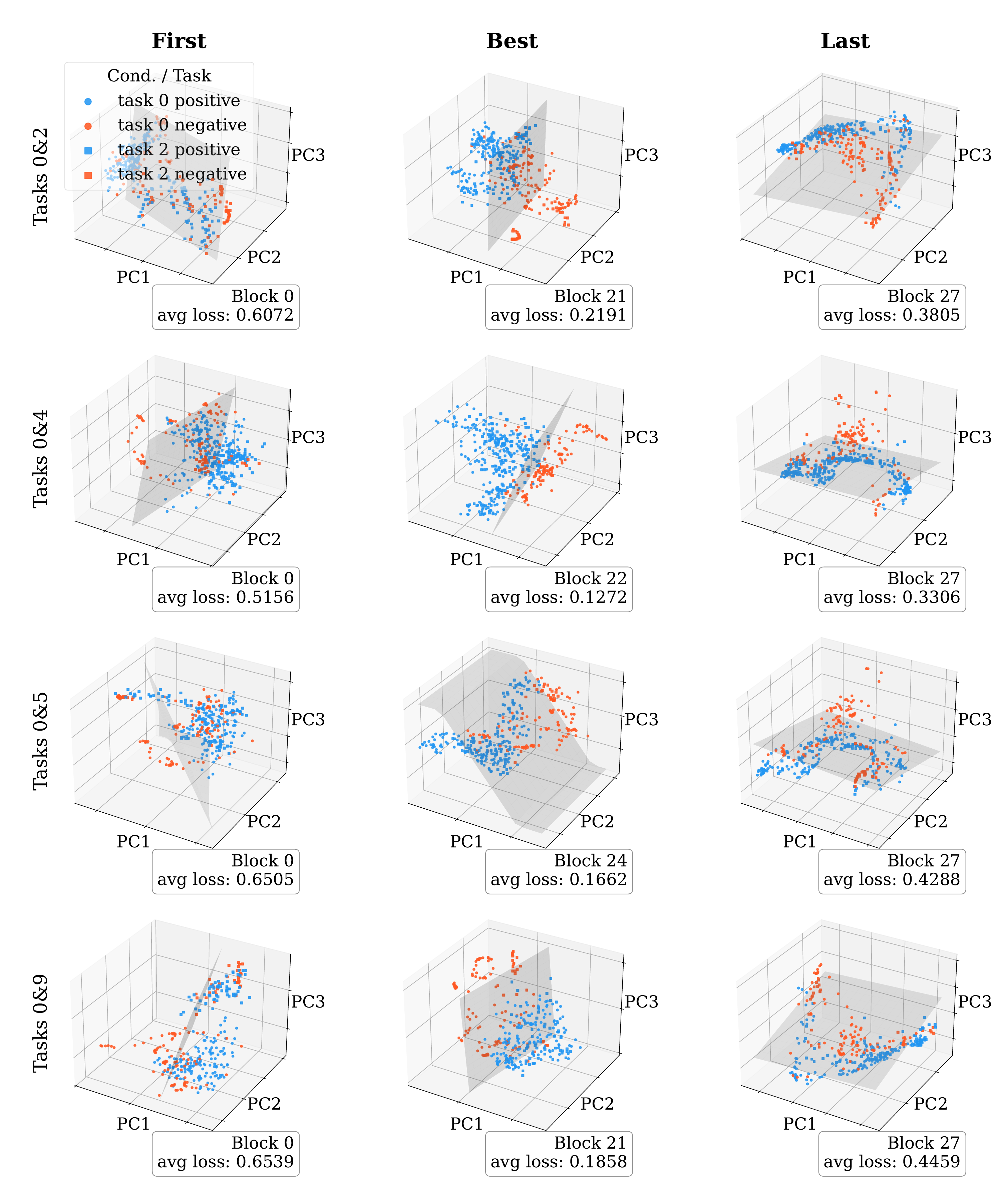}
    \caption{Pairwise separability for Cosmos-Policy with camera position perturbation. }
    \label{fig:cosmos_camera_pairs}
\end{figure}

We report the feature separation for three considered robustness features: perturbation to initial gripper position, initial camera position, and corruption of camera inputs with Gaussian noise, for all tasks in the LIBERO-10 dataset \cite{liu2023libero}. The results are summarized in Figs.~\ref{fig:cosmos_full_noise}-\ref{fig:ling_full_noise}. As reported in Sec.~\ref{sec:mechanistic}, we observe linear separation across tasks and perturbations in Cosmos-Policy (Figs.~\ref{fig:cosmos_full_noise}-\ref{fig:cosmos_full_gripper}), but do not observe such separation in LingBot-VA (Figs.~\ref{fig:ling_full_cam}-\ref{fig:ling_video_gripper}). For Cosmos-Policy, where we observe meaningful separation, we also present pairwise separability plots (Fig.~\ref{ap:fig:cosmos_noise_pairs}-\ref{fig:cosmos_camera_pairs}), generated by aggregating the activations for pairs of tasks and otherwise following the same procedure as before.

\FloatBarrier

\section{Full Model Separation}\label{app:full_model_separation}

For completeness, we report the full mechanistic separation plots across all layers in the following.

\begin{figure}[h]
    \centering
    \includegraphics[width=1\linewidth]{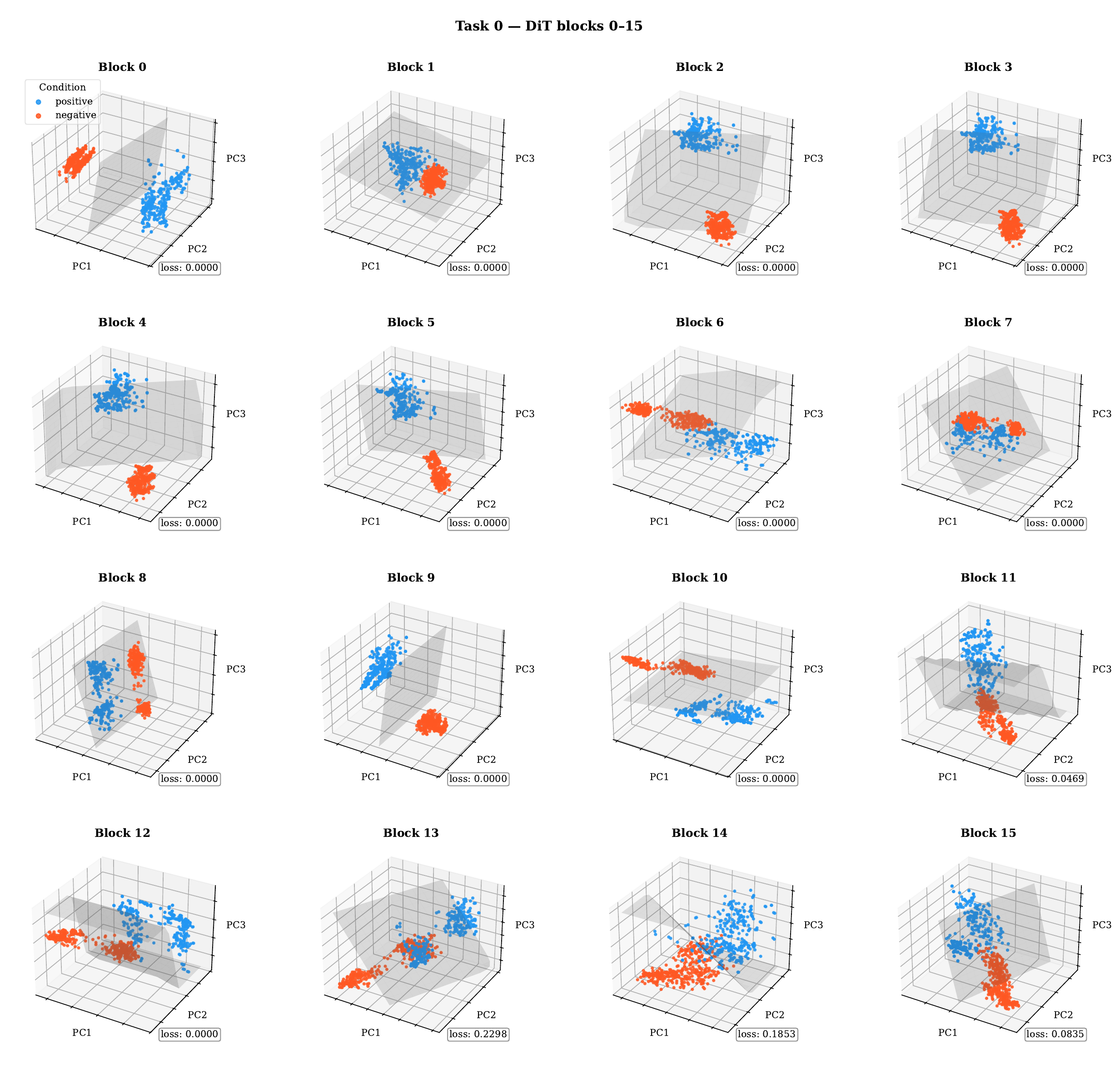}
    \caption{Cosmos-Policy Task 0 noise corruption}
    \label{fig:cosmos_allblocks_t0_p1}
\end{figure}

\begin{figure}
    \centering
    \includegraphics[width=1\linewidth]{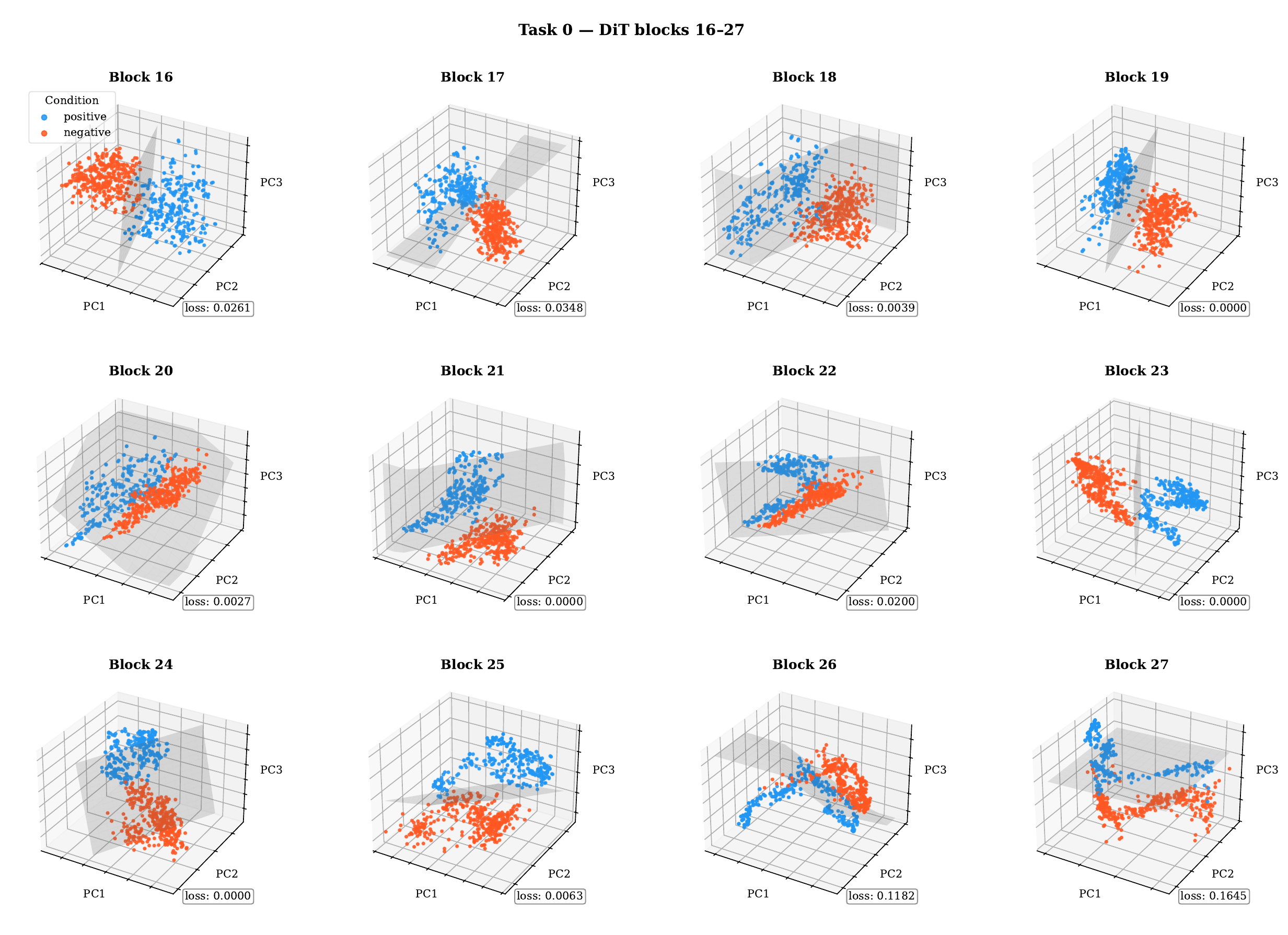}
    \caption{Cosmos-Policy Task 0 noise corruption}
    \label{fig:cosmos_allblocks_t0_p2}
\end{figure}

\begin{figure}
    \centering
    \includegraphics[width=1\linewidth]{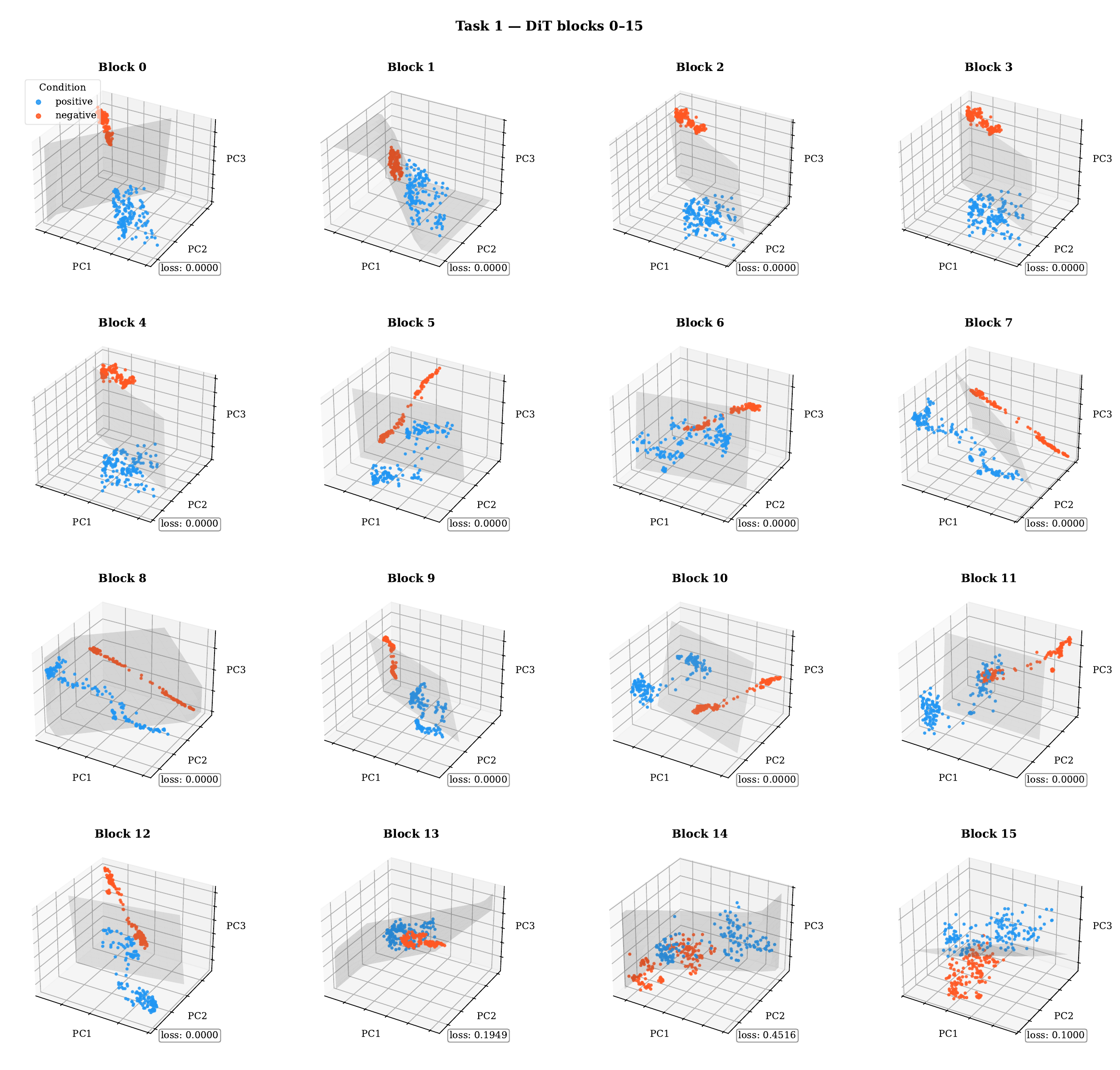}
    \caption{Cosmos-Policy Task 1 noise corruption}
    \label{fig:cosmos_allblocks_t1_p1}
\end{figure}

\begin{figure}
    \centering
    \includegraphics[width=1\linewidth]{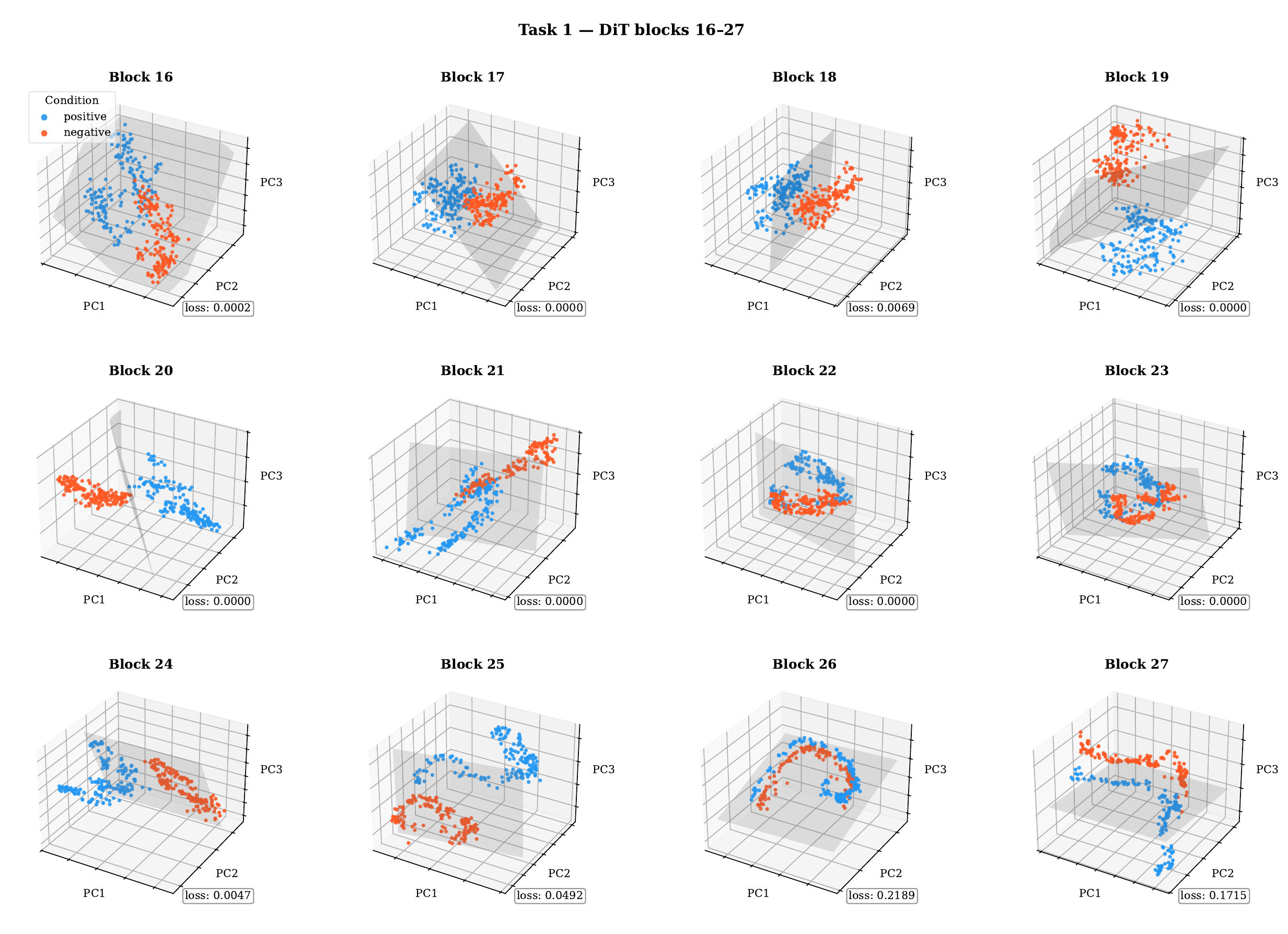}
    \caption{Cosmos-Policy Task 1 noise corruption}
    \label{fig:cosmos_allblocks_t1_p2}
\end{figure}

\begin{figure}[h]
    \centering
    \includegraphics[width=1\linewidth]{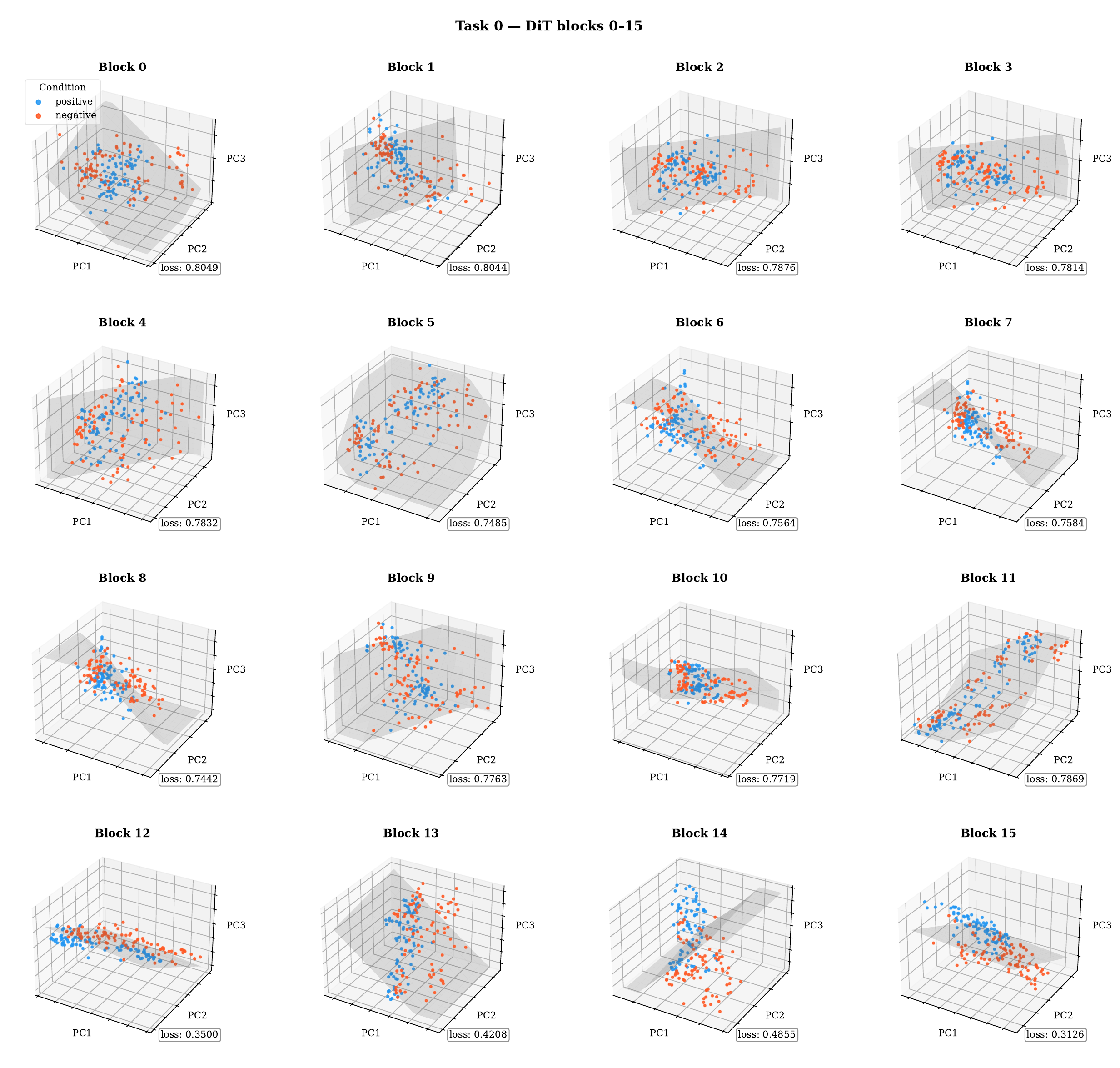}
    \caption{Cosmos-Policy Task 0 camera perturbation}
    \label{fig:cosmos_allblocks_t0_p1_cam}
\end{figure}

\begin{figure}
    \centering
    \includegraphics[width=1\linewidth]{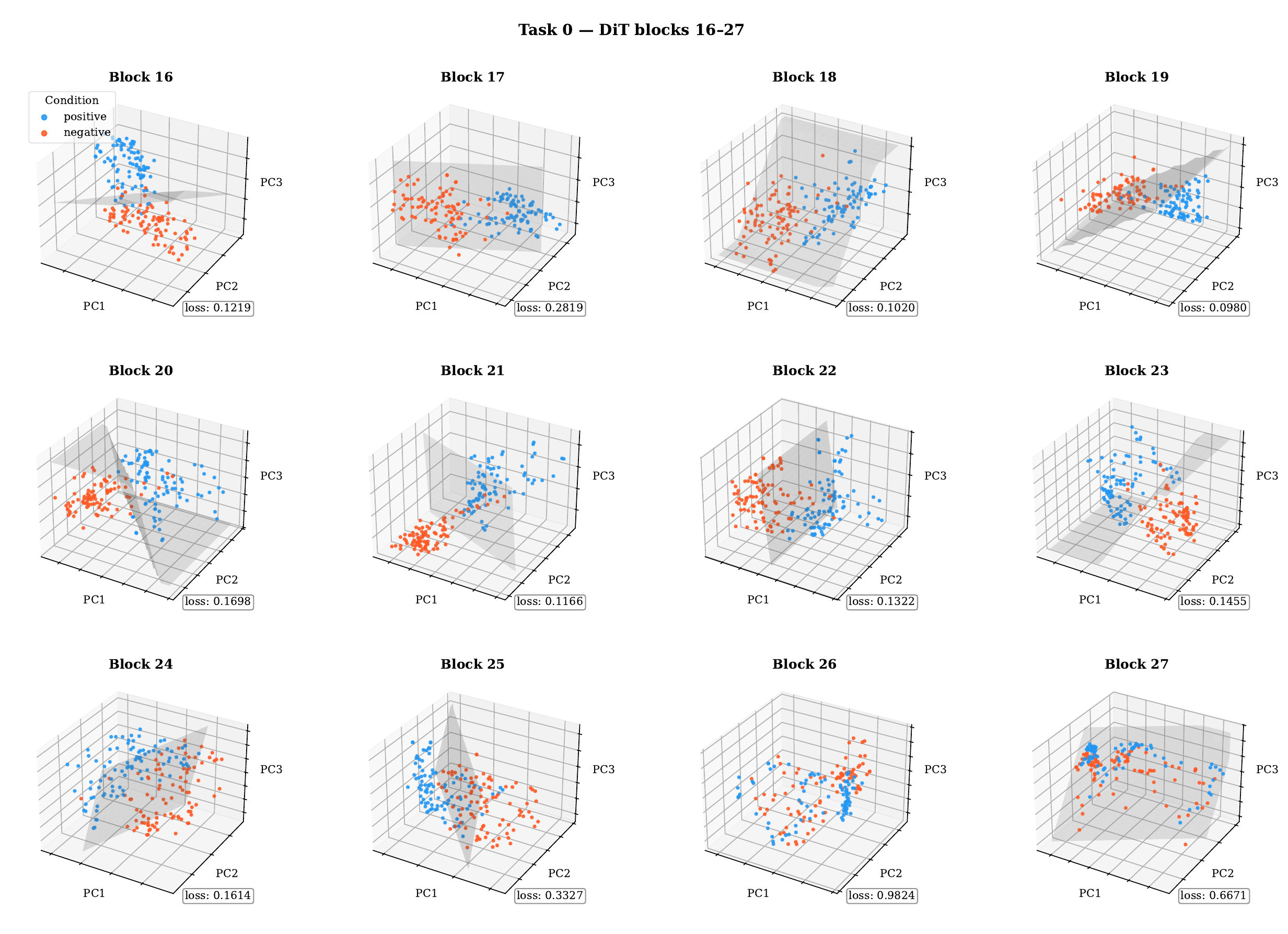}
    \caption{Cosmos-Policy Task 0 camera perturbation}
    \label{fig:cosmos_allblocks_t0_p2_cam}
\end{figure}

\begin{figure}
    \centering
    \includegraphics[width=1\linewidth]{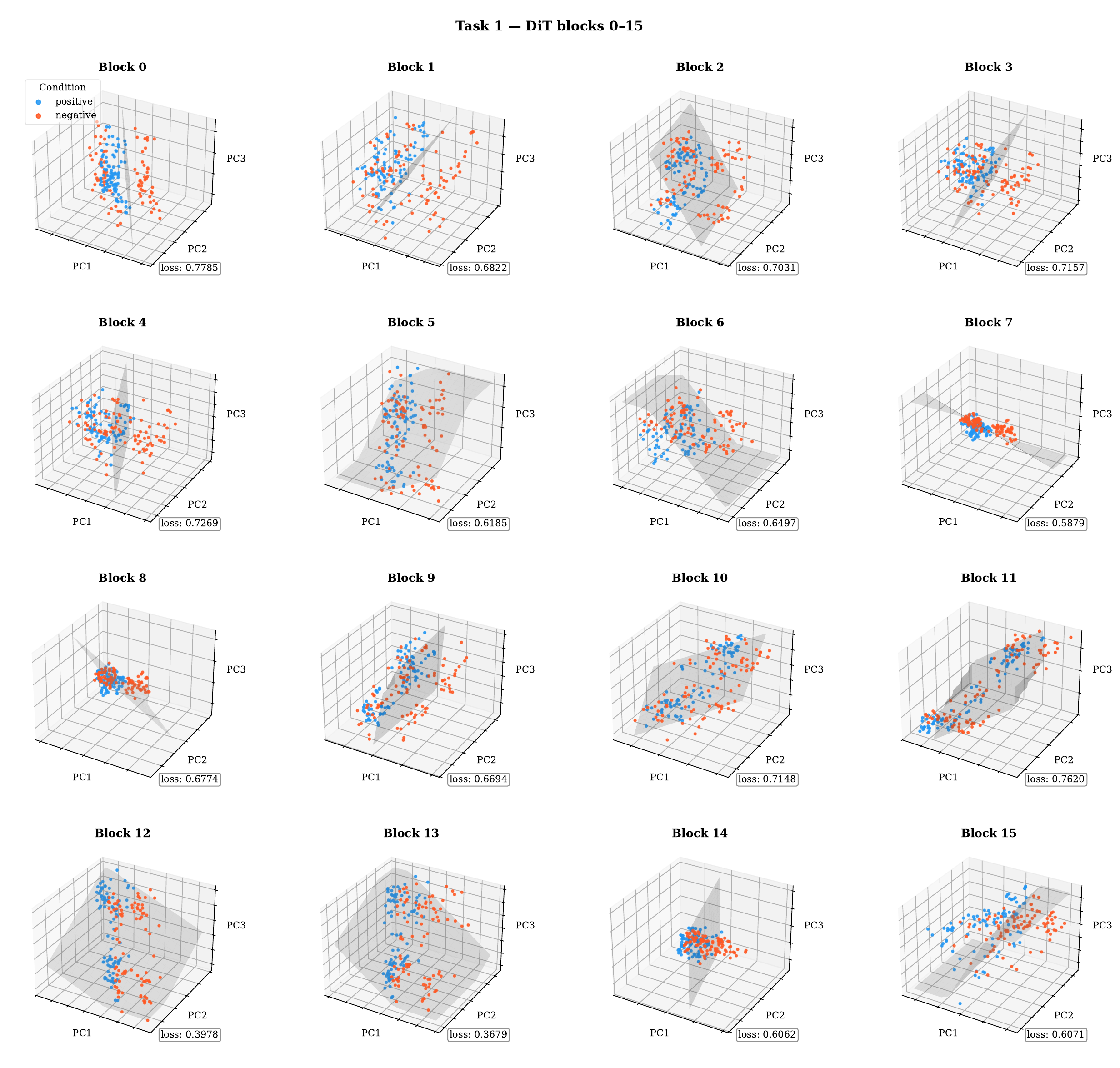}
    \caption{Cosmos-Policy Task 1 camera perturbation}
    \label{fig:cosmos_allblocks_t1_p1_cam}
\end{figure}

\begin{figure}
    \centering
    \includegraphics[width=1\linewidth]{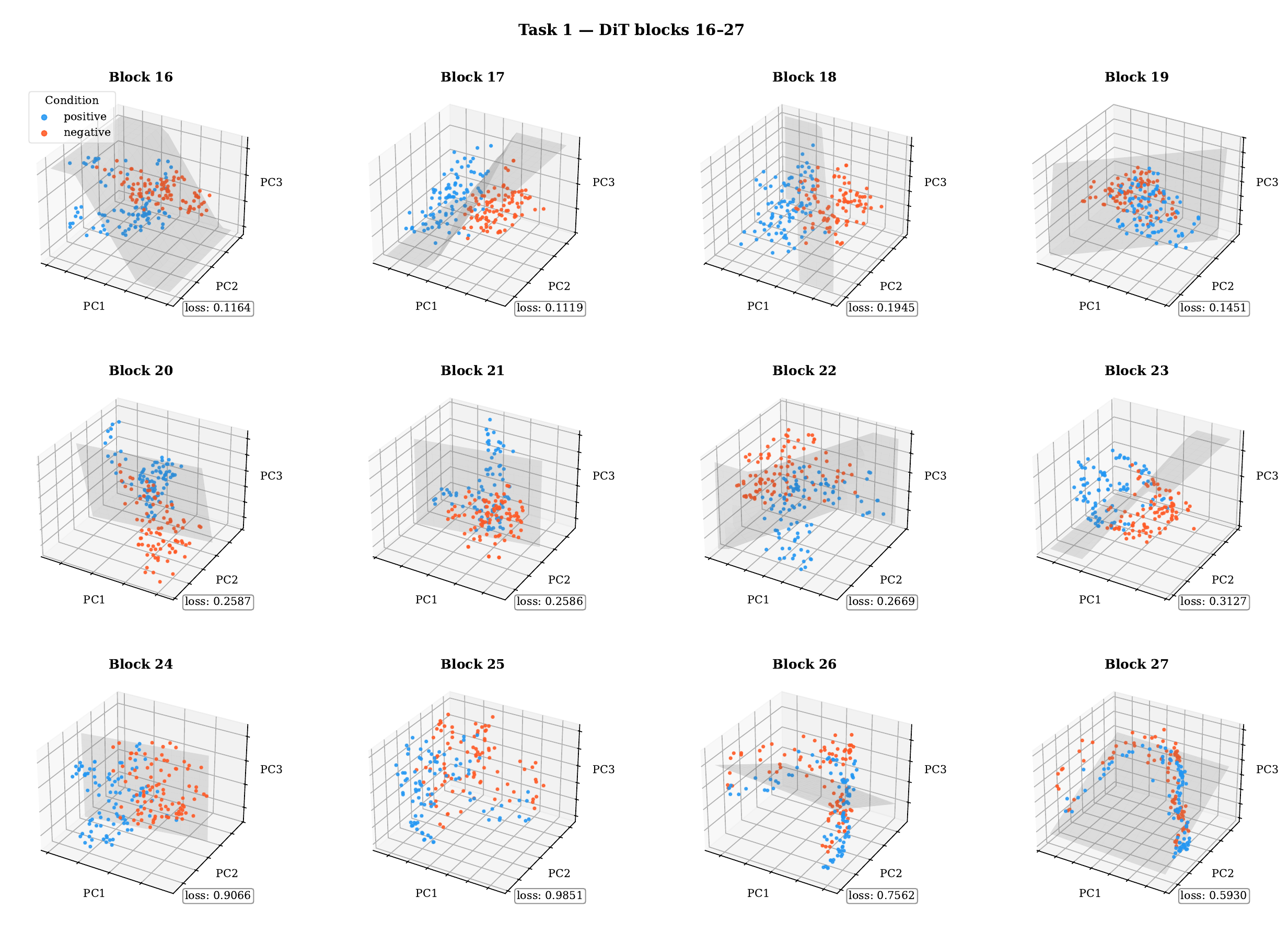}
    \caption{Cosmos-Policy Task 1 camera perturbation}
    \label{fig:cosmos_allblocks_t1_p2_cam}
\end{figure}

\begin{figure}[h]
    \centering
    \includegraphics[width=1\linewidth]{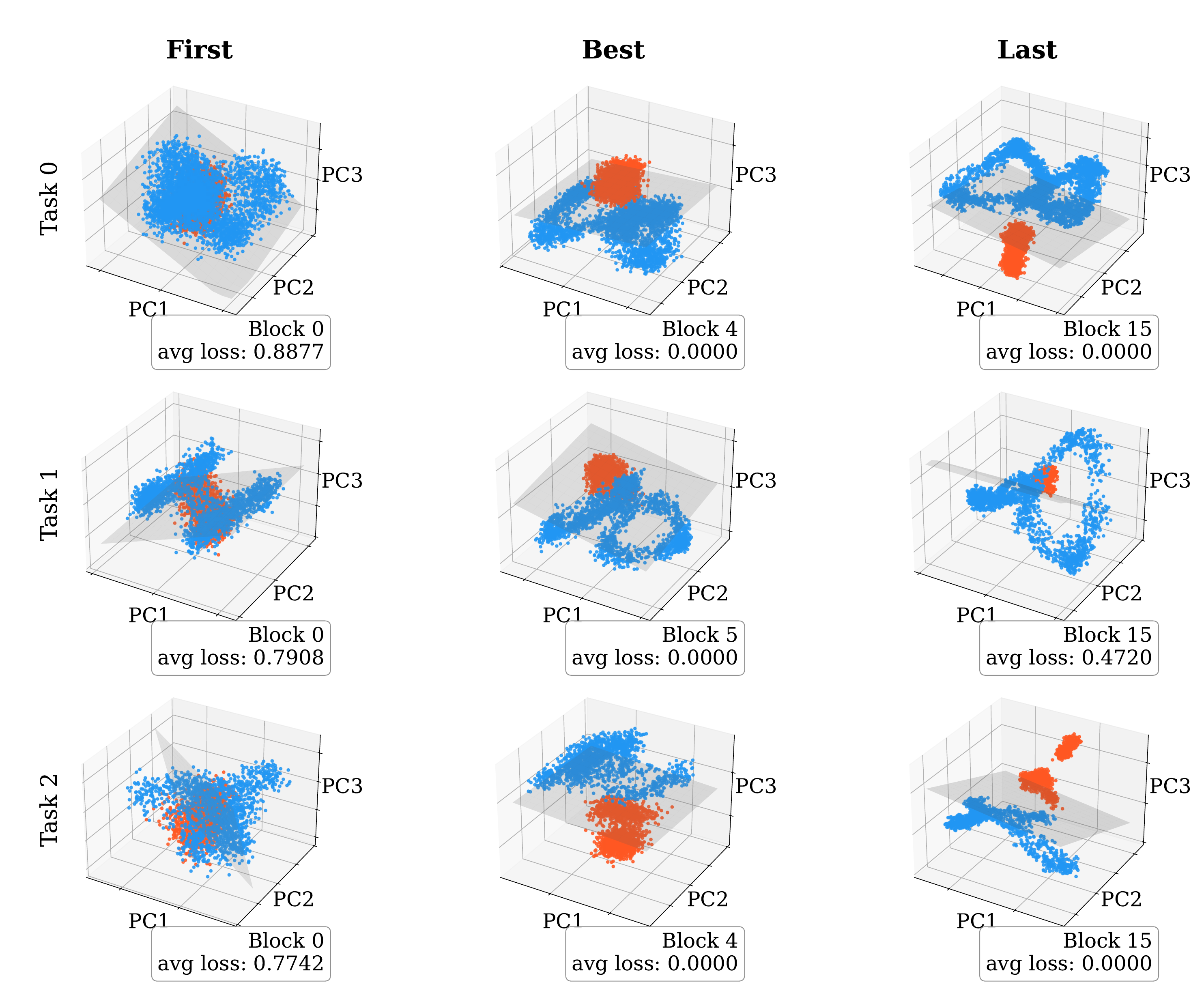}
    \caption{Feature separability: DiT4DiT Gaussian noise corruption for tasks 0-2.}
    \label{fig:dit4dit_noise}
\end{figure}

\begin{figure}
    \centering
    \includegraphics[width=1\linewidth]{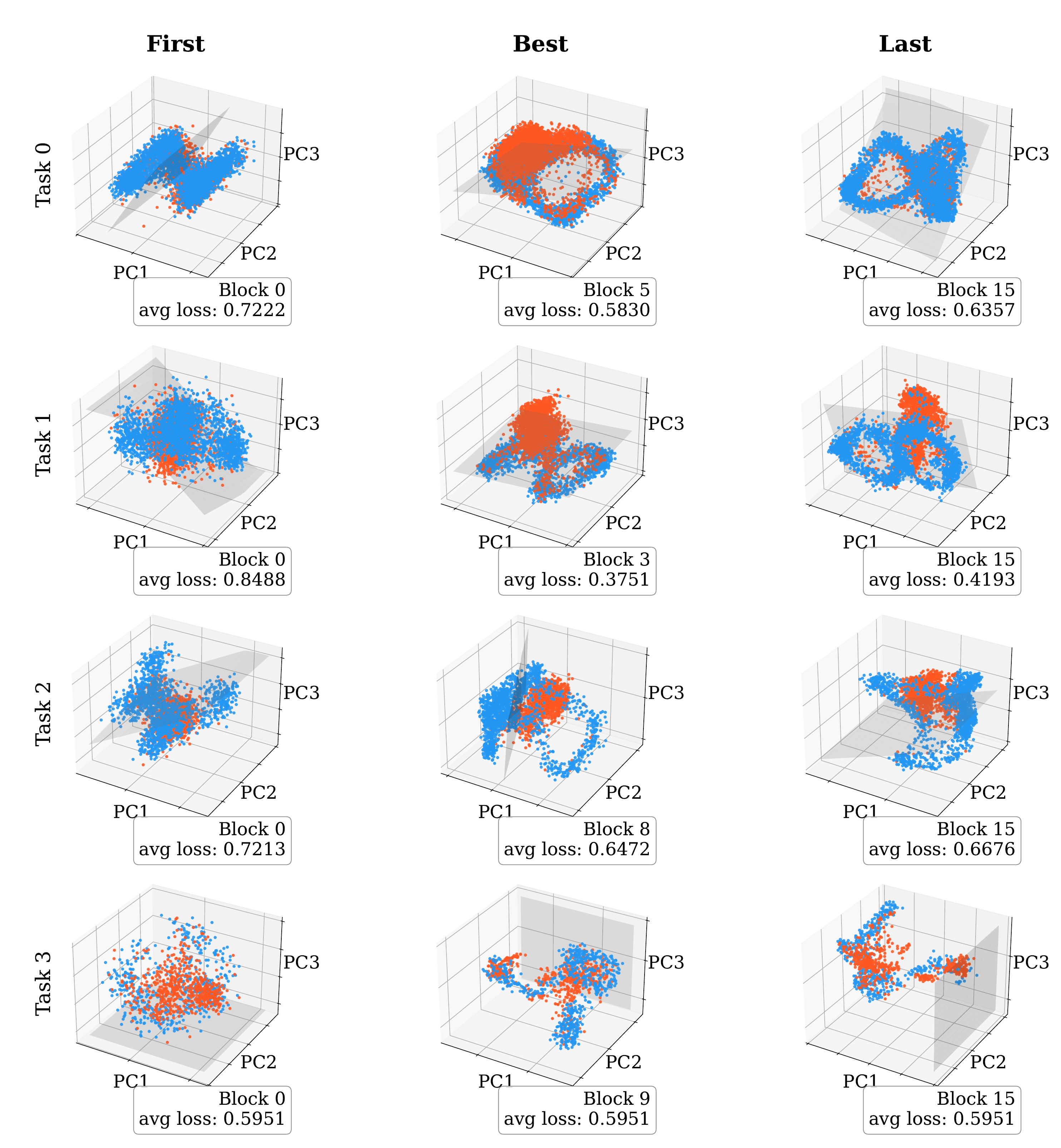}
    \caption{Feature separability: DiT4DiT gripper perturbation for tasks 0-3.}
    \label{fig:dit4dit_gripper}
\end{figure}

\begin{figure}
    \centering
    \includegraphics[width=1\linewidth]{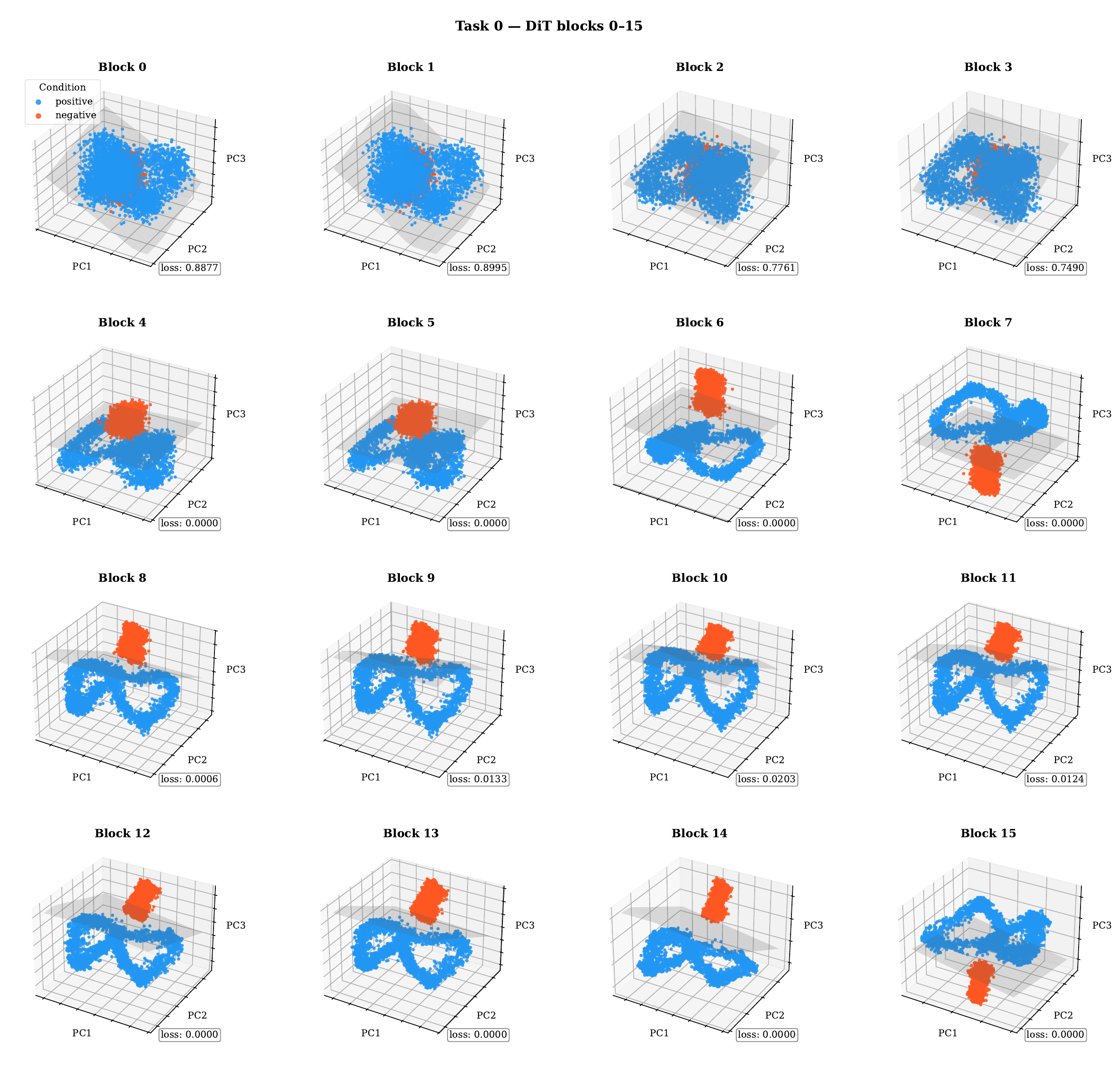}
    \caption{DiT4DiT Task 0 noise corruption}
    \label{fig:dit4dit_allblocks_t0}
\end{figure}

\begin{figure}
    \centering
    \includegraphics[width=1\linewidth]{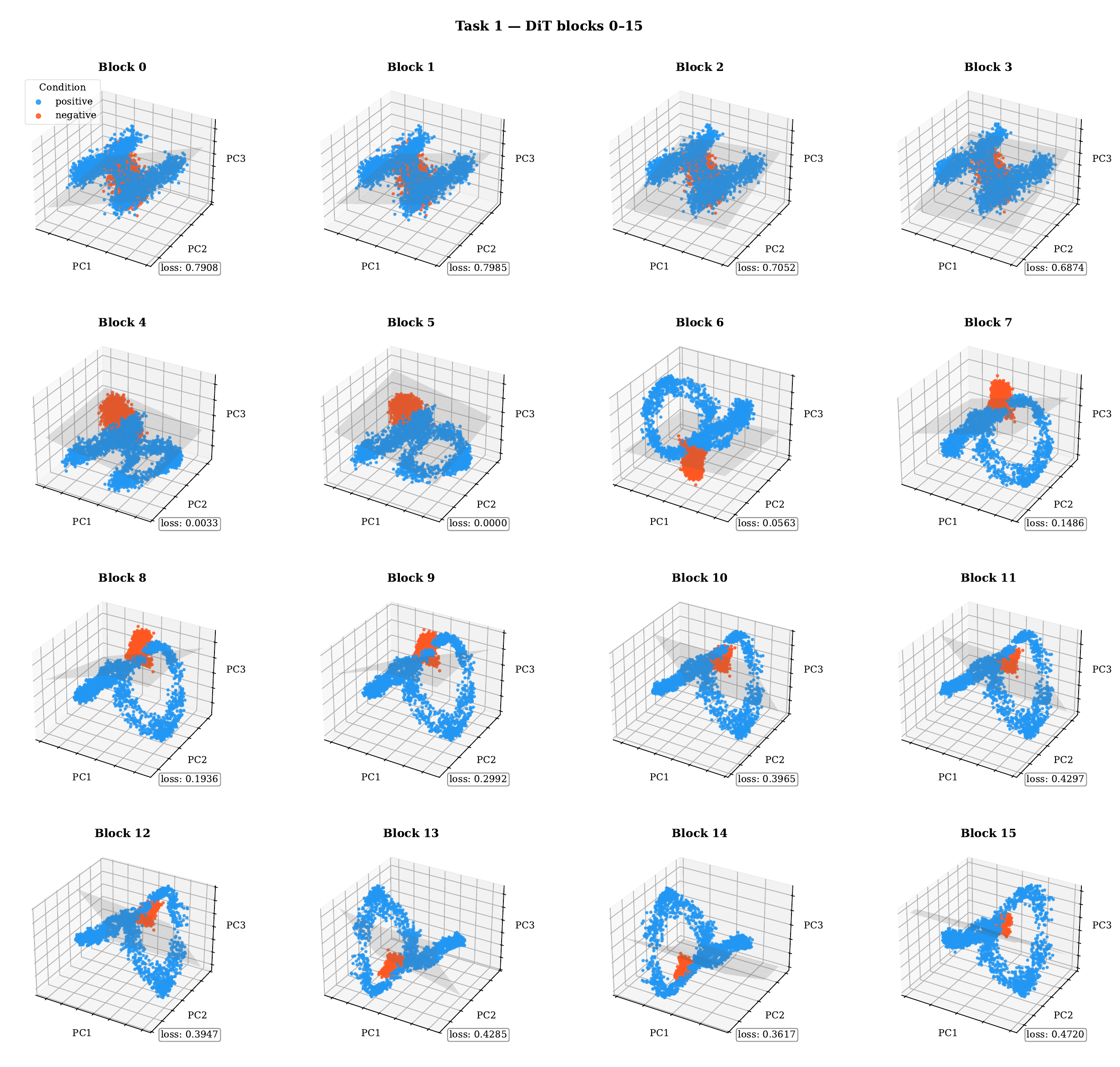}
    \caption{DiT4DiT Task 1 noise corruption}
    \label{fig:dit4dit_allblocks_t1}
\end{figure}

\begin{figure}
    \centering
    \includegraphics[width=1\linewidth]{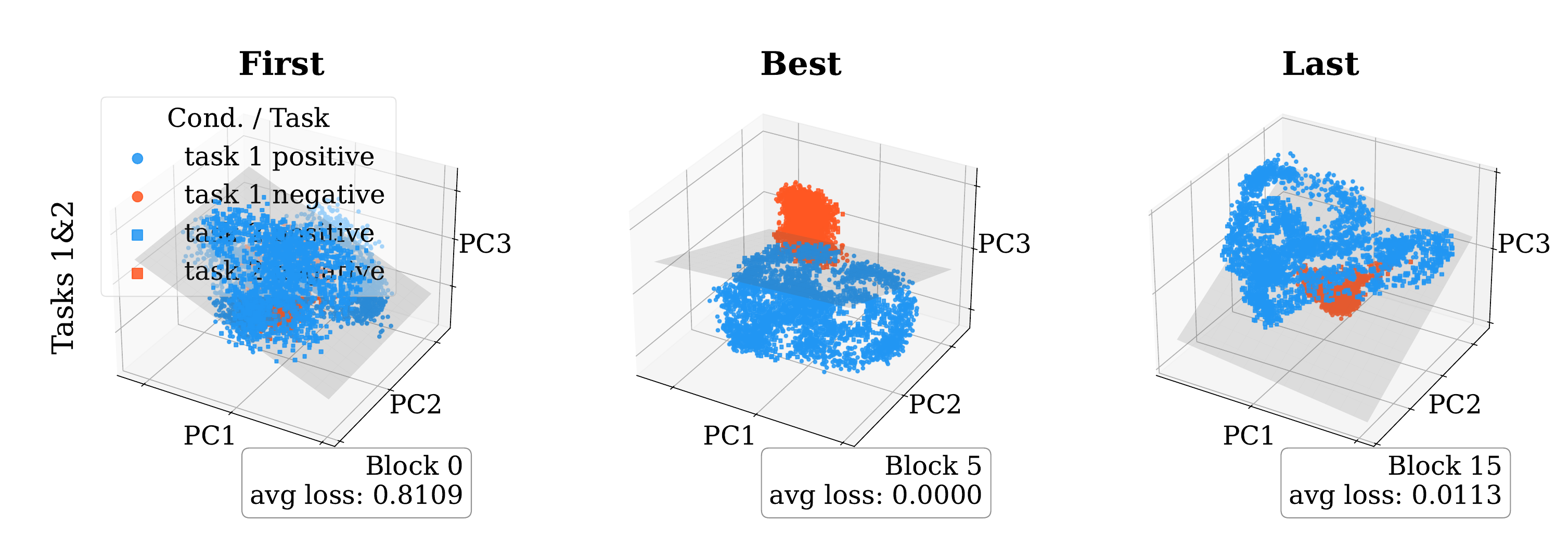}
    \caption{Pairwise Gaussian noise corruption feature separation on DiT4DiT.}
    \label{fig:dit4dit_noise_pairs}
\end{figure}

\begin{figure}
    \centering
    \includegraphics[width=1\linewidth]{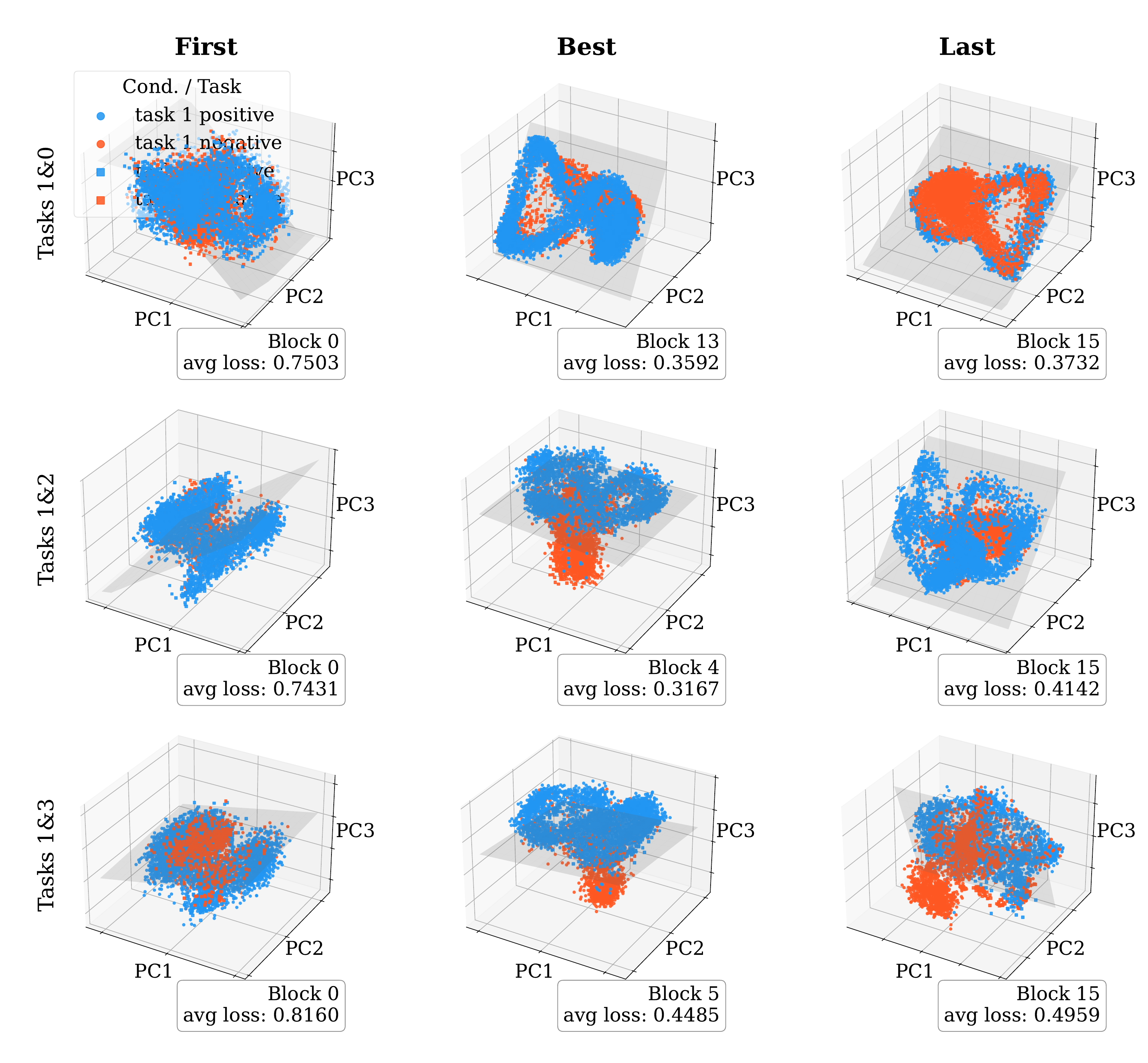}
    \caption{Pairwise gripper perturbation feature separation on DiT4DiT.}
    \label{fig:dit4dit_gripper_pairs}
\end{figure}

% \begin{figure}
%     \centering
%     \includegraphics[width=1\linewidth]{figs/lingbot_task00_allblocks_p1.pdf}
%     \caption{LingBot-VA Task 0 noise corruption}
%     \label{fig:lingbot_task00_allblocks_p1}
% \end{figure}

% \begin{figure}
%     \centering
%     \includegraphics[width=1\linewidth]{figs/lingbot_task00_allblocks_p2.pdf}
%     \caption{LingBot-VA Task 0 noise corruption}
%     \label{fig:lingbot_task00_allblocks_p2}
% \end{figure}

% \begin{figure}
%     \centering
%     \includegraphics[width=1\linewidth]{figs/lingbot_task01_allblocks_p1.pdf}
%     \caption{LingBot-VA Task 1 noise corruption}
%     \label{fig:lingbot_task01_allblocks_p1}
% \end{figure}

% \begin{figure}
%     \centering
%     \includegraphics[width=1\linewidth]{figs/lingbot_task01_allblocks_p2.pdf}
%     \caption{LingBot-VA Task 1 noise corruption}
%     \label{fig:lingbot_task01_allblocks_p2}
% \end{figure}

\end{document}